\documentclass[10pt,twocolumn,letterpaper]{article}

\usepackage[pagenumbers]{iccv}      
\usepackage{mathtools}
\usepackage{ulem}
\usepackage{comment}
\usepackage{amsmath, amsthm, amssymb}
\usepackage{array}
\usepackage{algorithm}
\usepackage{algpseudocode}
\usepackage{graphicx}
\usepackage{subcaption}
\usepackage{tikz} 
\usetikzlibrary{arrows.meta, positioning}
\usepackage[dvipsnames, table]{xcolor}

\usepackage{multirow, booktabs}
\usepackage{makecell}
\usepackage{tcolorbox}
\newtheorem{Proposition}{Proposition}
\usepackage{minitoc}

\usepackage{tikz}

\definecolor{iccvblue}{rgb}{0.21,0.49,0.74}
\usepackage[pagebackref,breaklinks,colorlinks,allcolors=iccvblue]{hyperref}

\def\paperID{2644} 
\def\confName{ICCV}
\def\confYear{2025}

\title{Multi-turn Consistent Image Editing}

\author{
Zijun Zhou$^{1}$ \quad
Yingying Deng$^{*}$ \quad
Xiangyu He$^{1}$ \quad
Weiming Dong$^{1}$ \quad
Fan Tang$^{2}$ \\
$^1$Institute of Automation, Chinese Academy
of Sciences, Beijing, China \\
$^2$Institute of Computing Technology,
Chinese Academy of Sciences, Beijing, China\\
}

\begin{document}
\twocolumn[{%
\renewcommand\twocolumn[1][]{#1}%
\maketitle

\begin{center}
    \captionsetup{type=figure}
    \includegraphics[width=\linewidth]{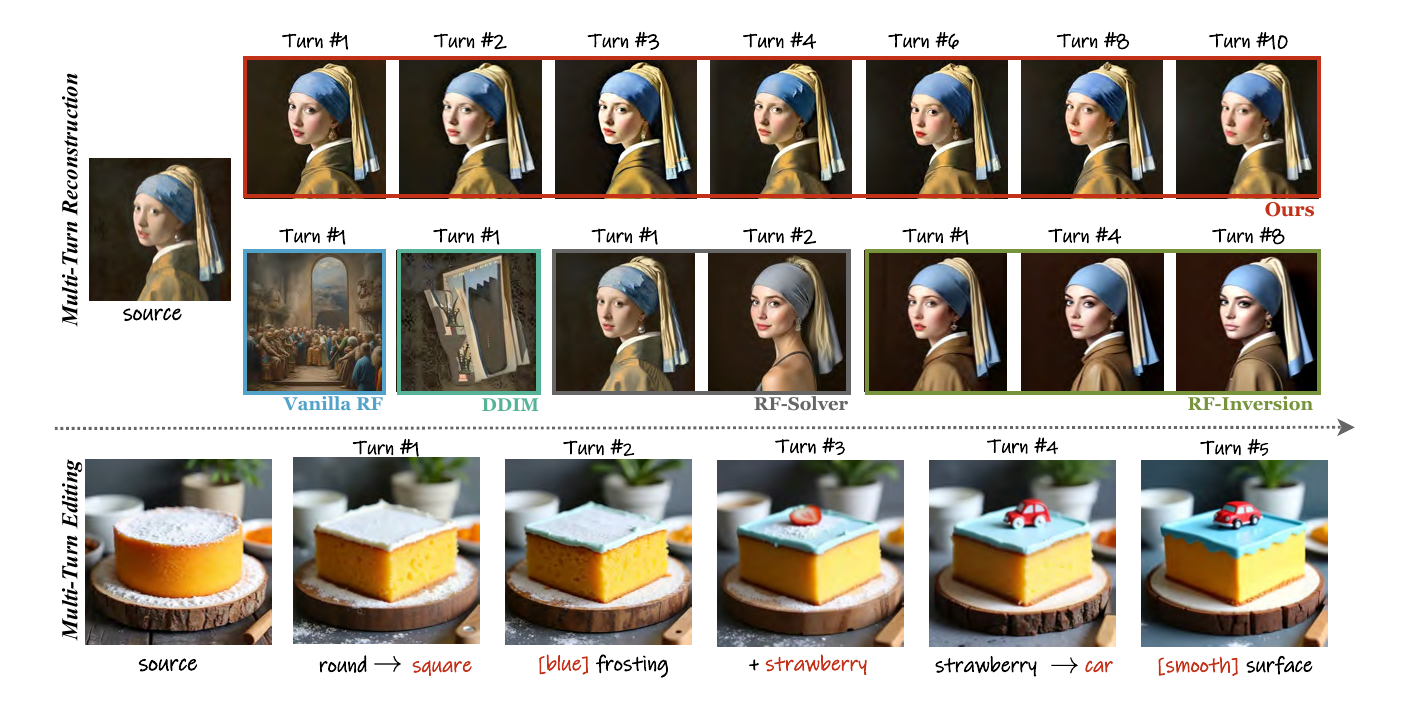}
    \vspace{-1.2cm}
    \caption{Our method efficiently preserves the original image's features during multi-turn image reconstruction. Additionally, it enables flexible editing capabilities in multi-turn editing tasks, providing the user with an iterative editing framework.}
    \label{fig:teaser}
\end{center}
}]
\renewcommand{\thefootnote}{} 
\footnotetext{*Corresponding author}
\renewcommand{\thefootnote}{\arabic{footnote}}

\begin{abstract}
Many real-world applications, such as interactive photo retouching, artistic content creation, and product design, require flexible and iterative image editing. However, existing image editing methods primarily focus on achieving the desired modifications in a single step, which often struggles with ambiguous user intent, complex transformations, or the need for progressive refinements. 
As a result, these methods frequently produce inconsistent outcomes or fail to meet user expectations.
To address these challenges, we propose a multi-turn image editing framework that enables users to iteratively refine their edits, progressively achieving more satisfactory results. 
Our approach leverages flow matching for accurate image inversion and a dual-objective Linear Quadratic Regulators (LQR) for stable sampling, effectively mitigating error accumulation.
Additionally, by analyzing the layer-wise roles of transformers, we introduce a adaptive attention highlighting method  that enhances editability while preserving multi-turn coherence.
Extensive experiments demonstrate that our framework significantly improves edit success rates and visual fidelity compared to existing methods.
\end{abstract}    
\begin{figure*}[htbp]
    \centering
    \includegraphics[width=\linewidth]{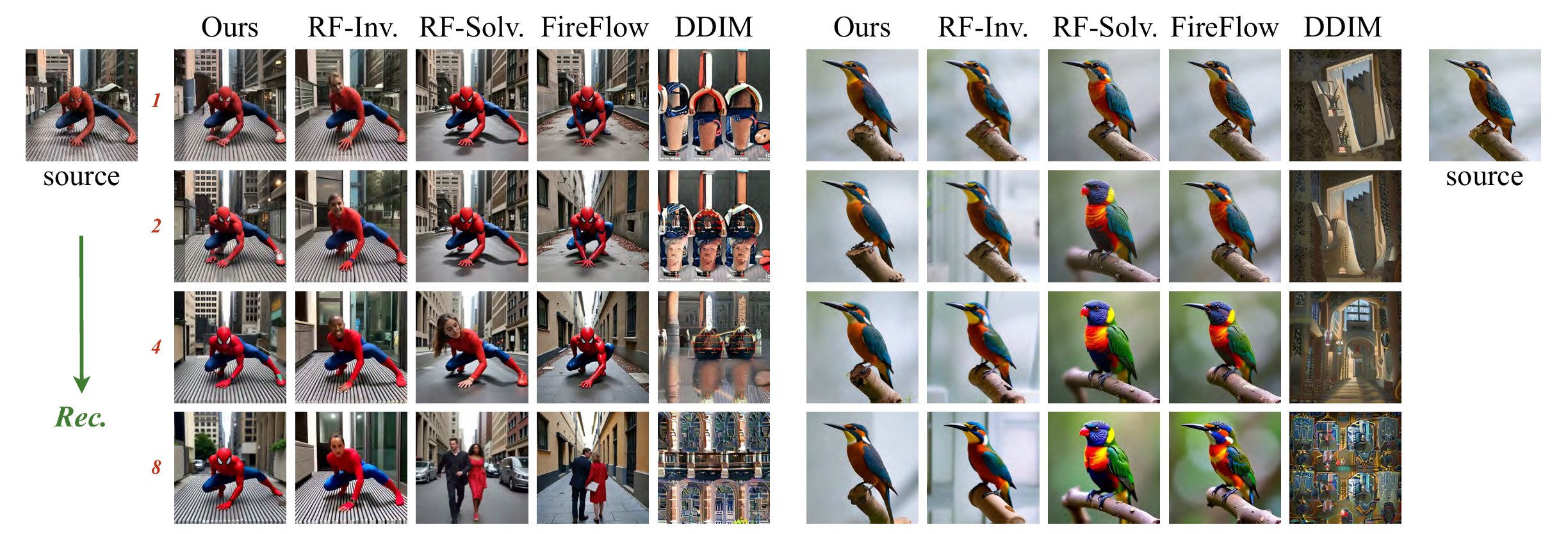}
    \caption{\textbf{Multi-turn Reconstruction Results.} This figure compares image reconstructions using our method and baseline methods across 1, 2, 4, and 8 reconstruction iterations. Our method effectively preserves color, background, structure, and semantic consistency across multiple reconstruction rounds, outperforming the baseline methods.}
    \label{fig:multi_turn_rec}
    \vspace{-0.6cm}

\end{figure*}

\section{Introduction}
\label{sec:intro}

Current image editing methodologies often strive for a single-step editing solution that perfectly aligns with a given textual prompt. 
This paradigm, however, proves inadequate for practical applications like product design, where user specifications are often inherently ambiguous and necessitate progressive refinement. 
A more effective framework should incorporate iterative editing capabilities, enabling users to sequentially refine outputs through multiple editing cycles. 
Such an approach would provide enhanced control over the final result by allowing continuous adjustments based on intermediate outcomes, as illustrated in ~\cref{fig:teaser}.
Consequently, further exploration of multi-turn image editing frameworks is essential to unlock their potential for iterative image refinement.

An intuitive approach to multi-turn image editing involves directly integrating existing single-step methods, leveraging the significant advancements in diffusion-based inversion \cite{song2020denoising, ho2020denoising, nichol2021improved, mokady2023null, miyake2023negative, ju2023direct} and related editing techniques.
These single-step methods often employ techniques such as attention map replacement \cite{cao2023masactrl, hertz2022prompt, tumanyan2023plug, nam2024dreammatcher, gu2024swapanything, chefer2023attend, kumari2023multi, deng2023z}, mask application \cite{cao2023masactrl, couairon2022diffedit, huang2023region}, and domain-specific pre-trained models~\cite{huang2024diffstyler, kim2022diffusionclip, wang2023stylediffusion} to mitigate inversion inaccuracies and preserve image structure. However, this strategy often lacks the robustness required for reliable multi-turn editing, as these techniques are insufficient to prevent the accumulation of errors across multiple iterations.
Consequently, edited results in multi-turn frameworks tend to exhibit increasing artifacts and semantic biases, deviating significantly from natural image characteristics.

Flow matching \cite{lipman2023flow, liu2022flow, esser2024scaling} has emerged as a powerful technique for image generation and editing. By directly estimating the transformation from noisy to clean images, rather than predicting noise as in diffusion-based methods, flow matching offers a more efficient and direct framework. This results in simplified distribution transfer, fewer inference steps, and ultimately, more precise editing and reconstruction. This has led to its adoption in state-of-the-art models like SD3 \cite{esser2024scaling} and FLUX.1-dev \cite{FLUX1dev}.
Existing image editing research has explored flow matching \cite{liu2022flow, lipman2023flow, albergo2023building, esser2024scaling, avrahami2024stable} as a method for accurate image inversion in single-turn editing.
Beyond single-turn editing, ReFlow-based models have significant potential for multi-turn editing due to their efficiency in inference steps and accurate inversion, which are crucial prerequisites for this task. 
However, as shown in Figure \ref{fig:multi_turn_rec}, challenges such as accumulated errors in multi-turn editing still need to be addressed. Additionally, the trade-off between preserving content and ensuring sufficient editing flexibility in a multi-turn framework remains unexplored.

In this paper, we present a novel framework that leverages FLUX models to facilitate robust and controllable multi-turn image editing.
To ensure long-term coherence and restrict the distribution of edited images in multi-turn tasks, we integrate a dual-objective Linear Quadratic Regulator (LQR) control mechanism into our framework. 
This LQR mechanism considers both the outputs of preceding turns and the initial input image, establishing a long-term dependency in the editing process.
Although dual-objective LQR's stabilization capability is essential for reliable multi-turn editing, the method's stringent regularization constraints may inadvertently reduce editing flexibility.
To achieve a balance between stability and flexibility during the editing process, we propose an adaptive attention guidance method aimed at directing the editing focus toward salient regions. This adaptive attention mechanism utilizes medium-to-low activated regions as spatial guidance signals to generate a probabilistic editing mask. By employing attention reweighting, this approach selectively concentrates on target areas while preserving non-target regions.

The key contributions are summarized as follows:

\begin{itemize}
\item A dual-objective Linear Quadratic Regulator (LQR) approach that builds upon the flow matching inversion process to ensure stable image distribution across multiple editing turns.
\item An adaptive attention mechanism, guided by analysis of intermediate attention layers within the DiT architecture, to enhance the precision and localization of edits.
\item A multi-turn interactive image editing framework that empowers users to iteratively refine images with consistent and predictable results.
\end{itemize}
The subsequent sections of this paper are structured as follows: ~\cref{sec:Preliminary} introduces preliminaries on rectified flow's high-order solvers and LQR control. ~\cref{sec:motivation} outlines the motivation behind our framework, including the dual-objective LQR and high-order ODE solver. 
~\cref{sec:method} details our methodology, covering dual-objective LQR guidance and adaptive attention guidance.
Finally,  ~\cref{sec:experiment} presents both quantitative and qualitative experimental results.
A more detailed discussion of related work, technical proofs, and experiments can be found in the supplemental material.


\section{Related Work}
\label{sec:related work}
\paragraph{Image Inversion:}
\label{sec2_1:Inversion}
Image inversion transforms a clean image into a latent Gaussian noise representation. Sampling from this representation enables controlled image editing and reconstruction.
Diffusion-based inversion originated with DDPM~\cite{ho2020denoising,nichol2021improved}, which progressively add noise to an image. DDIM~\cite{song2020denoising}, a deterministic variant of DDPMs, allows for significantly faster inversion. However, early inversion techniques often lacked sufficient accuracy. Null-text inversion~\cite{mokady2023null} addresses this limitation by optimizing a null-text embedding, effectively leveraging the inherent bias of the inversion process. Negative-Prompt-Inversion~\cite{miyake2023negative} mathematically derives the optimization process of null-text inversion, thereby accelerating the inversion process. Direct-Inversion~\cite{ju2023direct} incorporates the inverted noise corresponding to each timestep within the denoising process to mitigate content leakage.


\paragraph{Image Editing:}
\label{sec2_3:Image Editing}
To maintain the consistency of edited images with the source image, several approaches constrain the editing results. One strategy, employed by \cite{zhang2023adding, stracke2024ctrloralter, ye2023ip}, involves tuning additional parameters to inject source image information or providing structural control through masks, canny edges, or depth maps. Another prominent approach, stemming from Prompt2Prompt \cite{hertz2022prompt}, manipulates attention maps to preserve image structure, as seen in various editing methods \cite{cao2023masactrl, tumanyan2023plug, chefer2023attend, zhang2024attention, kumari2023multi, deng2023z, chung2024style, hertz2024style, liu2024towards, chen2024training, wang2024compositional, nam2024dreammatcher, guo2024focus}. Furthermore, mask-based techniques have proven effective in enhancing both preservation and editability. For instance, \cite{couairon2022diffedit, cao2023masactrl, huang2023region, liu2023text} utilize automatically generated masks for more accurate text-guided image generation.
Flow-based image editing methods \cite{liu2022flow, lipman2023flow, albergo2023building, esser2024scaling, avrahami2024stable} have demonstrated strong performance in single-turn editing. Building upon this foundation, RF-Inversion \cite{rout2024semantic} employs a single-objective LQR control framework.
FireFlow \cite{deng2024fireflow} and RF-Solver \cite{wang2024taming} further refine the process by focusing on reducing single-step simulation error through second-order ODE solvers.
Our work addresses the specific challenges of multi-turn editing.
We leverage second-order ODEs for accurate single-step inversion and, crucially, introduce a dual-objective LQR and adaptive attention guidance to maintain coherence and control across multiple editing steps.

Joseph et al. \cite{joseph2024iterative} explored techniques for manipulating images directly within the latent space. ChatEdit \cite{cui2023chatedit} and TextBind \cite{li2024textbind} leverage the summarization capabilities of LLMs to streamline editing workflows. Similarly, Yang et al.~\cite{yang2023idea2img} employ a self-refinement strategy using GPT-4V~\cite{gpt4} to support interactive image editing. 
While these methods enhance editing efficiency, they underutilize the image generation model’s full potential.
In contrast, our work focuses on multi-turn image editing by directly optimizing the image generation model’s capabilities, enabling consistent edits across multiple iterations without relying on external language models.


\section{Preliminary}
\label{sec:Preliminary}
\paragraph{Rectified Flow:}
\label{sec3_1:Preliminaries}
Liu et al.~\cite{liu2022flow} proposed an ordinary differential equation (ODE) model to describe the distribution transfer from $x_0 \sim \pi_0$ to $x_1 \sim \pi_1$. They defined this transfer as a straight-line path, given by $x_t = t x_1 + (1 - t) x_0$, where $t \in [0,1]$. This can be expressed as the following differential equation:
\begin{equation}
\frac{d x_t}{d t}=x_1-x_0.
\end{equation}
To model this continuous process, they sought a velocity field $v$ that minimizes the objective:
\begin{equation}
\min _v \int_0^1 \mathbb{E}\left[\left\|\left(x_1-x_0\right)-v\left(x_t, t\right)\right\|^2\right] \mathrm{d} t.
\end{equation}
In practice, this continuous ODE is approximated using a discrete process, where the velocity field $v(x_t, t)$ is parameterized by a neural network. Typically, $x_1 \sim \pi_1$ is assumed to be Gaussian noise, and $x_0 \sim \pi_0$ represents the target image. The discrete inversion process is then formulated as:
 \begin{equation}
     X_{t+\Delta t} = X_{t} + v(\theta, t)\Delta t,
 \end{equation}
where $v(\theta)$ denotes the neural network with parameters $\theta$.

\begin{figure*}[t]
  \centering
  \begin{subfigure}{\linewidth}
\includegraphics[width=\linewidth]{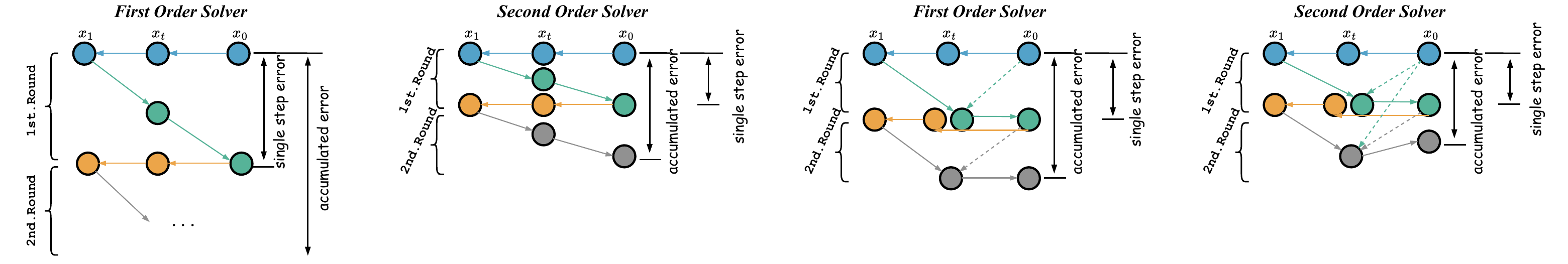}
  \end{subfigure}
 
  \begin{subfigure}{0.22\linewidth}
    \centering
    \caption{Vanilla ReFlow with $1^{st}$ order ODE solver}
  \end{subfigure}
  \hfill
  \begin{subfigure}{0.22\linewidth}
    \centering
    \caption{FireFlow/RF-solver with $2^{nd}$ order ODE solver}
    \label{fig:motivation_2}
  \end{subfigure}
  \hfill
  \begin{subfigure}{0.22\linewidth}
    \centering
    \caption{RF-inversion with $1^{st}$ order ODE solver}
    \label{fig:motivation_3}
  \end{subfigure}
  \hfill
  \begin{subfigure}{0.22\linewidth}
    \centering
    \caption{Our approach with $2^{nd}$ order ODE Solver}
    \label{fig:motivation_4}
  \end{subfigure}
\vspace{-2mm}
  \caption{We visualize the differences in single-step and multi-round accumulative errors during inversion ($\leftarrow$) and editing ($\searrow$) across different ReFlow-based editing methods. (a) Vanilla ReFlow struggles with structure preservation during inversion due to the truncation error of the Euler method. (b) While a second-order ODE solver reduces truncation error in a single step, the accumulated error over multiple editing rounds remains significant. (c) Incorporating the source image as guidance (dotted $\swarrow$) via LQR improves performance in a single step but becomes less effective as accumulated errors increase with more steps. (d) Our approach addresses this issue by integrating both techniques, leveraging a dual-objective LQR coupled with a high-order solver to enhance stability and accuracy.}
  \label{fig:motivation}
\end{figure*}

\paragraph{High-order Solver:} To enhance the accuracy of this discretization, 
RF-Solver~\cite{wang2024taming} and Fireflow~\cite{deng2024fireflow} employ second-order ODE solvers, which reduce the approximation error from $\mathcal{O}(\Delta t^2)$ to $\mathcal{O}(\Delta t^3)$ for the same step size $\Delta t$ as used in standard ODE methods. This improvement enables comparable results with fewer sampling steps. In practice, these methods implement the standard midpoint method, increasing accuracy by evaluating the velocity field at an intermediate point. 
In the discrete setting, for time $t \in [0, 1]$ and a positive time increment $\Delta t>0$, the inversion process updates the state forward in time according to:
 \begin{equation}
 \label{equ:midpoint}
     X_{t+\Delta t} = X_{t} + v(\theta, t+\frac{\Delta t}{2})\Delta t.
 \end{equation}
Additionally, Fireflow~\cite{deng2024fireflow} introduces an acceleration technique by caching intermediate velocity field results, reducing the required sampling steps to eight with the same truncation error as midpoint method.

\paragraph{Linear Quadratic Regulator (LQR) Control:}


RF-Inversion~\cite{rout2024semantic} introduces the Linear Quadratic Regulator (LQR) method to effectively guide image generation. When dealing with images or noise originating from atypical distributions, an explicit guidance term is incorporated. This ensures that images from atypical distributions can be inverted into typical noise, and likewise, atypical noise can be transformed back into typical images. Assuming \(x_1 \sim \pi_1\) represents the Gaussian noise space and \(x_0 \sim \pi_0\) represents the image space, the discrete inversion process over time \(t \in [0,1]\) is described by:
\begin{equation}
\label{equ:lqr_inversion}
X_{t + \Delta t} = X_t + [v_t(X_t) + \eta(v_t(X_t \mid X_1) - v_t(X_t))] \Delta t.
\end{equation}
This process guides the inversion toward typical noise. In this equation, \(v_t(x_t \mid x_1)\) is derived by solving an LQR problem, resulting in \(v_t(x_t \mid x_1) = \frac{x_1 - x_t}{1-t}\).




\section{Motivation}
\label{sec:motivation}
\paragraph{Single step error v.s. multi-round error.}
In image editing, flow matching acts as a discrete approximation of a continuous ordinary differential equation (ODE). While employing high-order solvers~\cite{deng2024fireflow} or increasing the number of timesteps~\cite{wang2024taming} reduces single-step errors—potentially enhancing editing performance in theory—practical implementations encounter notable challenges under multi-round constraints. When the forward and reverse processes are performed multiple times, especially in iterative editing scenarios, multi-round truncation errors become a significant concern as shown in~\cref{fig:multi_turn_rec}.

Multi-round truncation error arises not just from individual steps but from the accumulation of these errors over a sequence of operations. High-order methods do minimize local truncation errors, but when these methods are applied iteratively, the cumulative error can become substantial, overshadowing initial gains in precision from reducing single-step errors. The reversibility of the process also introduces an additional layer of complexity. Numerical methods are typically not perfectly reversible; the pathway through which errors propagate in the forward direction may differ from that in the reverse direction. This asymmetry can further exacerbate the accumulation of errors, especially over multiple editing cycles.


In practical applications of the ReFlow model, these considerations highlight the limitations of reducing single-step error alone, as shown in~\cref{fig:motivation_2}. Instead, comprehensive strategies are needed to address the cumulative nature of global truncation errors and stability challenges in multi-round editing processes.



\paragraph{Single step guidance v.s. multi-turn guidance.} Another group of methods \cite{rout2024semantic} relies on the source image as a reference, performing precise single-step edits. However, these methods falter in multi-round editing contexts where cumulative error becomes a critical issue. The crux of the problem lies in the way the original LQR-based approach references only the last edited image $Y_{i}$, gradually diverging from the source image $Y_0$
over multiple iterations. While well-suited for single-step optimization as $Y_{i}$ equals to $Y_0$, this technique accumulates discrepancies across successive rounds of edits due to its inability to realign with the original image's core characteristics, as shown in~\cref{fig:motivation_3}.

Multiple condition generation addresses this shortcoming by incorporating both $Y_0$ and $Y_n$ as simultaneous conditions for transformation. This dual-reference approach ensures that each round of editing remains anchored to the source image's foundational elements, thereby minimizing drift over time, shown in Figure \ref{fig:motivation_4}.


\section{Method}
\label{sec:method}

\begin{figure}[t]
  \centering
   \includegraphics[width=1.\linewidth]{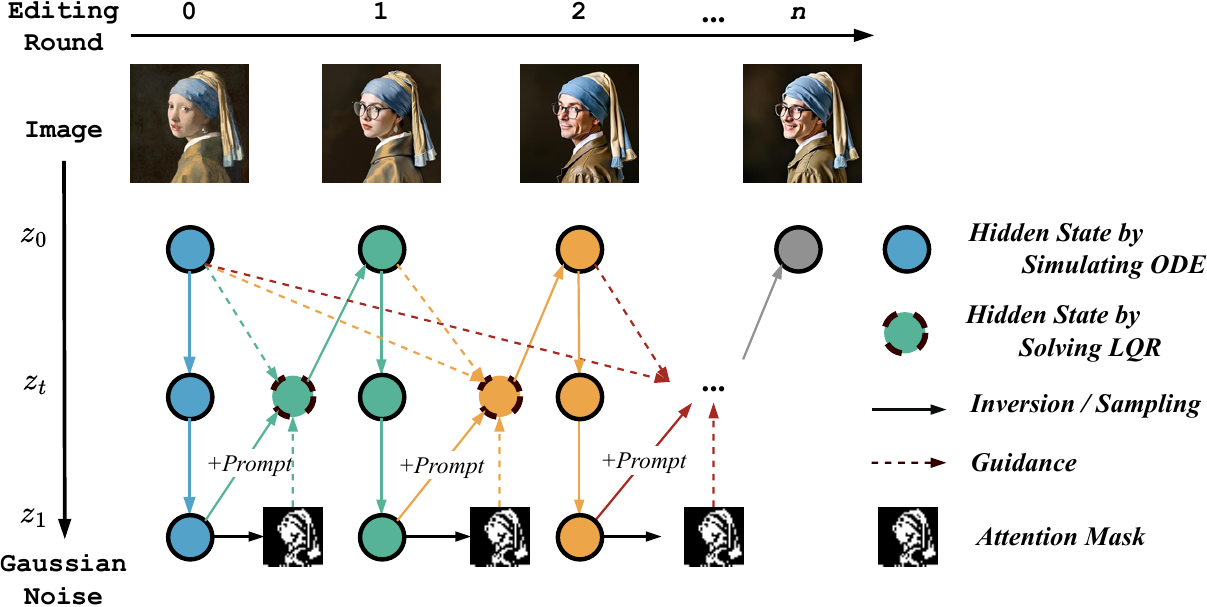}

   \caption{\textbf{Multi-turn editing pipeline.} In each editing iteration, a high-accuracy rectified flow inversion maps the image back to the Gaussian noise space, followed by sampling to generate the edited images. To better constrain the distribution of edits across multiple turns, the original image and previous editing results serve as guidance during subsequent sampling. Additionally, a highlighted region in the attention mask further preserves the content structure of the edited outputs.
   }
   \label{fig:pipeline}
\end{figure}

\subsection{Dual-objective LQR Guidance}
\label{sec4_1:multi turn guidance}

We develop an optimal control strategy to efficiently transform any image $X_0$ (whether corrupted or not) into a state that reflects multiple random noise conditions, represented by samples $X_1 \sim p_1, X_2 \sim p_2, \ldots, X_n \sim p_n$. 
\begin{equation}
\begin{split}
    \label{eq:multi-lqr}
    V(c) \coloneqq \int_0^1 \frac{1}{2} \left\|c\left(Z_t, t\right)\right\|_2^2 \, \mathrm{d}t + \sum_i^n \frac{\lambda_i}{2} \left\|Z_1 - X_i\right\|_2^2, \\
    \mathrm{d}Z_t = c\left(Z_t, t\right) \, \mathrm{d}t, \quad Z_0 = \mathbf{X}_0.
\end{split}
\end{equation}
This formulation is equivalent to leveraging a weighted average approach in a $d$-dimensional vector space $\mathbb{R}^d$ to achieve a balanced transformation:
\begin{equation}
\begin{split}
    \label{eq:multi-lqr}
    V(c) \coloneqq \int_0^1 \frac{1}{2} \left\|c\left(Z_t, t\right)\right\|_2^2 \, \mathrm{d}t + \frac{\lambda}{2} \left\|Z_1 - \hat{X}\right\|_2^2, \\
    \mathrm{d}Z_t = c\left(Z_t, t\right) \, \mathrm{d}t, \quad Z_0 = \mathbf{X}_0,
\end{split}
\end{equation}
where $\hat{X} = \frac{\sum_{i=1}^n \lambda_i X_i}{\sum_{i=1}^n \lambda_i}$ represents the weighted synthesis of the noise samples. The function $V(c)$ quantifies the total energy of the control $c: \mathbb{R}^d \times [0, 1] \rightarrow \mathbb{R}^d$. By optimizing $V(c)$ over the set of permissible controls, denoted by $\mathcal{C}$, we address the multi-condition generation challenge through a Linear Quadratic Regulator (LQR) framework.

\begin{Proposition}
\label{prop:multi-lqr}
Given $Z_0 = \mathbf{X}_0$ and the composite target $\hat{X}=\frac{\sum_{i=1}^n \lambda_i X_i}{\sum_{i=1}^n \lambda_i}$, the optimal control solution for the LQR problem \eqref{eq:multi-lqr}, denoted by $c^*\left(\cdot, t\right)$, aligns with the conditional vector field $u_t\left(\cdot|X_1,...,X_n\right)$, guiding the transformation along the interpolated path $X_t = t\hat{X} + (1-t)X_0$. Specifically, this results in $c^*\left(\mathbf{z}_t, t\right) = u_t\left(\mathbf{z}_t|\hat{X}\right) = \frac{\hat{X} - \mathbf{z}_t}{1-t}$.
\end{Proposition}

Based on Proposition 1 of dual-objective LQR guidance, we establish a framework for iterative image inversion and sampling, constraining the distribution of edited images per round to enable accurate and controlled editing. Additionally, we solve the second-order ODE (Equation~\ref{equ:midpoint}) using the FireFlow acceleration algorithm~\cite{deng2024fireflow}, enhancing the speed and editing capacity of single-step simulations within the framework.

In practice, we employ a single-objective LQR for the inversion process and a dual-objective LQR to guide the sampling process.
Let the clean image space be denoted by $x_0\sim \pi_0$ and the Gaussian noise space by $x_1\sim \pi_1$.
For inversion, we employ a single-objective LQR to map an image, whether corrupted or uncorrupted, back to the Gaussian noise space $\pi_1$, using a second-order ODE solver:
\begin{equation}
\begin{split}
    X_{t + \Delta t} &= X_t + \big[v_{t+\frac{\Delta t}{2}}(X_t) \\
    &\quad + \eta(v_{t+\frac{\Delta t}{2}}(X_t \mid X_{0}) + v_{t+\frac{\Delta t}{2}})\big] \Delta t.
\end{split}
\end{equation}
For the sampling process, we leverage the initial image and the result from the previous editing step as dual objectives to inform the LQR control within an invertible flow model.
Specifically, consider the $k$-th editing step, where the initial image is denoted as $X_{0,0}$ and the result from the $(k-1)$-th step as $X_{k-1,0}$.
Given a time step $\Delta t > 0$,
the dual-objective LQR sampling process is defined as follows:
\begin{equation}
\label{equ:double_lqr_sample}
\begin{cases}
\begin{aligned}
X_{t - \Delta t} &= X_t +  \big[-v_{t-\frac{\Delta t}{2}}(X_t) - \\
&\quad \eta(v_{t-\frac{\Delta t}{2}}(X_t \mid X_{\text{dual}}) + v_{t-\frac{\Delta t}{2}}(X_t))\big] \Delta t, \\
X_{\text{dual}} &= X_{0,0} + \lambda(X_{k-1,0} - X_{0,0}),
\end{aligned}
\end{cases}
\end{equation}
where $\eta$ and $\lambda$ are parameters controlling the influence of the guidance terms, $v_t\left(X_t \mid X_{\text {dual }}\right)$ is intended to encapsulate the dual-objective influence.

\newcommand{\imgbox}[1]{%
  \begin{tcolorbox}[
    colframe=Maroon,             
    boxrule=1.3pt,              
    boxsep=0pt,               
    left=0pt, right=0pt, top=0pt, bottom=0pt, 
    sharp corners,             
    before skip=0pt,          
    after skip=0pt,           
  ]
    \includegraphics[width=\linewidth]{#1} 
  \end{tcolorbox}%
}

\begin{figure*}[htbp]
    \centering
    \captionsetup[subfigure]{labelformat=empty, justification=centering, singlelinecheck=false}
    \setlength{\tabcolsep}{2.5pt}
    \begin{tabular}{ccccccccccc}
    \begin{subfigure}[t]{0.08\textwidth}
            \includegraphics[width=\linewidth]{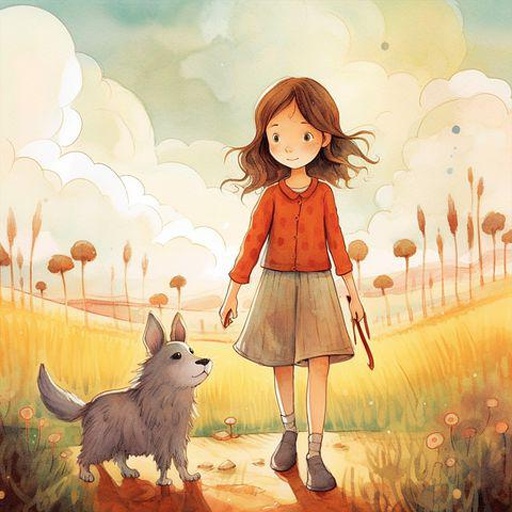}
            \caption{source}
            \label{fig:sub1i}
        \end{subfigure} &
        \begin{subfigure}[t]{0.08\textwidth}
            \includegraphics[width=\linewidth]{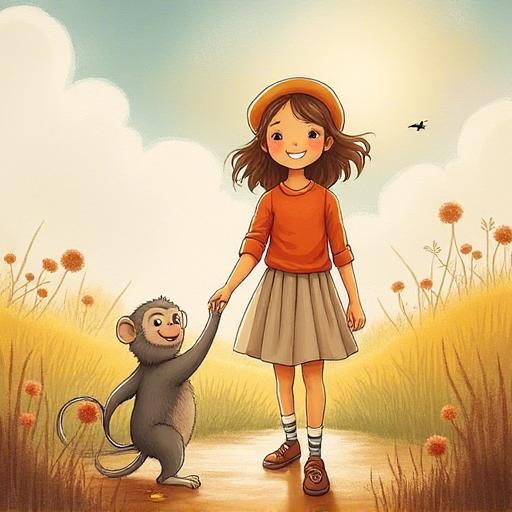}
            \caption{\sout{dog} $\rightarrow$ \\ \textcolor{BrickRed}{\textit{monkey}}}
            \label{fig:sub1i}
        \end{subfigure} &
        \begin{subfigure}[t]{0.08\textwidth}
            \includegraphics[width=\linewidth]{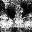}
            \caption{block 1}
            \label{subfig:sub1a}
        \end{subfigure} &
        \begin{subfigure}[t]{0.08\textwidth}
            \includegraphics[width=\linewidth]{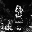}
            \caption{block 2}
            \label{subfig:sub1b}
        \end{subfigure} &
        \begin{subfigure}[t]{0.08\textwidth}
            
        \includegraphics[width=\linewidth]{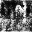}
            \caption{block 3}
            \label{subfig:sub1c}
        \end{subfigure} &
        \begin{subfigure}[t]{0.08\textwidth}
        \imgbox
            {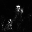}
            \caption{block 6}
            \label{subfig:sub1d}
        \end{subfigure} &
        \begin{subfigure}[t]{0.08\textwidth}
            \includegraphics[width=\linewidth]{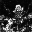}
            \caption{block 8}
            \label{subfig:sub1e}
        \end{subfigure} &
        \begin{subfigure}[t]{0.08\textwidth}
            \includegraphics[width=\linewidth]{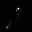}
            \caption{block 9}
            \label{fig:sub1f}
        \end{subfigure} &
        \begin{subfigure}[t]{0.08\textwidth}
            \includegraphics[width=\linewidth]{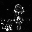}
            \caption{block 12}
            \label{fig:sub1g}
        \end{subfigure} &
        \begin{subfigure}[t]{0.08\textwidth}
        \setlength{\fboxrule}{1pt}
        \imgbox
            {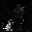}
            \caption{block 16}
            \label{fig:sub1h}
        \end{subfigure} &
        \begin{subfigure}[t]{0.08\textwidth}
        \imgbox
       {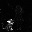}
            \caption{block 18}
            \label{fig:sub1i}
        \end{subfigure}
        \\
        
        \begin{subfigure}[t]{0.08\textwidth}
            \includegraphics[width=\linewidth]{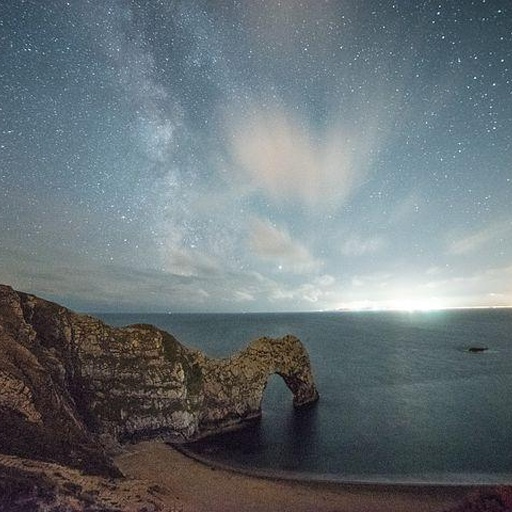}
            \caption{source}
            \label{fig:sub2a}
        \end{subfigure} &
        \begin{subfigure}[t]{0.08\textwidth}
            \includegraphics[width=\linewidth]{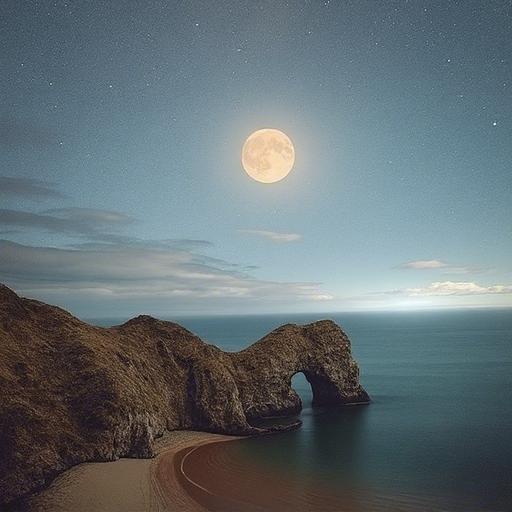}
            \caption{+ \textcolor{BrickRed}{\textit{moon}}}
            \label{fig:sub2b}
        \end{subfigure} &
        \begin{subfigure}[t]{0.08\textwidth}
            \includegraphics[width=\linewidth]{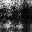}
            \caption{block 1}
            \label{fig:sub2c}
        \end{subfigure} &
        \begin{subfigure}[t]{0.08\textwidth}
        
            \includegraphics[width=\linewidth]{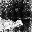}
            \caption{block 2}
            \label{fig:sub2d}
        \end{subfigure} &
        \begin{subfigure}[t]{0.08\textwidth}
        \imgbox
        {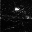}
            \caption{block 5}
            \label{fig:sub2e}
        \end{subfigure} &
        \begin{subfigure}[t]{0.08\textwidth}
            \includegraphics[width=\linewidth]{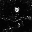}
            \caption{block 6}
            \label{fig:sub2f}
        \end{subfigure} &
        
        \begin{subfigure}[t]{0.08\textwidth}
       \includegraphics[width=\linewidth]{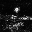}
    \caption{block 8}
    \label{fig:sub2g}
\end{subfigure} &
        
        \begin{subfigure}[t]{0.08\textwidth}
        \imgbox
        {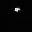}
            \caption{block 10}
            \label{fig:sub2h}
        \end{subfigure} &
        \begin{subfigure}[t]{0.08\textwidth}
        \imgbox
        {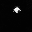}
            \caption{block 11}
            \label{fig:sub2i}
        \end{subfigure}&
        \begin{subfigure}[t]{0.08\textwidth}
            \includegraphics[width=\linewidth]{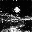}
            \caption{block 16}
            \label{fig:sub1i}
        \end{subfigure}&
        \begin{subfigure}[t]{0.08\textwidth}
        \setlength{\fboxrule}{1pt}
            \setlength{\fboxsep}{0pt}

           \imgbox
           {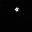}
            
            \caption{block 18}
            \label{fig:sub1i}
        \end{subfigure}
    \end{tabular}
    
    \caption{\textbf{Self-attention map visualizations} from selected FLUX double blocks (19 total) illustrate layer-specific roles in the editing process (e.g., global, local, details). Top row: attention maps corresponding to the ``monkey'' text token. Bottom row: maps for the ``moon'' token. The attention map highlighted by a red box denotes correctly activated maps.}
    \label{fig:mask_observe}
\end{figure*}

\subsection{Adaptive Attention Guidance}
\label{sec:attention guidance}
Our framework leverages flow reversal and LQR-based optimal control for distributional consistency across iterative edits. 
While LQR ensures stability—critical for multi-turn editing—its strong regularization can limit editability. To balance stability and flexibility, we introduce adaptive attention modulation, guiding edits towards salient regions for precise, localized modifications while preserving unaffected areas.

Unlike Stable Diffusion (SD1-5~\cite{rombach2022high}, SD2-1, SDXL~\cite{podell2023sdxl}), which processes image and text information through cross-attention~\cite{hertz2022prompt, kumari2023multi, cao2023masactrl}, FLUX utilizes double blocks to jointly process text and image embeddings.
Following the observation by Xu et al. \cite{xu2024headrouter} that FLUX's lower-left self-attention quadrant encodes text-to-image spatial influence.
With each column representing a text token's modulation, we exploit this column-wise interaction for fine-grained analysis and to implement an adaptive attention control strategy.

As shown in~\cref{fig:mask_observe}, which illustrates a token mapping column reshaped into a visualization attention map, different FLUX double blocks exhibit distinct editing behaviors.
As shown in the top row, the first and third double blocks primarily influence the entire image, while the second and twelfth focus on the main object. Notably, the sixteenth and eighteenth blocks precisely activate the region corresponding to ``monkey," aligning with the desired editing area. 
This analysis reveals a discernible trend: highly activated maps tend to perform global editing, while lower activated maps focus on finer details.

Given that maintaining coherence across multiple editing turns is essential for effective multi-turn image editing, we emphasize the importance of performing finer and more localized edits in each turn. 
To achieve this, we propose adaptively identifying and using medium-to-low activated maps as guidance in our framework.
This process generates a mask that highlights the focus area for editing, reducing the impact on unaffected regions and facilitating localized, controlled edits.

We employ the attention map at time-step $k$ and block $l$, defined as:
\begin{equation}
s_{k,l}=\operatorname{softmax}\left(\frac{Q \cdot K^T}{\sqrt{d}}\right).
\end{equation}
Following prior work~\cite{epstein2023diffusion, dalva2024fluxspace}, we rescale the attention values to the interval $[0,1]$ via:
\begin{equation}
s_{k,l}^{\prime}=\sigma\left(10 *\left(\operatorname{normalize}\left(s_{k,l}\right)-0.5\right)\right),
\end{equation}
where $\sigma(\cdot)$ is the sigmoid function and $\operatorname{normalize(\cdot)}$ applies min-max normalization.
Let $S_k^{\prime}=\{s_{k,1}^{\prime}, s_{k,2}^{\prime}, \ldots, s_{k,19}^{\prime}\}$ represent the set of $19$ rescaled self-attention maps at step $k$. 
To adaptively select \textbf{medium-low activated maps} for editing guidance,
we define an activation magnitude function $activation(s_{k,l})$, where $a_{k,l}=activation(s_{k,l})=\sum s_{k,l}$ represents the sum of all elements in the attention map $s_{k,l}$. 
Next, we arrange the tensors in $S_k^{\prime}$ in ascending order-based on their activation levels.
This sorted sequence is denoted as:
\begin{equation}
\begin{aligned}
A_k &= \textit{Sort} \left\{a_{k,1}, a_{k,2}, \ldots, a_{k,19}\right\} = \left\{a_{k,1}^{\prime}, a_{k,2}^{\prime}, \ldots, a_{k,19}^{\prime}\right\}, \\
&\quad \text{where } a_{k,1}^{\prime} \leq \ldots \leq a_{k,19}^{\prime}.
\end{aligned}
\end{equation}
Let $ A_{i:j}= \{a_{k,l}^{\prime} \mid l \in \mathbb{Z}, i \leq l \leq j \}$ denote the subset of maps indexed from $i$ to $j$ $(1 \leq i<j \leq 19)$, corresponding to medium-low activation levels.
The mask $M_k$  is generated by averaging these selected maps:
\begin{equation}
\label{equ:dynamic_attn}
\bar{v}_{i: j}=\frac{1}{j-i+1} \sum_{l=i}^j a_{k,l}^{\prime},
\end{equation}
and thresholding the result to amplify focused regions while suppressing others:
\begin{equation}
\label{equ:soft_thre}
M_k= \begin{cases}h_{\text {factor }} & \text { if } \bar{v}_{i: j} \geq \tau \\ r_{\text {factor }} & \text { otherwise }\end{cases}
\end{equation}
where $h_{factor}$ and $r_{factor}$ control amplification/reduction, and $\tau$ is a predefined threshold. Finally, $M_k$ modulates the attention computation at step $k+1$:
\begin{equation}
s_{k+1,l}=\operatorname{softmax}\left(\frac{Q \cdot K^{\top}}{\sqrt{d}}\right) \odot M_k,
\end{equation}
where $\odot$ denotes element-wise multiplication.

\newcommand{\downarrowwithtext}[2][OliveGreen, very thick]{%
    \begin{tikzpicture}[baseline=(current bounding box.center)]
        \draw[->, line width=1pt, draw=#1, >=Latex] (0, 0) -- (0, -2); 
        \node[below=2pt, align=center, text=#1] at (0, -2) {#2}; 
    \end{tikzpicture}%
}

\begin{figure*}[htbp]
    \centering
    \setlength{\tabcolsep}{2pt} 
    \renewcommand{\arraystretch}{0.8} 

    \begin{minipage}[t]{0.09\textwidth}
        \centering
        \vspace{-3.4cm}
        \includegraphics[width=\linewidth]{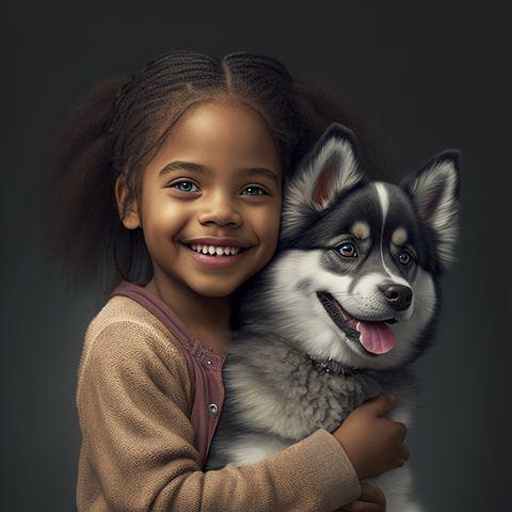}
        \vspace{-0.8cm}
        \begin{center}
        source
    \end{center}
        \vspace{0.3cm} 
        \scalebox{0.9}{\downarrowwithtext[OliveGreen, ultra thick]{\textit{Editing} \\ \textit{Turn}}}
        
    \end{minipage}%
    \hspace{2em}
    \begin{minipage}[t]{0.84\textwidth}
        \begin{tabular}{p{0.3cm}cccccccc}
            &Ours & RF-Inv. & StableFlow & FlowEdit & RF-Solver & FireFlow &  MasaCtrl & PnPInv. \\
            \scalebox{0.9}{\raisebox{0.5cm}{\rotatebox{90}{\textcolor{BrickRed}{\textit{aged}}}}}&
            \includegraphics[width=0.11\textwidth]{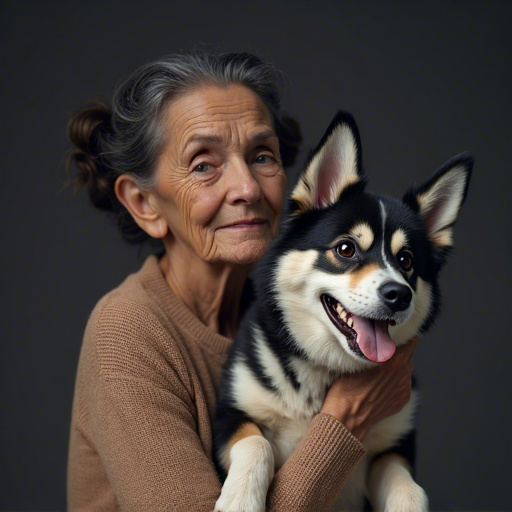} &
            \includegraphics[width=0.11\textwidth]{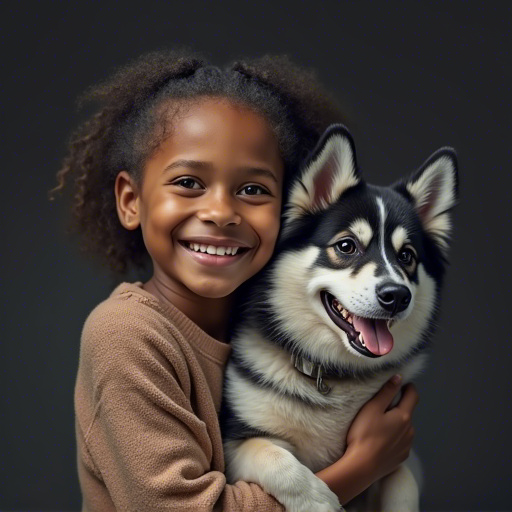} &
            \includegraphics[width=0.11\textwidth]{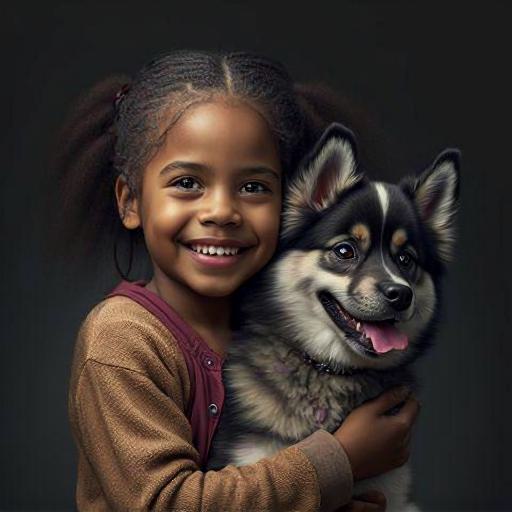} &
            \includegraphics[width=0.11\textwidth]{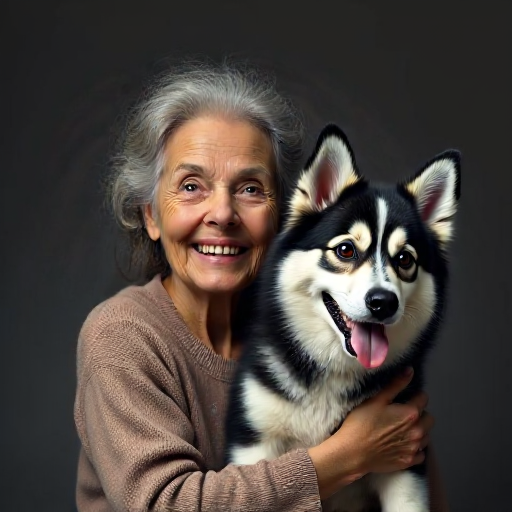} &
            \includegraphics[width=0.11\textwidth]{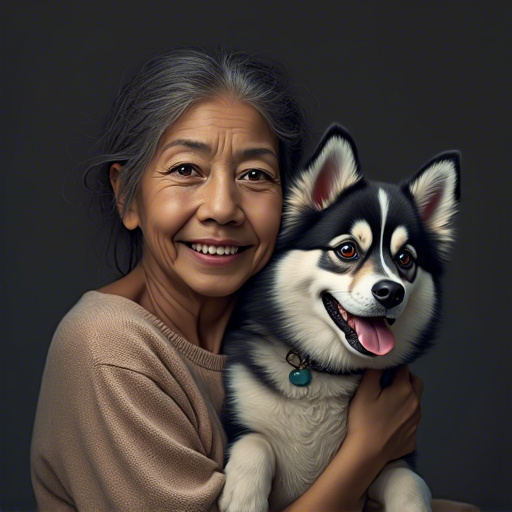} &
            \includegraphics[width=0.11\textwidth]{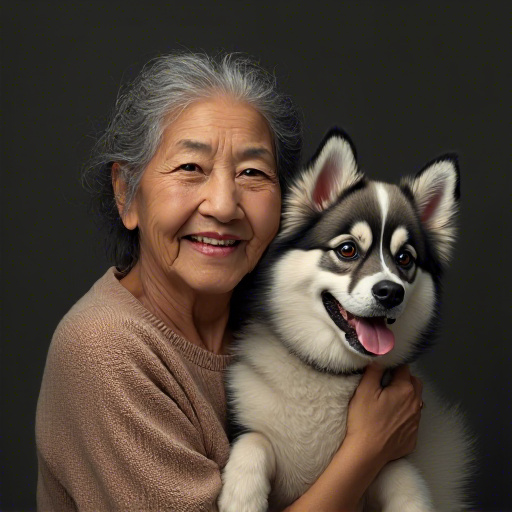} &
            \includegraphics[width=0.11\textwidth]{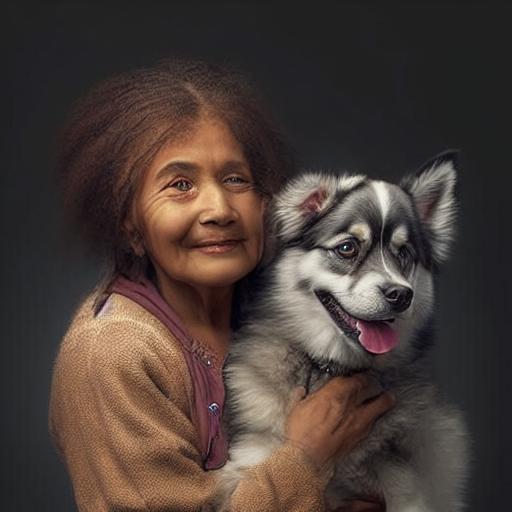} &
            \includegraphics[width=0.11\textwidth]{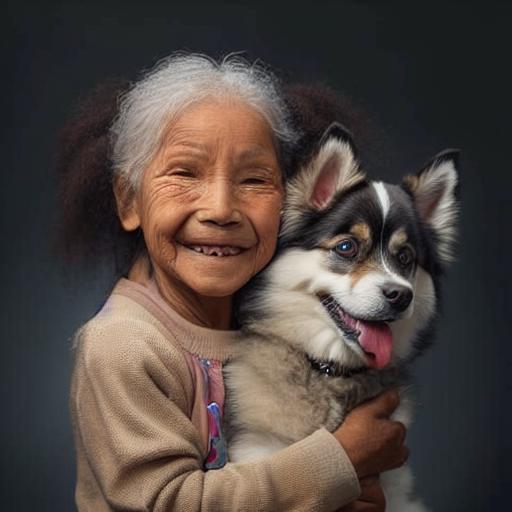} \\

            \scalebox{0.9}{\raisebox{0.2cm}{\rotatebox{90}{\sout{dog} $\rightarrow$ \textcolor{BrickRed}{\textit{cat}}}}}&
            \includegraphics[width=0.11\textwidth]{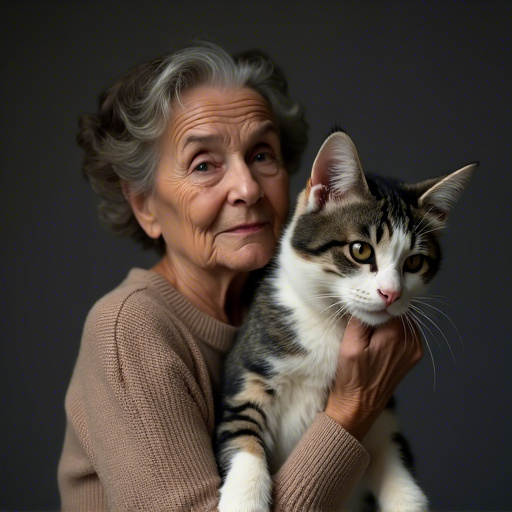} &
            \includegraphics[width=0.11\textwidth]{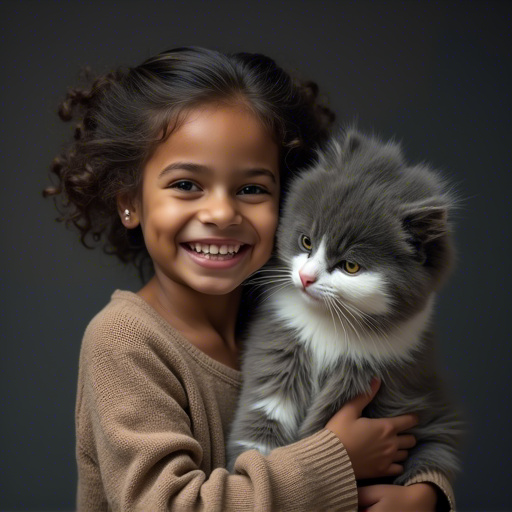} &
            \includegraphics[width=0.11\textwidth]{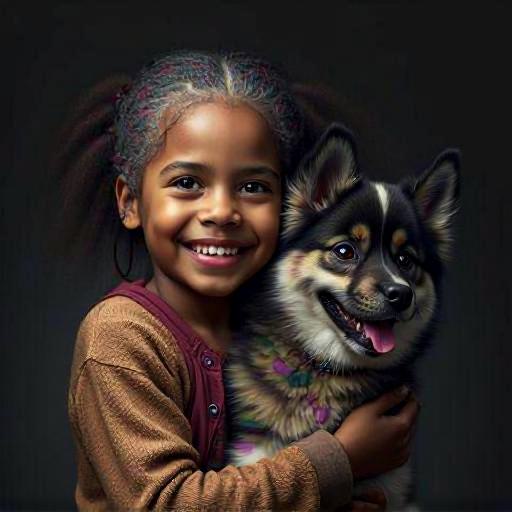} &
            \includegraphics[width=0.11\textwidth]{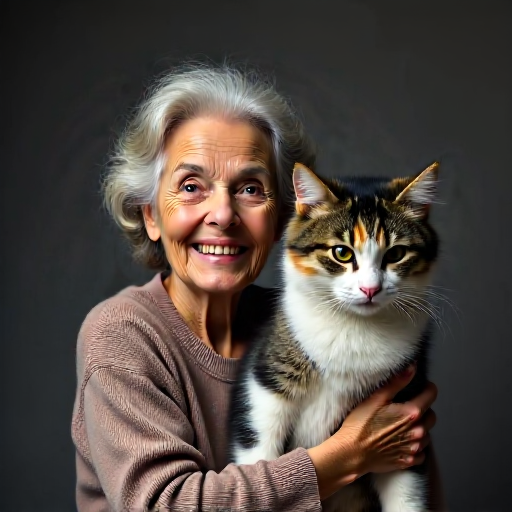} &
            \includegraphics[width=0.11\textwidth]{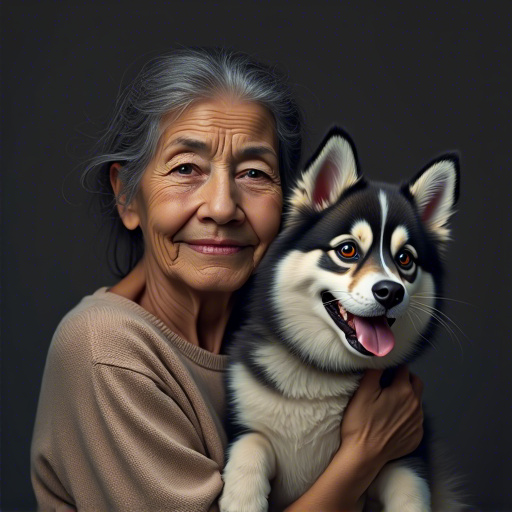} &
            \includegraphics[width=0.11\textwidth]{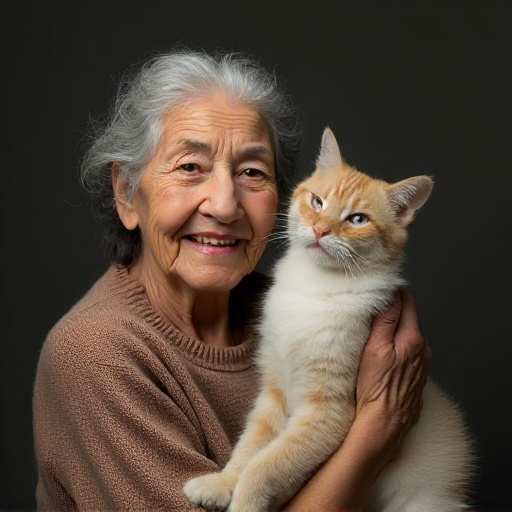} &
            \includegraphics[width=0.11\textwidth]{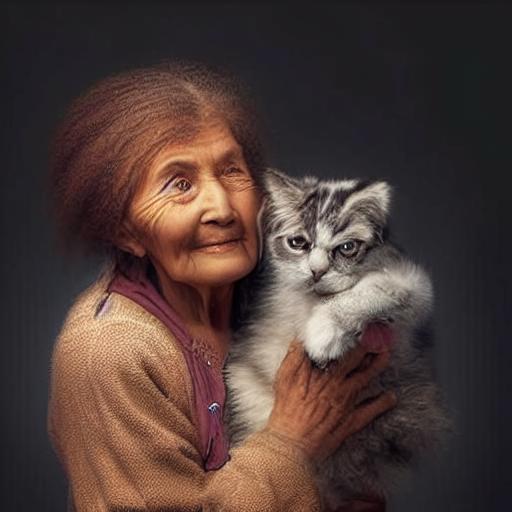} &
            \includegraphics[width=0.11\textwidth]{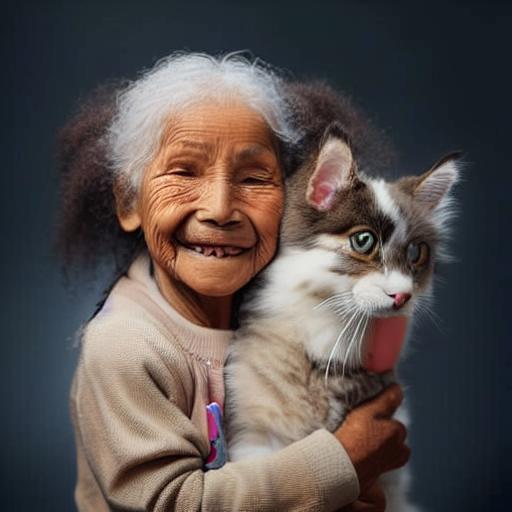} \\
            
            \scalebox{0.9}{\raisebox{0cm}{\rotatebox{90}{+ \textcolor{BrickRed}{\textit{blue collar}}}}}&
            \includegraphics[width=0.11\textwidth]{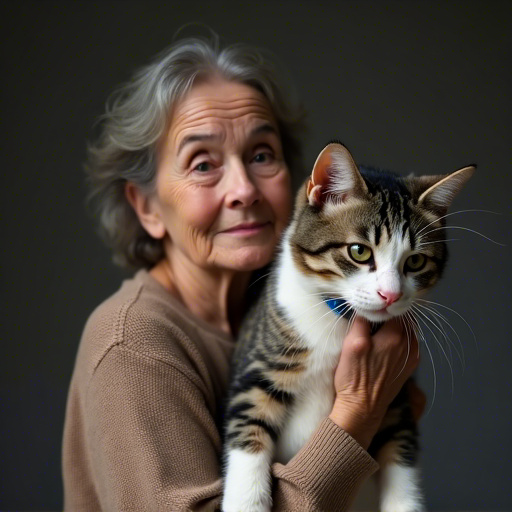} &
            \includegraphics[width=0.11\textwidth]{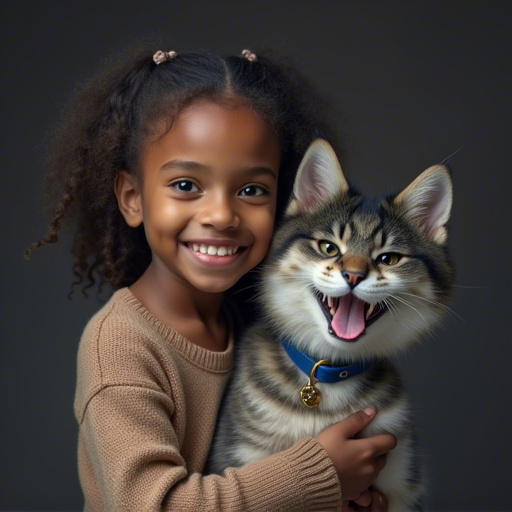} &
            \includegraphics[width=0.11\textwidth]{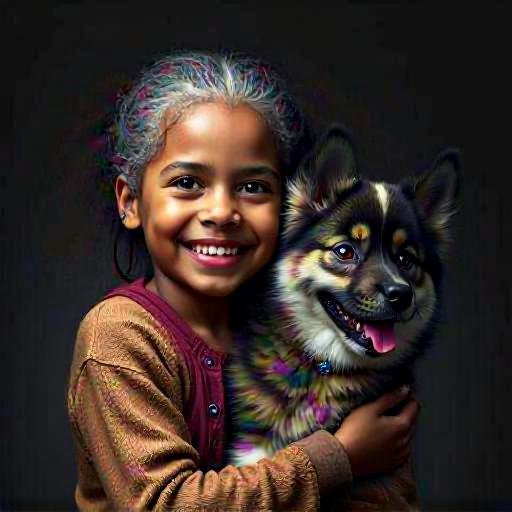} &
            \includegraphics[width=0.11\textwidth]{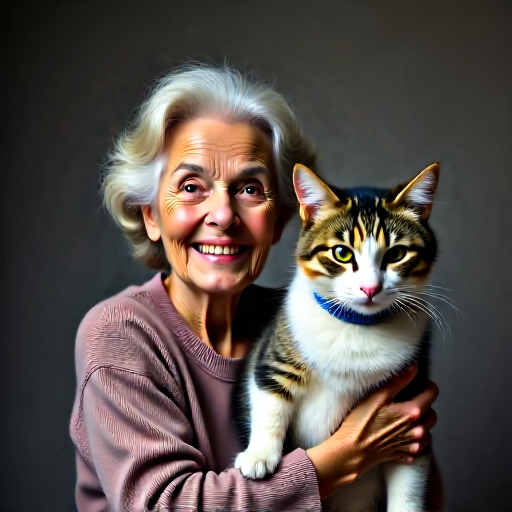} &
            \includegraphics[width=0.11\textwidth]{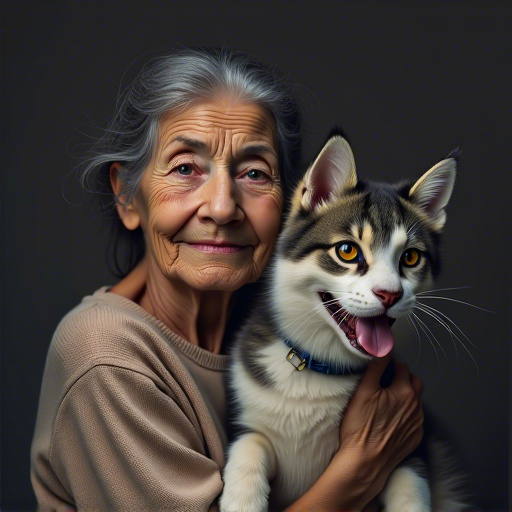} &
            \includegraphics[width=0.11\textwidth]{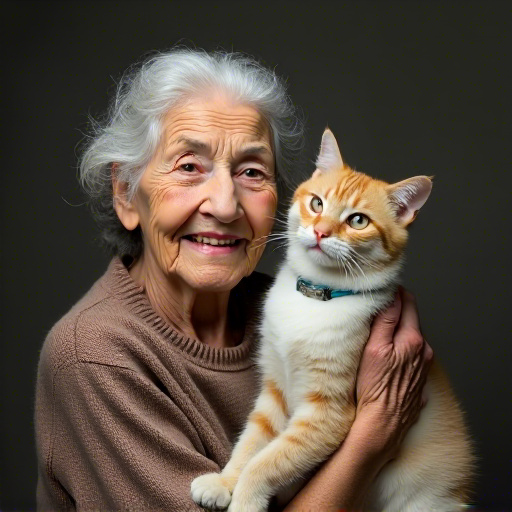} &
            \includegraphics[width=0.11\textwidth]{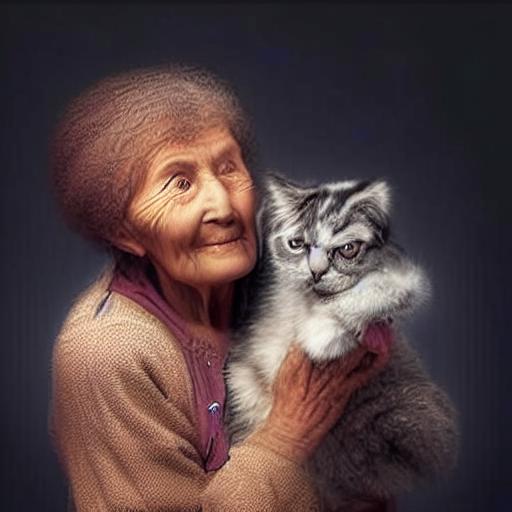} &
            \includegraphics[width=0.11\textwidth]{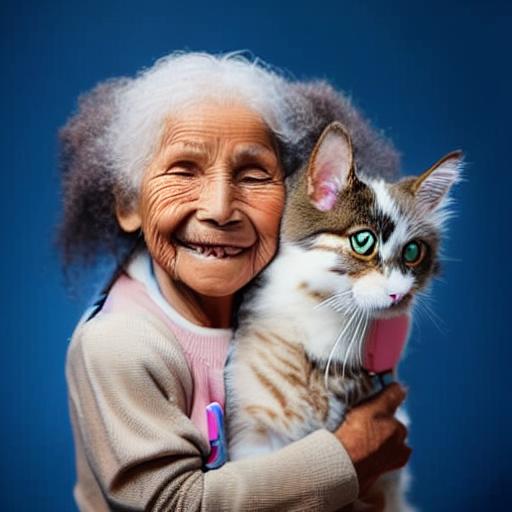} \\
            
            \scalebox{0.9}{\raisebox{0.1cm}{\rotatebox{90}{\textcolor{BrickRed}{\textit{[red]}} dress}}}&
           \includegraphics[width=0.11\textwidth]{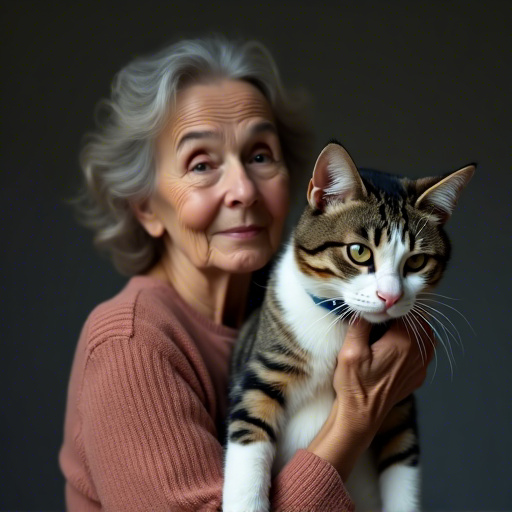} &
            \includegraphics[width=0.11\textwidth]{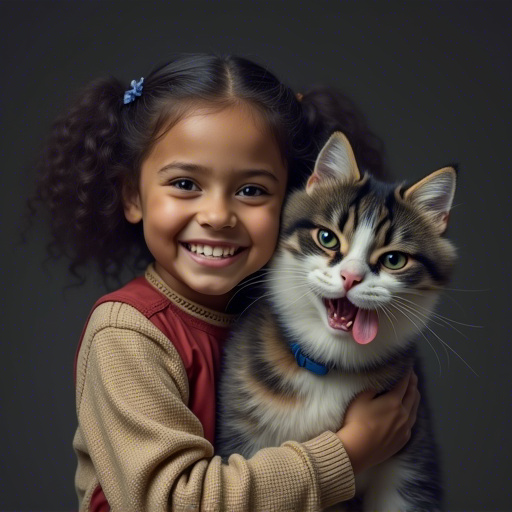} &
            \includegraphics[width=0.11\textwidth]{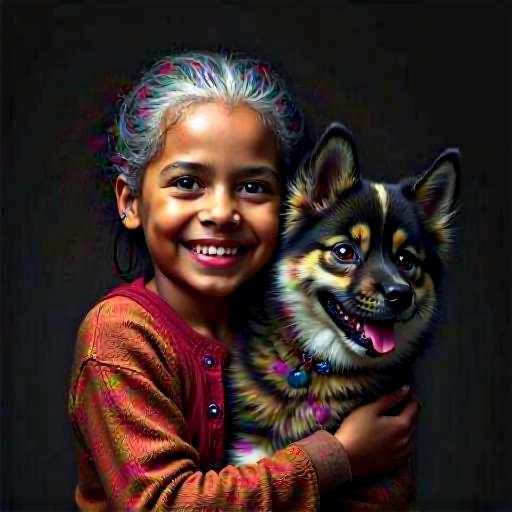} &
            \includegraphics[width=0.11\textwidth]{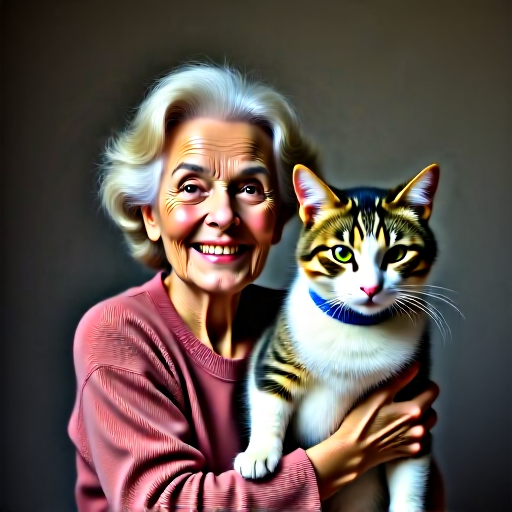} &
            \includegraphics[width=0.11\textwidth]{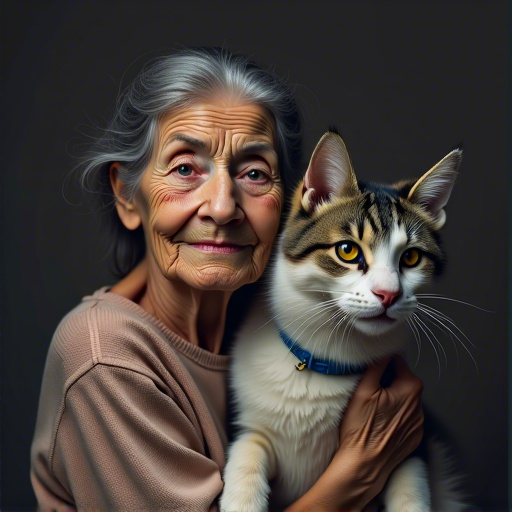} &
            \includegraphics[width=0.11\textwidth]{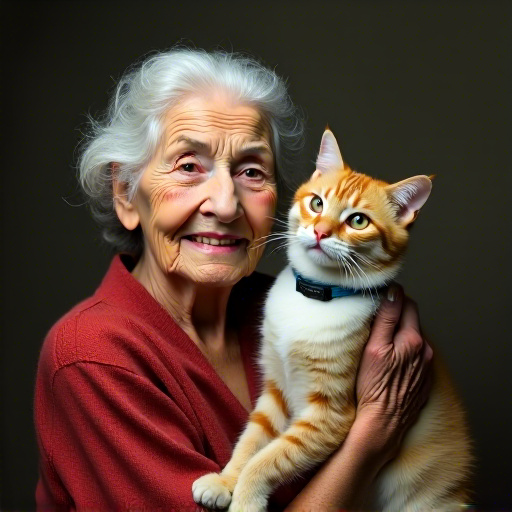} &
            \includegraphics[width=0.11\textwidth]{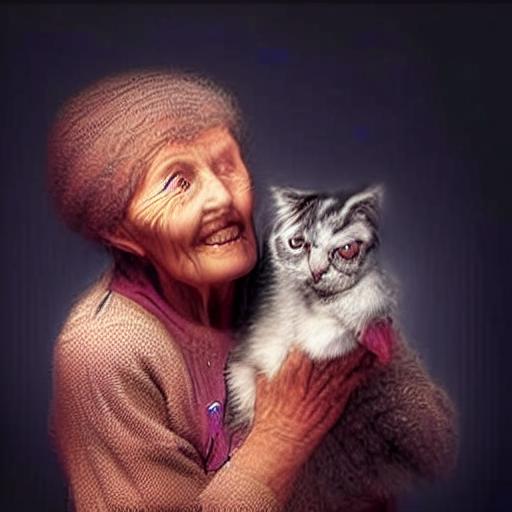} &
            \includegraphics[width=0.11\textwidth]{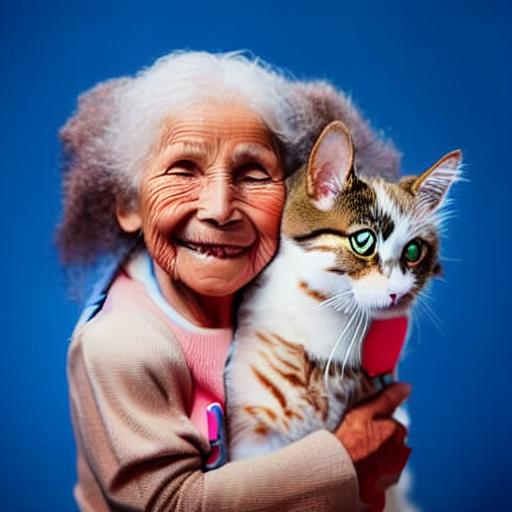}
        \end{tabular}
    \end{minipage}

\vspace{0.7cm}

    \begin{minipage}[t]{0.09\textwidth}
        \centering
        \vspace{-3.4cm}
        \includegraphics[width=\linewidth]{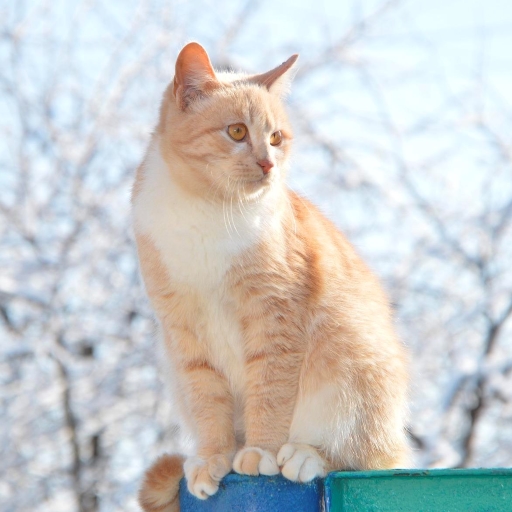}
      \vspace{-0.8cm} \begin{center}
        source
    \end{center}
        
        \vspace{0.3cm} 
        \scalebox{0.9}{\downarrowwithtext[OliveGreen, ultra thick]{\textit{Editing} \\ \textit{Turn}}}

    \end{minipage}%
    \hspace{2em}
    \begin{minipage}[t]{0.84\textwidth}
        \begin{tabular}{p{0.3cm}cccccccc}
            &Ours & RF-Inv. & StableFlow & FlowEdit & RF-Solver & FireFlow &  MasaCtrl & PnPInv. \\
            \scalebox{0.9}{\raisebox{0.5cm}{\rotatebox{90}{+~\textcolor{BrickRed}{\textit{hat}}}}}&
            \includegraphics[width=0.11\textwidth]{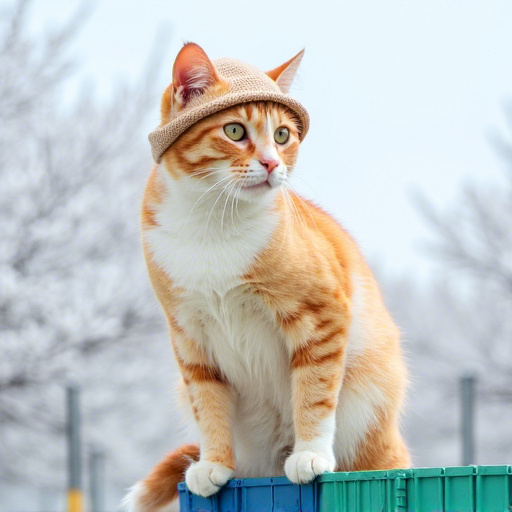} &
            \includegraphics[width=0.11\textwidth]{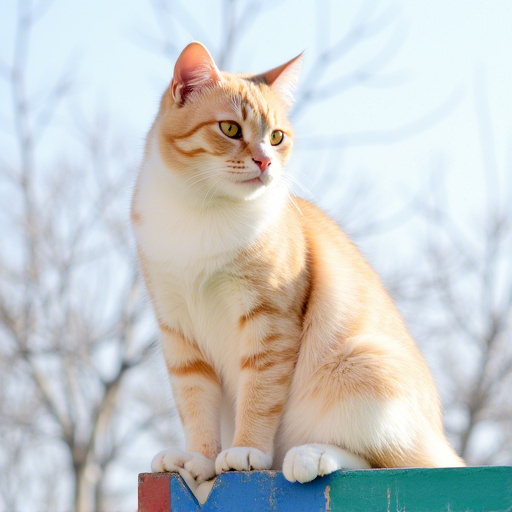} &
            \includegraphics[width=0.11\textwidth]{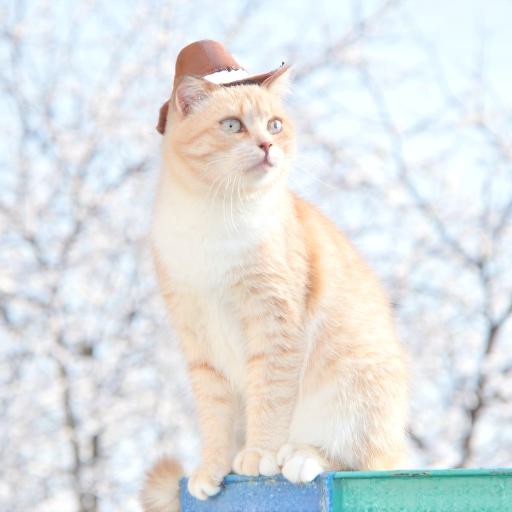} &
            \includegraphics[width=0.11\textwidth]{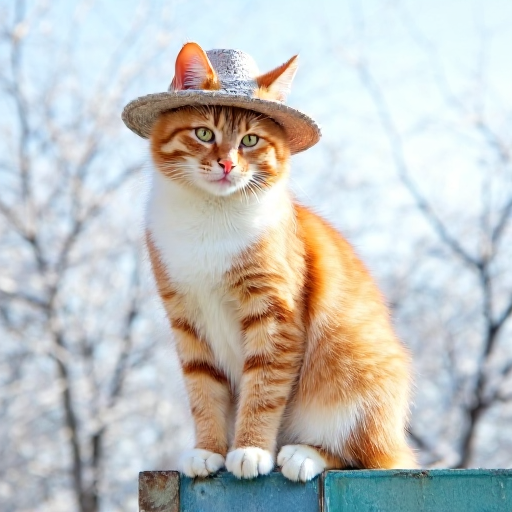} &
            \includegraphics[width=0.11\textwidth]{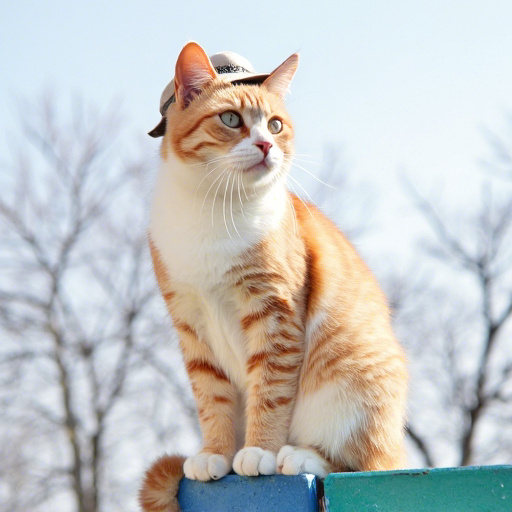} &
            \includegraphics[width=0.11\textwidth]{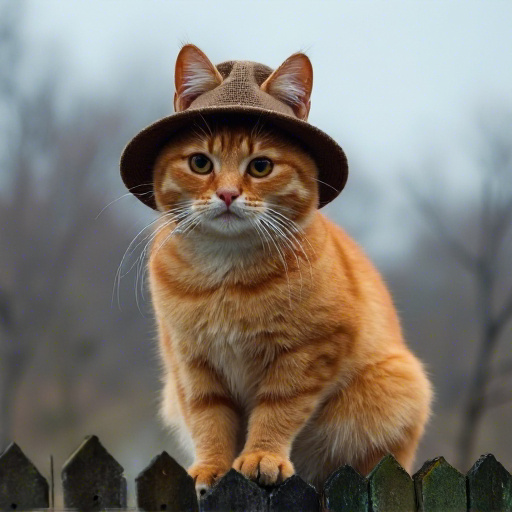} &
            \includegraphics[width=0.11\textwidth]{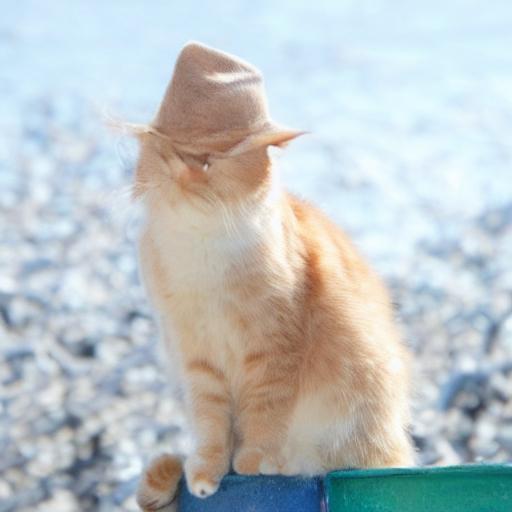} &
            \includegraphics[width=0.11\textwidth]{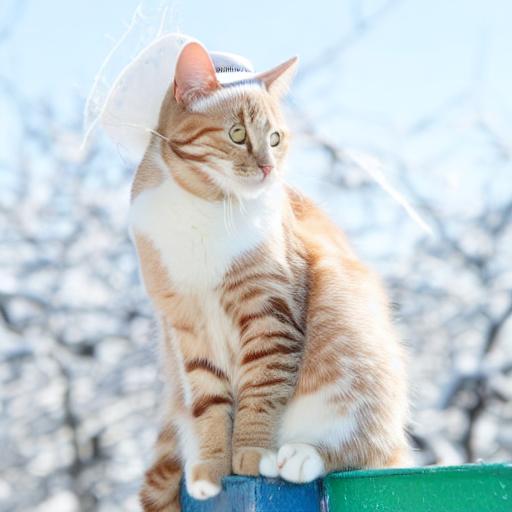} \\

            \scalebox{0.9}{\raisebox{0.2cm}{\rotatebox{90}{\textcolor{BrickRed}{\textit{[blue]}} hat}}}&
            \includegraphics[width=0.11\textwidth]{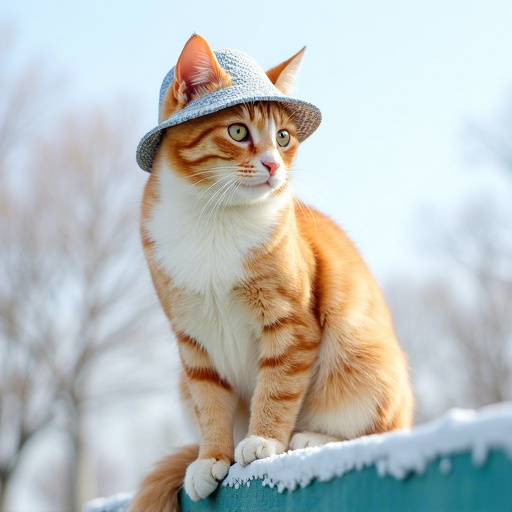} &
            \includegraphics[width=0.11\textwidth]{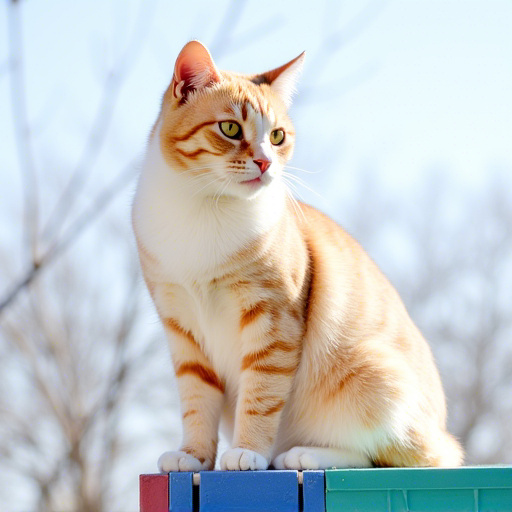} &
            \includegraphics[width=0.11\textwidth]{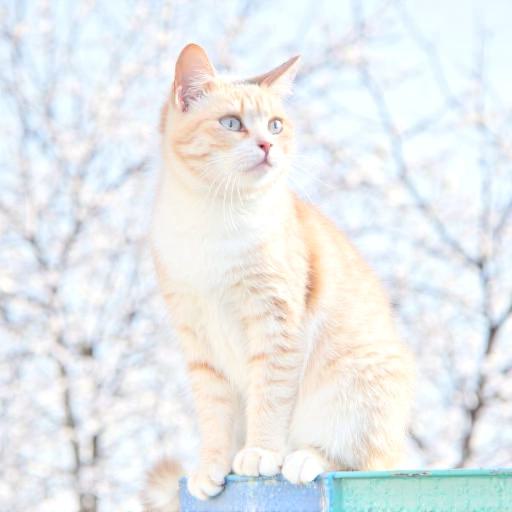} &
            \includegraphics[width=0.11\textwidth]{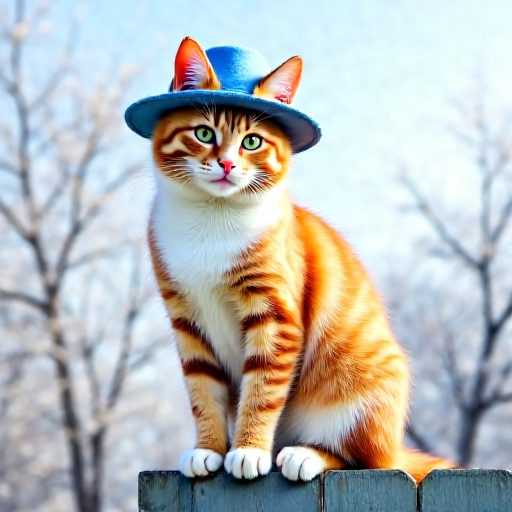} &
            \includegraphics[width=0.11\textwidth]{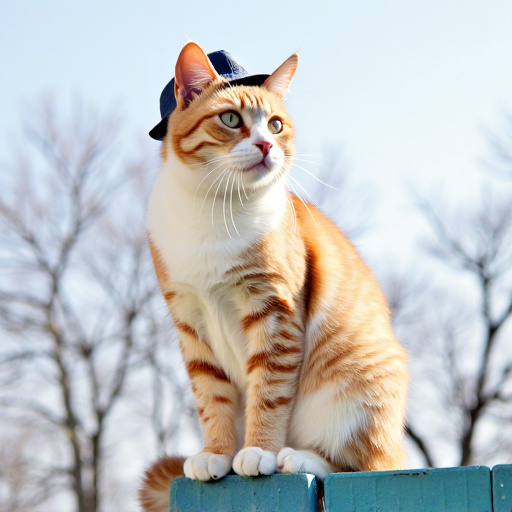} &
            \includegraphics[width=0.11\textwidth]{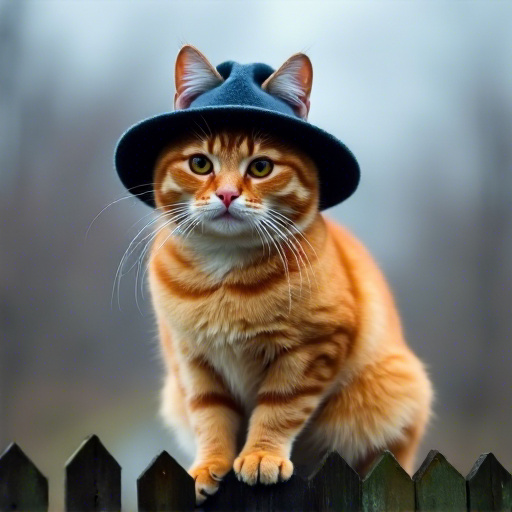} &
            \includegraphics[width=0.11\textwidth]{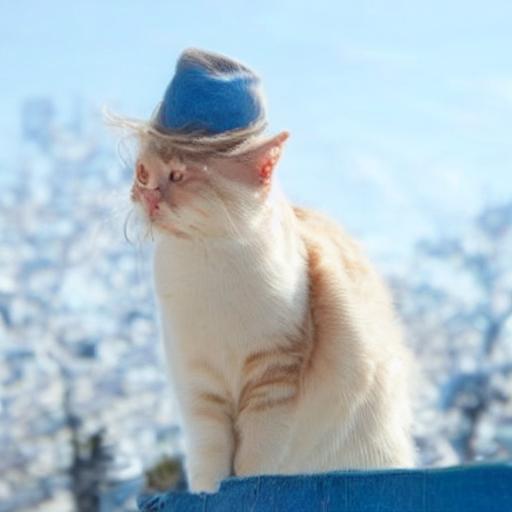} &
            \includegraphics[width=0.11\textwidth]{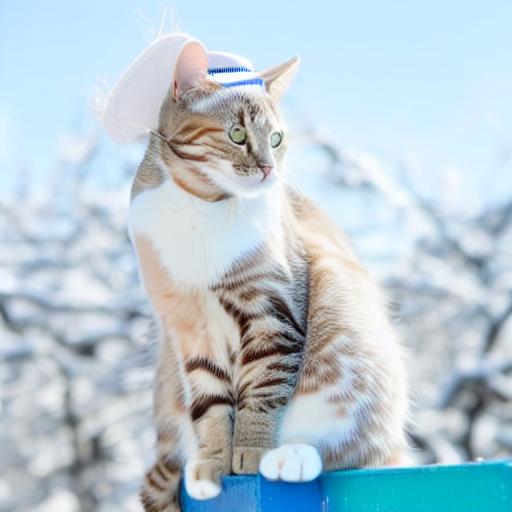} \\
            
            \scalebox{0.9}{\raisebox{0cm}{\rotatebox{90}{+ \textcolor{BrickRed}{\textit{sunglasses}}}}}&
            \includegraphics[width=0.11\textwidth]{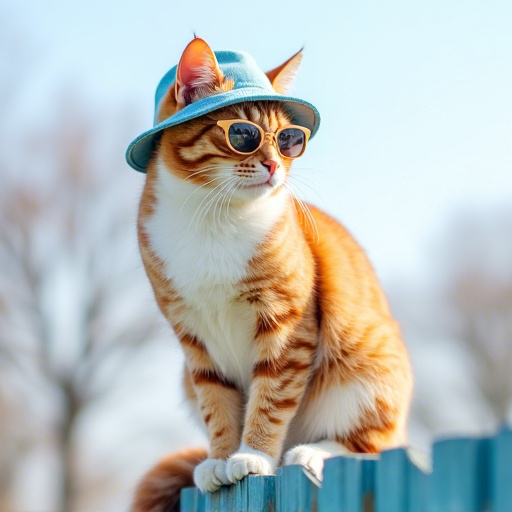} &
            \includegraphics[width=0.11\textwidth]{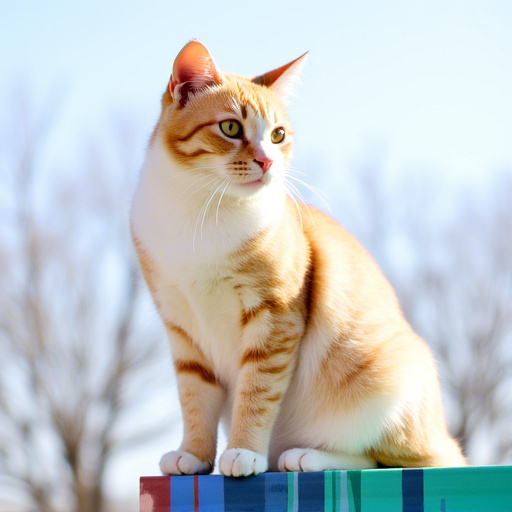} &
            \includegraphics[width=0.11\textwidth]{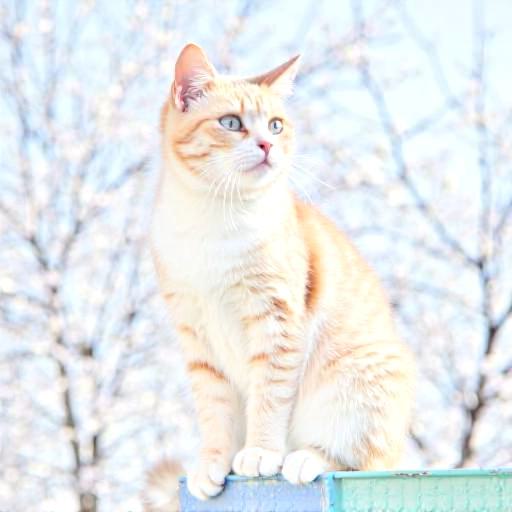} &
            \includegraphics[width=0.11\textwidth]{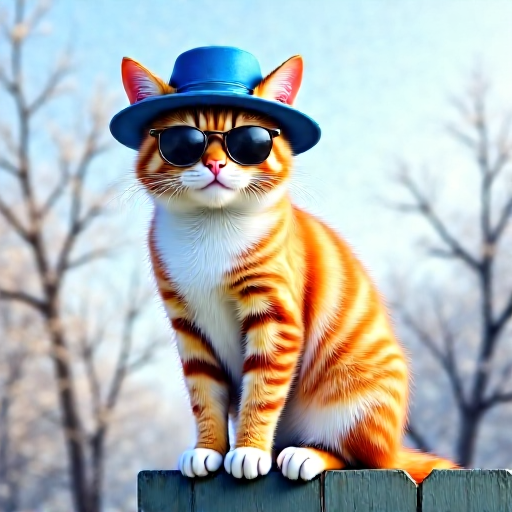} &
            \includegraphics[width=0.11\textwidth]{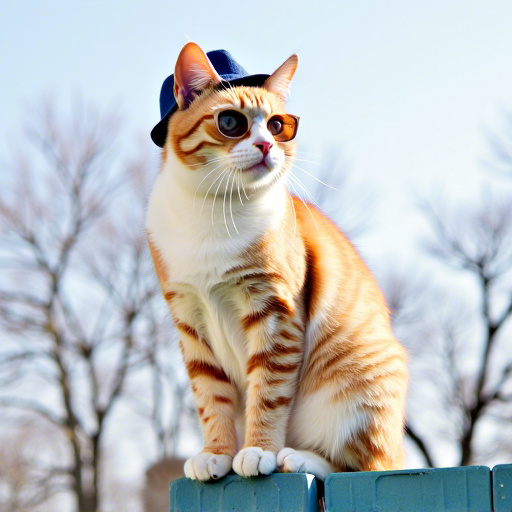} &
            \includegraphics[width=0.11\textwidth]{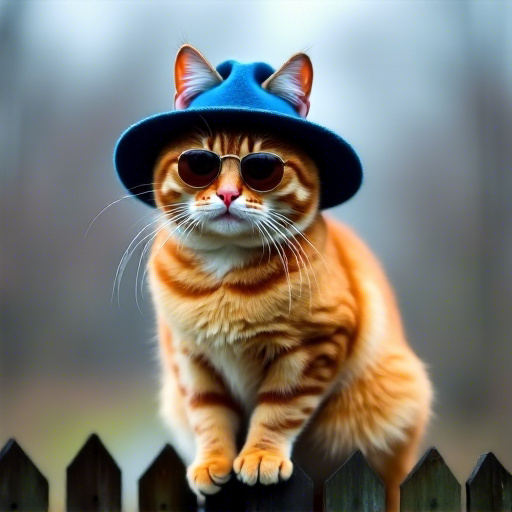} &
            \includegraphics[width=0.11\textwidth]{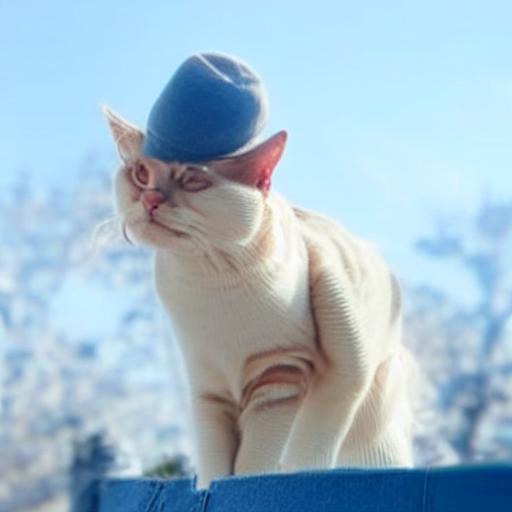} &
            \includegraphics[width=0.11\textwidth]{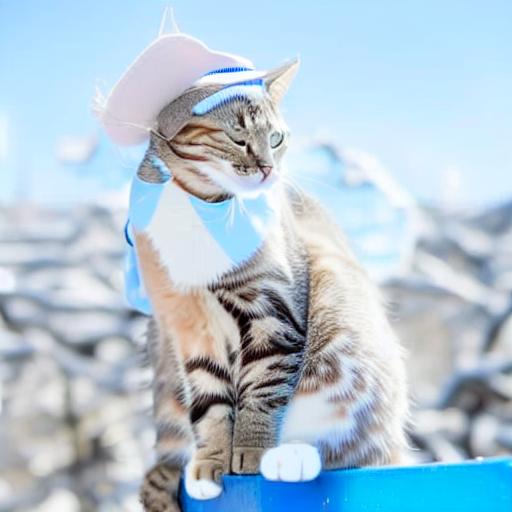} \\
            
            \scalebox{0.9}{\raisebox{0.3cm}{\rotatebox{90}{+~\textcolor{BrickRed}{\textit{scarf}} }}}&
           \includegraphics[width=0.11\textwidth]{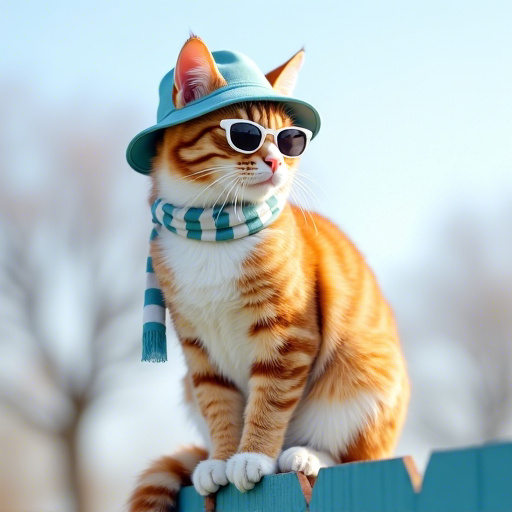} &
            \includegraphics[width=0.11\textwidth]{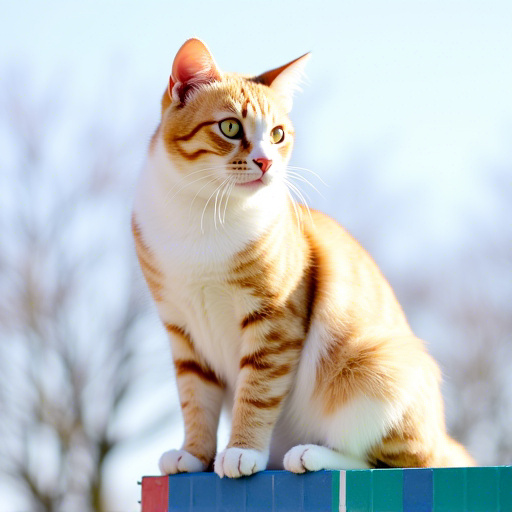} &
            \includegraphics[width=0.11\textwidth]{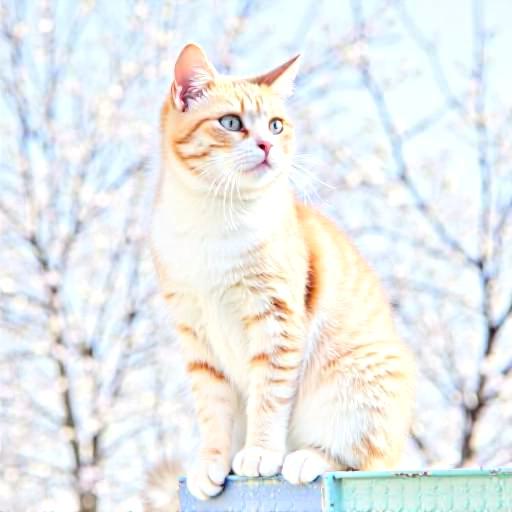} &
            \includegraphics[width=0.11\textwidth]{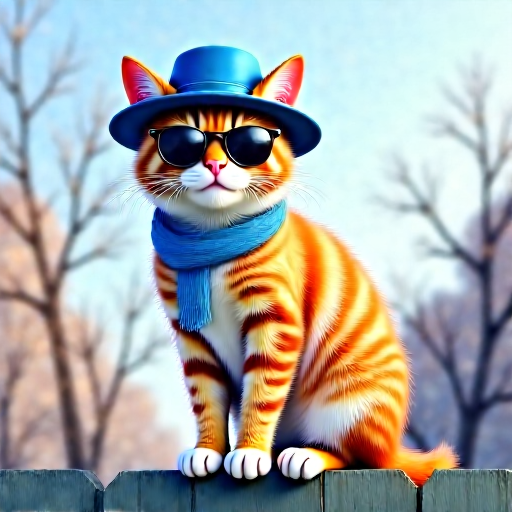} &
            \includegraphics[width=0.11\textwidth]{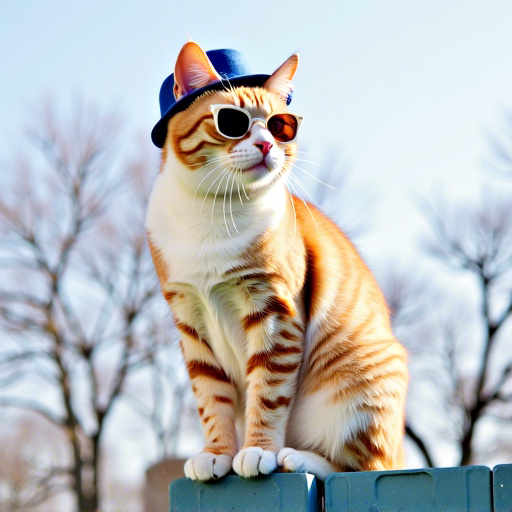} &
            \includegraphics[width=0.11\textwidth]{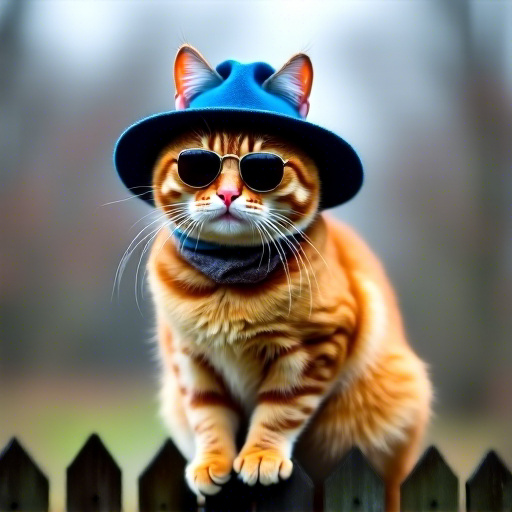} &
            \includegraphics[width=0.11\textwidth]{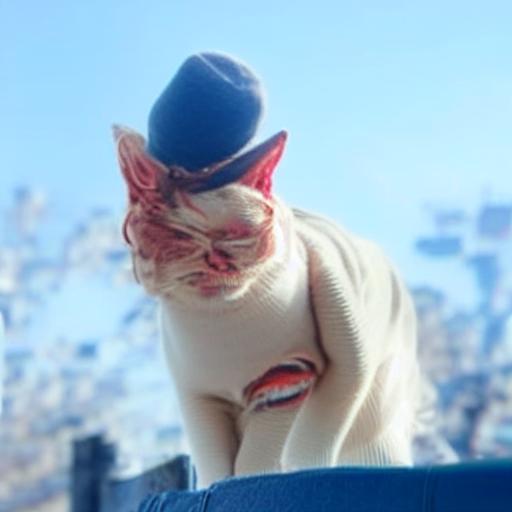} &
            \includegraphics[width=0.11\textwidth]{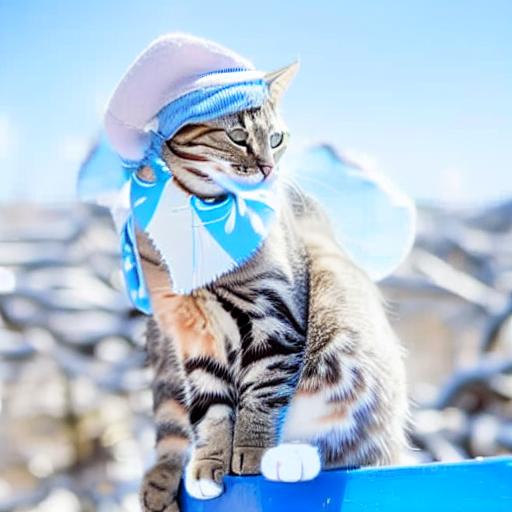}
        \end{tabular}
    \end{minipage}

    \captionsetup{skip=0pt} 
    \caption{\textbf{Qualitative comparison of multi-turn editing results against baseline methods.} Note that our method effectively preserves the original image structure while achieving high-quality edits.}
    \label{fig:multi_turn_editing}

\end{figure*}
\section{Experiment}
\label{sec:experiment}

\subsection{Implementation Details}

\noindent\textbf{Baselines: }
We compare our method against Rectified Flow-based inversion methods including RF-Inverison~\cite{rout2024semantic},
StableFlow~\cite{avrahami2024stable}
RF-Solver~\cite{wang2024taming}, 
FireFlow~\cite{deng2024fireflow} and FlowEdit~\cite{kulikov2024flowedit}.
We also consider the Diffusion inversion-based methods including MasaCtrl~\cite{cao2023masactrl}, and PnP~\cite{tumanyan2023plug}.

\noindent\textbf{Datasets: }
Existing benchmarks do not adequately evaluate multi-turn image editing performance. Therefore, we created a novel dataset based on PIE-Bench \cite{ju2023direct}, a benchmark designed for single-turn image editing. PIE-Bench provides images paired with editing instructions. To extend this resource for multi-turn evaluation, we leveraged GPT-4~Turbo to generate four additional rounds of editing instructions, conditioned on the original prompt and the preceding editing instructions. This extended dataset enables convenient benchmarking of both single-turn and multi-turn image editing tasks.

\noindent\textbf{Metrics: }
To demonstrate our method's balance between content preservation and editability, we employ the following evaluation metrics: CLIP-T\cite{radford2021learning} measures prompt-image consistency; CLIP-I measures the similarity between the original and edited images; and FID \cite{zhang2018unreasonable} assesses the overall generation quality.

\noindent\textbf{Settings: }
Our method was implemented with 15 steps for both inversion and sampling, with parameters $\eta=0.9$ and $\lambda=0.7$ in  ~\cref{equ:double_lqr_sample} for the initial 4 sampling steps, $i=10$ and $j=14$ in ~\cref{equ:dynamic_attn}, $h_{factor}=2.0$ and $r_{factor}=0.8$ in~\cref{equ:soft_thre}.
Baseline methods were implemented using their official code and default settings: StableFlow\cite{avrahami2024stable} (50 steps), FlowEdit\cite{kulikov2024flowedit} (28 steps).
FireFlow~\cite{deng2024fireflow} was evaluated both in its original form and without the attention's V replacement variant (denoted as FireFlow-v).
RF-Solver\cite{wang2024taming} was implemented with 25 steps, accounting for its second-order ODE solver (50 effective steps totally) with V replacement. MasaCtrl\cite{cao2023masactrl} and PnPInversion\cite{tumanyan2023plug} used Stable Diffusion's standard 50-step inversion and sampling.


\subsection{Multi-turn Reconstruction}
\label{sec5_1:reconstruction}
The qualitative results of multi-turn reconstruction can be seen in~\cref{fig:multi_turn_rec}.
Diffusion-based DDIM inversion is not as accurate as flow-based methods, demonstrating that flow matching excels in both speed and accuracy, which shows great potential for multi-turn scene editing or reconstruction.
FireFlow \cite{deng2024fireflow} and RF-Solver \cite{wang2024taming} perform exceptionally well in single-step reconstruction, indicating that solving second-order ODEs in flow matching improves inversion accuracy and reduces reconstruction error.
However, these two methods still suffer from accumulated errors, causing the distribution of the reconstructed image to deviate from the original one.
RF-Inversion\cite{rout2024semantic} maintains semantic consistency and distribution well but tends to enforce certain patterns in the image.
In contrast, our method preserves the distribution and produces natural-looking results even as the number of editing rounds increases.


 \begin{table}[t]
		\centering
        \belowrulesep=0pt
        \aboverulesep=0pt
		\begin{tabular}{l|>{\centering\arraybackslash}p{1cm}>{\centering\arraybackslash}p{1.4cm}>{\centering\arraybackslash}p{1.3cm}>{\centering\arraybackslash}p{0.9cm}}
        \hline
          \textbf{Method} & FID $\downarrow $ & CLIP-T $\uparrow$ & CLIP-I $\uparrow$ & Steps
          \\ \hline
           RF-Inv. &
          5.740 & 24.094 & \underline{0.904} & 28\\
           StableFlow 
           &  20.624 & 24.234 & 0.899 & 50\\ 
            
	        FlowEdit &
           14.547 & 26.703 & 0.894 & 28\\ 
            
		    RF-Solver
            & 11.581 & 25.516 & \textbf{0.906} & 25
            \\ 

            FireFlow &
             7.970 & 26.500 & 0.897 & 8\\ 
             
            FireFlow$-$\textit{v} &
             12.375 & \textbf{28.281} & 0.873 & 8\\ 

            MasaCtrl &
            10.811 & 23.797 & 0.886 & 50\\ 

            PnPInv.
            & 10.262  & 25.765 & 0.872 & 50\\
            
            \rowcolor{gray!20}
            Ours
            & \underline{5.553} & \underline{26.831} & 0.894 & 15 \\  
             \rowcolor{gray!20}
            Ours
            & \textbf{5.396} & 25.828 & 0.902 & 8 \\\hline

		\end{tabular}
          \caption{\textbf{Quantitative results of fourth-turn editing.} The best results are highlighted in bold, while the second-best results are underlined. Our method achieves a balance between CLIP-I and CLIP-T scores while obtaining the best FID score in the fourth editing turn.}
        \label{tab:round_4_editing}
	\end{table}

\begin{table}[t]
		\centering
        \belowrulesep=0pt
        \aboverulesep=0pt
		\begin{tabular}{l|>{\centering\arraybackslash}p{1.1cm}>{\centering\arraybackslash}p{1.4cm}>{\centering\arraybackslash}p{1.3cm}>{\centering\arraybackslash}p{0.9cm}}
        \hline
          \textbf{Method} & FID $\downarrow $ & CLIP-T $\uparrow$ & CLIP-I $\uparrow$ & Steps
          \\ \hline

            \textit{Single-LQR}
            & 9.886 & 26.484 & 0.892 & 15 \\
   
            \textit{High-attn}
            & 6.316 & \underline{26.878} & 0.891 & 15 \\
             
            \textit{w/o attn}
            & 6.678 & 26.760 & 0.889 & 15 \\
             \rowcolor{gray!20}
            Ours
            & \underline{5.553} & 26.831 & \underline{0.894} & 15 \\
            
        \hline
		\end{tabular}
          \caption{\textbf{Ablation study on fourth-turn editing reults.}}
        \label{tab:ablation table}
	\end{table}
\subsection{Multi-turn Editing}
\label{sec5_2:multi-turn editing}
\cref{fig:multi_turn_editing} provides a qualitative comparison of multi-turn editing results, illustrating the performance of our method and several baseline techniques.
In our experiments, Diffusion Model (DM)-based methods, including MasaCtrl \cite{cao2023masactrl} and PnPInversion (Direct Inversion \cite{ju2023direct} for inversion, and PnP \cite{tumanyan2023plug} for sampling)
, performed poorly in multi-turn editing, failing to preserve the original image structure and generate accurate, high-quality edits.
While RF-Inversion \cite{rout2024semantic}, RF-Solver Edit \cite{wang2024taming}, and StableFlow \cite{avrahami2024stable} demonstrate accurate inversion by maintaining the original image structure, they often fail to produce the desired edits. For example, RF-Solver and StableFlow are unable to transform a ``dog" into a ``cat" (top subfigure) or add a ``scarf" (bottom subfigure). FireFlow \cite{deng2024fireflow} and FlowEdit \cite{kulikov2024flowedit} successfully perform the edits specified by the text prompts, but they compromise the original image structure to varying degrees, with FlowEdit exhibiting a tendency to generate images with increasing artifacts over multiple editing rounds. Our method overcomes these limitations by achieving a more adaptable balance between structure preservation and successful editing, allowing for both accurate and meaningful image manipulations.

Table \ref{tab:round_4_editing} presents quantitative results for the fourth editing turn, highlighting our approach's advantages in multi-turn scenarios. Our method achieves a relatively high CLIP-T score, demonstrating successful alignment with the editing prompt, while simultaneously maintaining high CLIP-I scores, indicating effective content preservation. Notably, our method also achieves the best FID score, suggesting that the generated images retain the characteristics of natural images and exhibit minimal distribution bias after multiple editing iterations.


\begin{figure}
    \centering
     \parbox{0.48\textwidth}{\centering hat $\rightarrow$ \textit{\textcolor{BrickRed}{[blue]} hat }}

    \setlength{\tabcolsep}{1pt}
    \begin{tabular}{cccc}
      \includegraphics[width=0.22\linewidth]{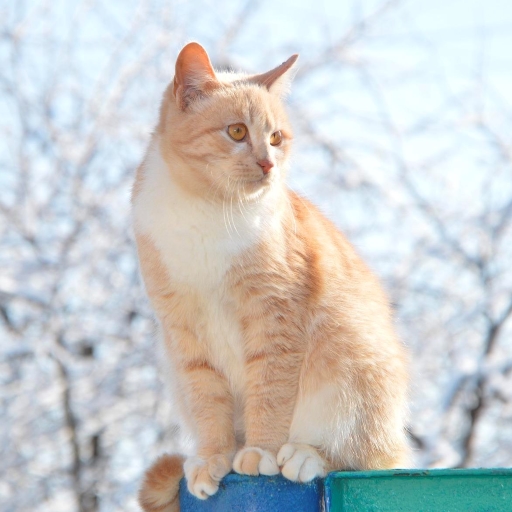}  &
      \includegraphics[width=0.22\linewidth]{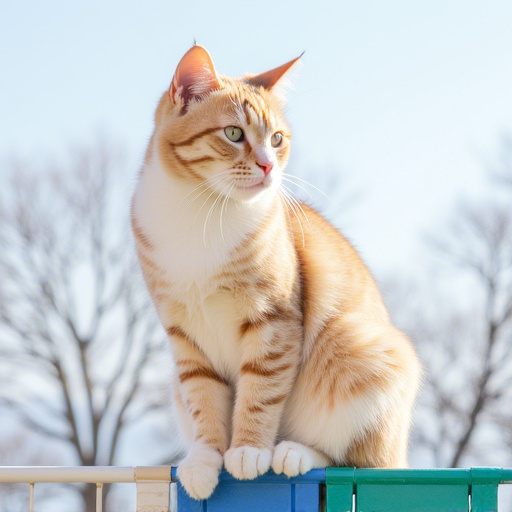}  &
      \includegraphics[width=0.22\linewidth]{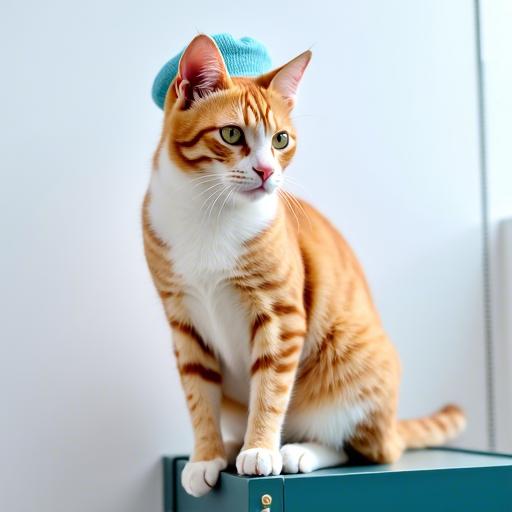} &
      \includegraphics[width=0.22\linewidth]{images/editing/cat/ours_a_cat_wearing_a_blue_hat_standing_on_fence_img_0.jpg} \\ 
    source & \makecell[t]{single-\\original} & \makecell[t]{single-\\previous} & dual \\
      
    \end{tabular}
    \caption{\textbf{Ablation study of single-objective LQR guidance.} Guidance based solely on the source image limits editability, while relying only on the previous step leads to accumulated error and artifacts.}
    \label{fig:ablation_lqr}
\end{figure}
\begin{figure}
    \centering
     \parbox{0.48\textwidth}{\centering \sout{man} $\rightarrow$ \textit{\textcolor{BrickRed}{superhero} }}

    \setlength{\tabcolsep}{1pt}
    \begin{tabular}{ccccc}
      \includegraphics[width=0.18\linewidth]{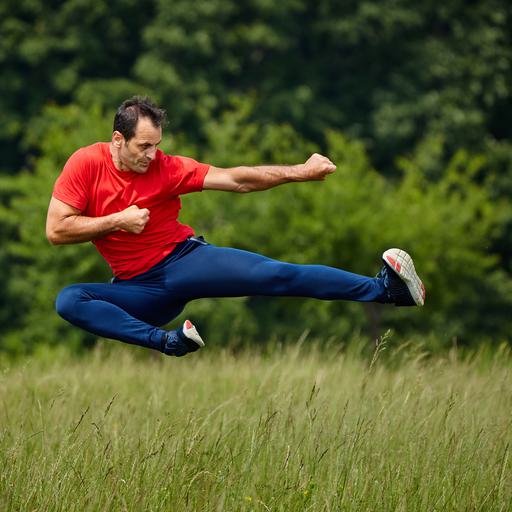}  &
      \includegraphics[width=0.18\linewidth]{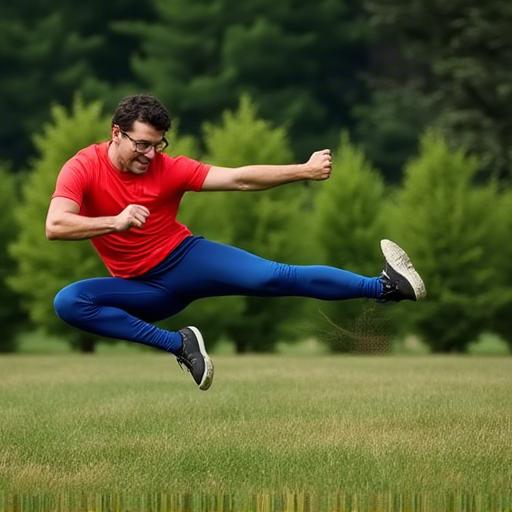}  &
      \includegraphics[width=0.18\linewidth]{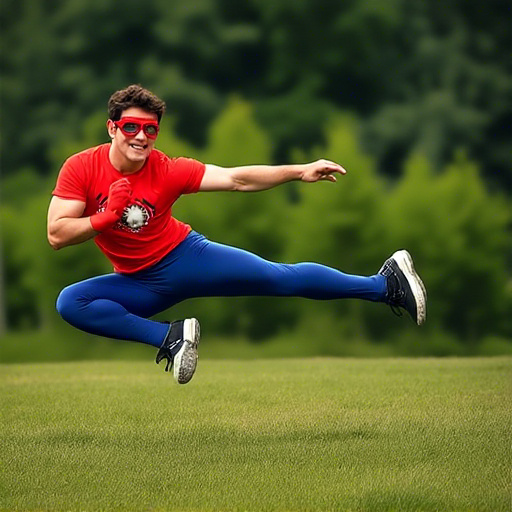} &
      \includegraphics[width=0.18\linewidth]{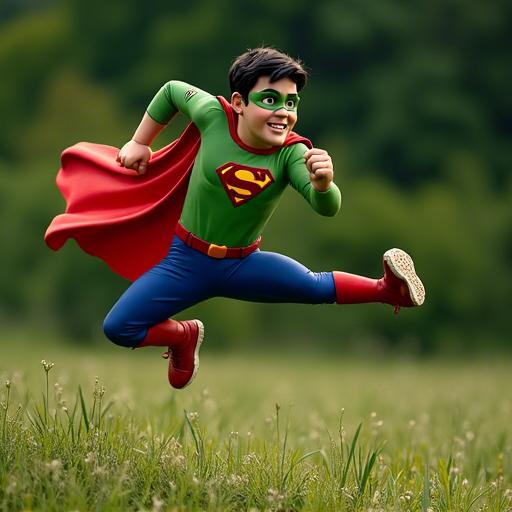} &
      \includegraphics[width=0.18\linewidth]{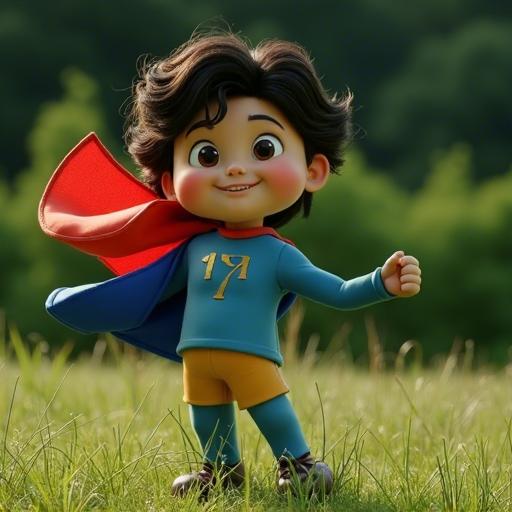} \\ 
    source & w/o attn & low & medium & high\\
      
    \end{tabular}
    \caption{\textbf{Ablation study of adaptive attention guidance.} Results demonstrate that editing without attention guidance struggles to affect salient areas, while increasing attention map activation leads to structural damage overly aggressive edits.}
    \label{fig:ablation_attention}
\end{figure}

\subsection{Ablation Study}
To evaluate the contribution of key components, we conducted ablation studies on: (1) the effect of using a single-objective LQR instead of our proposed dual-objective LQR (\cref{sec4_1:multi turn guidance}) in multi-turn editing; and (2) the impact of different attention map activation levels (\cref{sec:attention guidance}) on guiding editing performance.

We conduct a quantitative ablation study on the fourth-turn editing results, as shown in ~\cref{tab:ablation table}.
Relying solely on previous steps as single-objective LQR guidance leads to distribution bias, causing FID to increase significantly faster without the integration of dual-objective LQR guidance with the original image. Additionally, both highly activated attention guidance and the absence of attention mask guidance hinder content preservation. However, using highly activated attention as guidance improves editability.

~\cref{fig:ablation_lqr} shows that single-objective LQR guidance: 
LQR guidance based solely on the original image restricts editability, while relying only on previous steps leads to accumulated artifacts.
For the attention map ablation, we defined "low," "medium," and "high" activation levels based on the 19 double blocks in FLUX.1-dev 
(\cref{sec:attention guidance}), corresponding to the $12$$\sim$$17th$, $6$$\sim$$10th$, and top 5 most highly activated attention maps, respectively (\cref{fig:ablation_attention}).
Our results demonstrate that attention guidance is crucial for effective editing, as its absence resulted in limited editing of salient areas due to the strong LQR constriction. Furthermore, we observed that higher activation levels tended to damage the original image structure and background, while lower activation levels enabled more targeted editing of salient areas. For instance, when transforming a ``man'' into a ``superhero," the edits began with the glasses and cloak when using lower activation levels.

\section{Conclusion}

This paper investigated the workflow and necessities of multi-turn image editing, highlighting the limitations of existing approaches when adapted to this task. 
To overcome these limitations, we proposed a novel framework that integrates accurate flow matching inversion with a dual-objective LQR guidance method. 
Furthermore, we analyzed the roles of different transformer blocks within the DiT architecture and introduced a adaptive attention map selection mechanism to improve editability while preserving unaffected areas. 
Our experiments demonstrate the superior performance and adaptability of our method in multi-turn editing scenarios.

\noindent\textbf{Future Work:} 
Future work includes expanding our datasets and experiments to encompass a greater number of editing rounds, allowing for a more comprehensive evaluation of flow-based inversion techniques.
We also plan to investigate the synergies between multi-turn image editing and video editing, exploring how methods for ensuring temporal consistency in video can be adapted to maintain coherence across multiple editing iterations. 
Future work will also focus on automating token-specific attention map identification for enhanced editing precision and exploring the potential of single blocks, encoder spaces, and token spaces within the DiT architecture.

{
    \small
    \bibliographystyle{ieeenat_fullname}
    \normalem
    \bibliography{main}
}
\clearpage
\twocolumn[\begin{center}
    \large \textbf{Multi-turn Consistent Image Editing \\ Supplemental Material}
\end{center}]
\vspace{1em}

\setcounter{section}{0}
\renewcommand{\thesection}{\Alph{section}}

\tableofcontents
\newpage
\section{Datasets}
\label{sec:datasets}
\begin{figure}[htb]
    \centering
    \begin{subfigure}[b]{0.47\textwidth}
        \includegraphics[width=\linewidth]{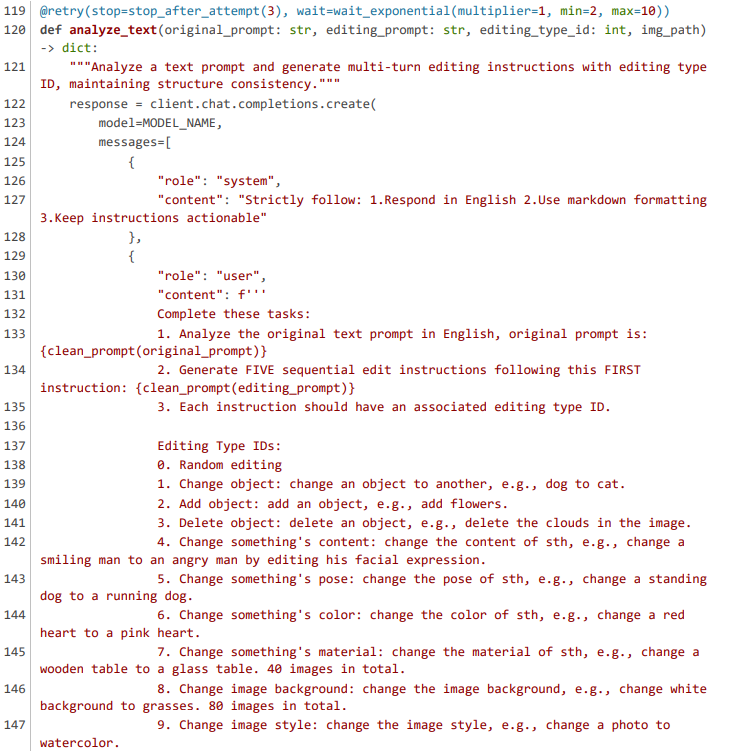}
        \label{fig:first}
    \end{subfigure}
    \begin{subfigure}[b]{0.47\textwidth}
        \includegraphics[width=\linewidth]{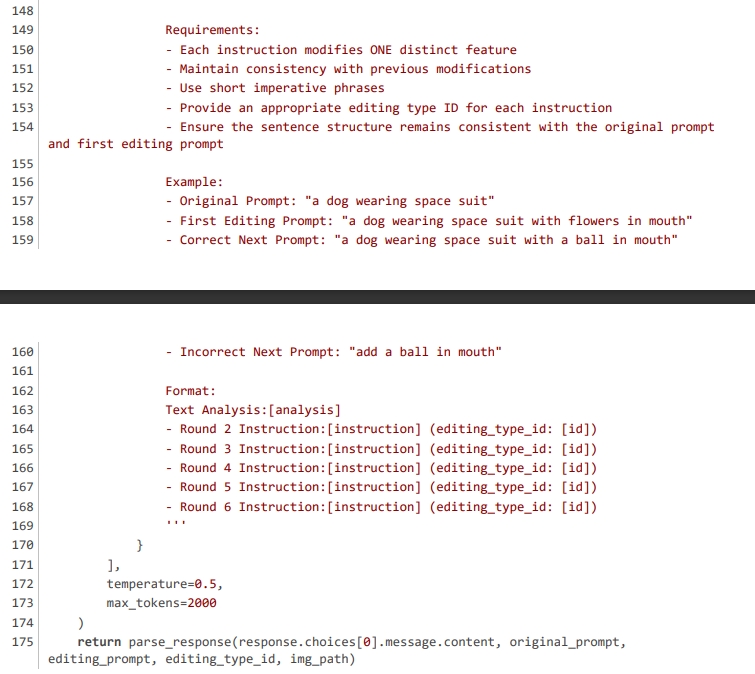}
        \label{fig:second}
    \end{subfigure}
    \caption{Prompts for GPT4-Turbo genrating multi-turn editing instrctions.}
    \label{fig:prompt_gpt}
\end{figure}

Since there are no existing datasets for multi-turn image editing, we propose an extended dataset based on PIE-Bench~\cite{ju2023direct} to facilitate evaluation. This extension allows for testing multi-turn editing while maintaining alignment with existing single-turn editing benchmarks. PIE-Bench consists of 10 editing types, as outlined below:
\begin{enumerate}
    \item Random editing: Random prompt written by volunteers or examples in previous research. 
    \item Change object: Change an object to another, e.g., dog to cat. 
    \item Add object: add an object, e.g., add flowers. 
    \item Delete object: delete an object, e.g., delete the clouds in the image. 
    \item Change sth's content: dhange the content of sth, e.g., change a smiling man to an angry man by editing his facial expression. 
    \item Change sth's pose: dhange the pose of sth, e.g., change a standing dog to a running dog.
    \item Change sth's color: change the color of sth, e.g., change a red heart to a pink heart. 
    \item Change sth's material: change the material of sth, e.g., change a wooden table to a glass table.
    \item Change image background: change the image background, e.g., change white background to grasses.
    \item Change image style: change the image style, e.g., change a photo to watercolor. 
\end{enumerate}
PIE-Bench is a dataset designed for single-turn editing, where each image is paired with an original prompt and an editing instruction. To extend it for multi-turn editing, we utilize OpenAI's GPT-4 Turbo to generate additional editing instructions. Based on the original prompt and the first-round editing instruction, we randomly select one of the ten editing types and generate five additional rounds of editing instructions for each image.
The prompts used for generating editing instructions are shown in ~\cref{fig:prompt_gpt}.

\section{Technical Proofs}
\label{sec:proof}
This section provides detailed technical proofs for the theoretical results discussed in this paper.
\subsection{Proof of Proposition 1}
\begin{proof}
The original problem with a single target is formulated as:
\[
V(c) \coloneqq \int_0^1 \frac{1}{2}\left\|c\left(Z_t, t\right)\right\|_2^2 \, \mathrm{d}t + \frac{\lambda}{2} \left\|Z_1 - X_1\right\|_2^2 
\]
The extended problem considering multiple targets is expressed as:
\[
V(c) \coloneqq \int_0^1 \frac{1}{2}\left\|c\left(Z_t, t\right)\right\|_2^2 \, \mathrm{d}t + \sum_{i=1}^n \frac{\lambda_i}{2} \left\|Z_1 - X_i\right\|_2^2 
\]
Rewriting the extended problem, the sum of squared distances is reformulated as:
\[
\sum_{i=1}^n \frac{\lambda_i}{2} \left\|Z_1 - Y_i\right\|_2^2 = \frac{\sum_{i=1}^n \lambda_i}{2} \left\|Z_1 - \mathbf{\mu}\right\|_2^2 + c
\]
where \(\mathbf{\mu}\) is defined as the weighted average of the targets:
\[
\mathbf{\mu} = \frac{\sum_{i=1}^n \lambda_i Y_i}{\sum_{i=1}^n \lambda_i}
\]
and $c$ corresponds to a constant value, which is irrelevant to $Z_1$. By defining a new target \(\mathbf{\mu}\) and a new weight \(\lambda' = \sum_{i=1}^n \lambda_i\), the extended problem simplifies to:
\[
V(c) \coloneqq \int_0^1 \frac{1}{2}\left\|c\left(Z_t, t\right)\right\|_2^2 \, \mathrm{d}t + \frac{\lambda'}{2} \left\|Z_1 - \mathbf{\mu}\right\|_2^2 
\]
This formulation is structurally identical to the single object LQR problem, where the target \(Y_1\) is replaced by \(\mathbf{\mu}\) and the weight \(\lambda\) is replaced by \(\lambda'\).
\end{proof}

\subsection{Solution to LQR Problem}
The standard approach to solving an LQR problem is the minimum principle theorem that can be found in control literature. We follow this approach and provide the full proof below for completeness. The Hamiltonian of the LQR problem is given by
\begin{align}
    H(\mathbf{z}_t, \mathbf{p}_t, \mathbf{c}_t, t) = \frac{1}{2}\left\|\mathbf{c}_t\right\|^2 + \mathbf{p}_t^T\mathbf{c}_t.
\end{align}
For \(\mathbf{c}_t^* = -\mathbf{p}_t\), the Hamiltonian attains its minimum value: \(H(\mathbf{z}_t, \mathbf{p}_t, \mathbf{c}_t^*, t) = -\frac{1}{2}\left\|\mathbf{p}_t \right\|^2\).
Using the minimum principle theorem, we get 
\begin{align}
\label{eq:lqr-1}
\frac{\mathrm{d} \mathbf{p}_t}{\mathrm{d} t} 
    &= \nabla_{\mathbf{z}_t} H\left(\mathbf{z}_t, \mathbf{p}_t, \mathbf{c}_t^*, t\right) = 0;\\
    \label{eq:lqr-2}
    \frac{\mathrm{d} \mathbf{z}_t}{\mathrm{d} t} 
    &= \nabla_{\mathbf{p}_t} H\left(\mathbf{z}_t, \mathbf{p}_t, \mathbf{c}_t^*, t\right) = -\mathbf{p}_t;\\
    \label{eq:lqr-3}
    \mathbf{z}_0 &= \mathbf{y}_0;\\
    \label{eq:lqr-4}
    \mathbf{p}_1 &= \nabla_{\mathbf{z}_1}  \left( \frac{\lambda}{2} \left\|\mathbf{z}_1 - \mathbf{y}_1\right\|_2^2\right) = \lambda \left(\mathbf{z}_1 - \mathbf{y}_1 \right).
\end{align}
From \eqref{eq:lqr-1}, we know \(\mathbf{p}_t\) is a constant \(\mathbf{p}\). 
Using this constant in \eqref{eq:lqr-2} and integrating from \(t \rightarrow 1\), we have \(\mathbf{z}_1 = \mathbf{z}_t - \mathbf{p} (1-t)\).
Substituting \(\mathbf{z}_1\) in \eqref{eq:lqr-3}, 
\begin{align*}
    \mathbf{p} = \lambda (\mathbf{z}_t - \mathbf{p}(1-t)-\mathbf{y}_1) = \lambda (\mathbf{z}_t - \mathbf{y}_1) - \lambda (1-t)\mathbf{p},
\end{align*}
which simplifies to 
\begin{align*}
    \mathbf{p} &= \left(1+\lambda (1-t)\right)^{-1} \lambda (\mathbf{z}_t - \mathbf{y}_1) \\
               &= \left(\frac{1}{\lambda}+(1-t)\right)^{-1} (\mathbf{z}_t - \mathbf{y}_1).
\end{align*}
Taking the limit \(\lambda \rightarrow \infty\), we get \(\mathbf{p} = \frac{\mathbf{z}_t - \mathbf{y}_1}{1-t}\) and the optimal controller \(\mathbf{c}_t^* = \frac{\mathbf{y}_1 - \mathbf{z}_t}{1-t}\). Since \(u_t(\mathbf{z}_t | \mathbf{y}_1) = \mathbf{y}_1 - \mathbf{y}_0\), the proof follows by substituting \(\mathbf{y}_0 = \frac{\mathbf{z}_t - t\mathbf{y}_1}{1-t}\).

In conclusion, the formulation with multiple targets can be regarded as a special case of the original single-target Linear Quadratic Regulator (LQR) problem. In this interpretation, the effective target is a weighted average of the individual targets, and the effective weight is the sum of the individual weights. This allows for the seamless application of the optimal control techniques developed for the single-target scenario to be extended to handle the multi-target problem by treating the weighted average target as the effective target.


\section{Limitations}
\label{sec:formatting}

\subsection{Editing Iterations}
As shown in ~\cref{fig:german}, our method effectively preserves the natural appearance of images across multiple editing rounds, whereas other methods exhibit noticeable artifacts. However, due to limitations in dataset generation, we created only five rounds of editing instructions. Additionally, errors from ChatGPT restricted our benchmark evaluation to four editing turns.

As a result, we have not yet fully explored the potential of our method across a larger number of editing iterations. As seen in the reconstruction results presented in this paper, most flow-based inversion methods begin to exhibit significant semantic drift by the fourth reconstruction. In contrast, our multi-turn reconstruction results demonstrate that even after 10 reconstruction steps reconstruction, our method maintains high-quality outputs.

Since our evaluation was limited to only four editing rounds, a comprehensive comparison between methods remains incomplete. Moving forward, we aim to extend the multi-turn dataset to support a greater number of editing iterations for a more thorough evaluation.

\subsection{First Round Editing}
LQR-guided methods are highly effective in aligning distributions, particularly in transforming atypical distributions into typical ones. This capability is essential for maintaining coherence in multi-turn editing. However, in single-turn editing, LQR guidance can disrupt the original flow matching process to some degree. Consequently, the performance of our method in the initial editing round is suboptimal. Future work could explore alternative methods to integrate information across editing iterations.

\section{Additional Experiments}
~\label{sup c: experiment}
 \begin{table*}[htbp]
		\centering
        \belowrulesep=0pt
        \aboverulesep=0pt
        \scalebox{0.9}{
		\begin{tabular}{l|cccccccccccc}
        \Xhline{1.5pt}
            \multirow{2}{*}{Methods}
			& \multicolumn{3}{c}{Round 1}
			&\multicolumn{3}{c}{Round 2}
            &\multicolumn{3}{c}{Round 3}
             &\multicolumn{3}{c}{Round 4}\\
			\cmidrule(lr){2-4} \cmidrule(lr){5-7} \cmidrule(lr){8-10}\cmidrule(lr){11-13}
		 
          &FID & Clip-T& Clip-I& 
          FID & Clip-T& Clip-I& 
          FID & Clip-T& Clip-I &
          FID & Clip-T& Clip-I 
          \\ 
			\Xhline{1pt}
			\rowcolor{gray!20}
            Ours
            & 2.554 & 26.19  & 0.910
            & 4.015 & \cellcolor{blue!15}26.56 &  0.903
            & \cellcolor{blue!15} 5.115 & \cellcolor{blue!15} 26.81 & 0.897
            &\cellcolor{green!15} 5.553 & \cellcolor{blue!15}26.83 & 0.894 \\ \hline
           RF-Inv.
           & 1.854 & 24.41 & 0.928
           &\cellcolor{green!15} 3.015 & 24.09 & 0.919
           & \cellcolor{green!15} 4.324 & 24.06 & 0.909
           &\cellcolor{blue!15} 5.740 & 24.10 &\cellcolor{blue!15} 0.904\\ \hline
           StableFlow 
           & 1.699 & 23.94 &\cellcolor{green!15} 0.940
           & 5.971 & 23.98 & \cellcolor{green!15} 0.932
           & 12.413 & 23.94 &\cellcolor{green!15} 0.914
           & 20.624 & 24.23 & 0.899\\ \hline
            
	        FlowEdit 
            &\cellcolor{green!15} 0.998 & \cellcolor{blue!15}26.28 &\cellcolor{blue!15} 0.932 
            & 3.706 & 26.34 & 0.914
            & 8.405 & 26.36  & 0.903
            & 14.547 & 26.70 & 0.894\\ \hline
            
		    RF-Solver
            & \cellcolor{blue!15} 1.450 & 25.58 & 0.931 
            & \cellcolor{blue!15}3.419 & 25.55 &\cellcolor{blue!15} 0.922 
            & 6.603 & 25.62 & \cellcolor{blue!15} 0.912
            & 11.581 & 25.52 & \cellcolor{green!15}0.906
            \\ \hline

            FireFlow
            & 5.579 & \cellcolor{green!15} 27.72 & 0.891 
            & 8.279 & \cellcolor{green!15}27.87 & 0.883
            & 8.405 &\cellcolor{green!15} 27.94 & 0.878
            & 12.375 &\cellcolor{green!15} 28.28 & 0.873\\ \hline

            MasaCtrl
            & 1.647 & 23.98 & 0.933 
            & 4.518 & 23.80 & 0.915 
            & 7.609 & 23.91 & 0.900
            & 10.811 & 23.80 & 0.886\\ \hline

            PnPInv.
            & 2.222  & 25.25 & 0.915 
            & 4.927 & 25.67 & 0.901 
            & 7.703  & 25.47 & 0.889 
            & 10.262  & 25.77 & 0.872\\
        \Xhline{1.5pt}
		\end{tabular}}
          \caption{\textbf{Quantitative Results of Multi-Turn Editing.} The best results are highlighted in green, while the second-best results are marked in purple. Our method demonstrates a balance between CLIP-I and CLIP-T while achieving the best FID score at the fourth-turn editing.}
        \label{tab:multi_turn_editing}
	\end{table*}

 \begin{table*}[htbp]
		\centering
        \belowrulesep=0pt
        \aboverulesep=0pt
        \scalebox{0.9}{
		\begin{tabular}{l|cccccccccccc}
        \Xhline{1.5pt}
            \multirow{2}{*}{Methods}
			& \multicolumn{2}{c}{Round 1}
			&\multicolumn{2}{c}{Round 2}
            &\multicolumn{2}{c}{Round 3}
             &\multicolumn{2}{c}{Round 4}\\
			\cmidrule(lr){2-3} \cmidrule(lr){4-5} \cmidrule(lr){6-7}\cmidrule(lr){8-9}
		 
          &CLIP-edit & Structure &CLIP-Edit& 
          Structure & CLIP-Edit& Structure& 
          CLIP-Edit & Structure
          \\ 
			\Xhline{1pt}
			\rowcolor{gray!20}
            Ours
            &\cellcolor{blue!15} 23.596 & 0.0475 
            & 23.120  & 0.0587
            & \cellcolor{blue!15} 23.294  & 0.0652 
            & \cellcolor{blue!15} 23.021  &\cellcolor{blue!15}  0.0580 \\ \hline
            
           RF-Inv.
           & 21.573 & 0.0326 
           & 21.834 & 0.0411
           & 21.920 &\cellcolor{green!15} 0.0471 
           & 21.945 & \cellcolor{green!15}0.0525 \\ \hline
           
           StableFlow 
           & 21.187 & \cellcolor{green!15}0.0190  
           & 21.581 & \cellcolor{blue!15} 0.0375
           & 21.926 & 0.0589  
           & 22.051 & 0.0785 \\ \hline
            
	        FlowEdit 
            & 23.393 & 0.0289 
            & 23.378  & 0.0493
            & 23.237  & 0.0668
            & 22.941  & 0.0813 \\ \hline
            
		    RF-Solver
           & 22.536 &\cellcolor{blue!15}  0.0249  
           & 23.101  &\cellcolor{green!15} 0.0359 
           & 23.229  &\cellcolor{blue!15}  0.0488 
           & 22.581 & 0.0611 \\ \hline

            FireFlow
           & \cellcolor{green!15}24.226 & 0.0780  
           & \cellcolor{green!15}23.843  & 0.1040 
           & \cellcolor{green!15}23.524  & 0.1240  
           & \cellcolor{green!15}23.208  & 0.1420 \\ \hline
            
            MasaCtrl
            & 21.073 & 0.0271  
            & 21.557 & 0.0456 
            & 21.621 & 0.0595 
            & 21.776 & 0.0709 \\ \hline

            PnPInv.
            & 22.502 & 0.0218  
            & 22.859  & 0.0424 
            & 22.788  & 0.0580  
            & 22.752  & 0.0692 \\ 
        \Xhline{1.5pt}
		\end{tabular}}
          \caption{\textbf{Quantitative Results of Multi-Turn Editing.} The best results are highlighted in green, while the second-best results are marked in purple.}
        \label{tab:multi_turn_editing_2}
	\end{table*}

In this section, we begin by presenting comprehensive experimental metrics across multiple editing rounds~\cref{sup: quantative}. 
Next, we showcase quantitative results demonstrating that our method is highly effective for multi-turn editing, excelling in both editability and structure preservation~\cref{sup:qualitative}.
Finally, we conduct additional ablation studies to analyze the functionality of key components.

\subsection{Quantitative Results}
\label{sup: quantative}
We utilize CLIP-T~\cite{radford2021learning} to measure image-text similarity, while CLIP-I and structure-distance metrics assess the similarity between the edited and original images. The Fréchet Inception Distance (FID) is employed to evaluate the quality of the generated images. Additionally, since PIE-bench provides a mask labeling the edited area, we use CLIP-Edit to measure image-text similarity specifically within the edited region.

\begin{figure}[t]
    \centering
    \includegraphics[width=0.8\linewidth]{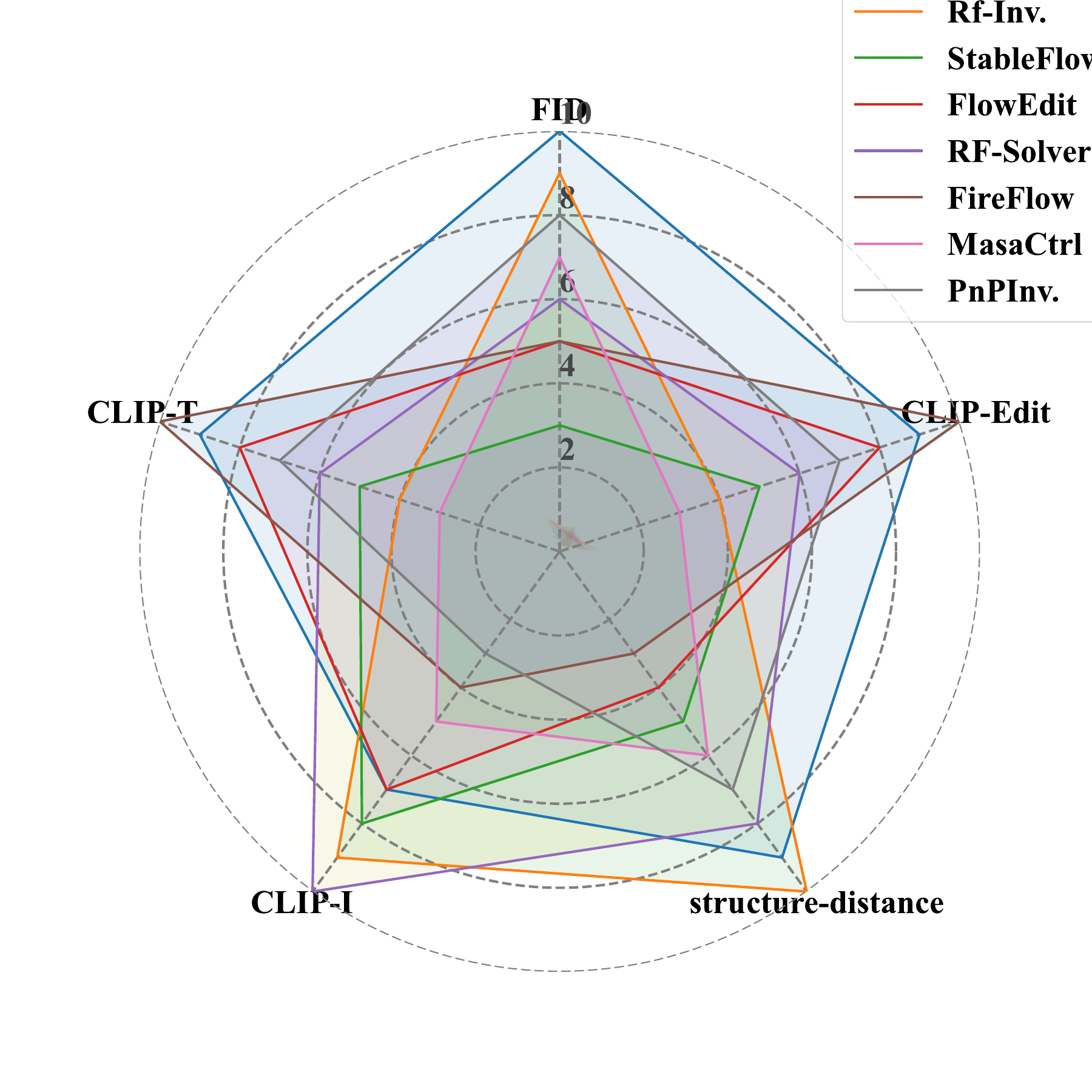}
    \caption{We rank the performance of our method compared to baseline methods in the fourth round of editing. Our method performs well in both text similarity and fidelity to the original image.}
    \label{fig:radar}
\end{figure}

Quantitative results are presented in ~\cref{tab:multi_turn_editing} and~\cref{tab:multi_turn_editing_2}. Our method demonstrates a strong balance between content preservation and editing capability, particularly in the fourth round of editing. Notably, for the FID and structure-distance metrics, our method maintains stable performance across multiple editing turns, whereas most competing methods exhibit a continuous increase in both structure distance and FID as the number of editing rounds grows. Furthermore, our multi-turn approach achieves comparable performance to state-of-the-art flow-based editing methods in the initial rounds and delivers outstanding performance in later rounds.
To comprehensively evaluate overall performance, we compare our method with baseline methods on fourth-turn editing results. All metrics are normalized to a 0-10 ranking and visualized using a radar plot, which shows that our method strikes a balance across all metrics.(~\cref{fig:radar})

\subsection{Qualitative Results}
\label{sup:qualitative}
Existing metrics cannot accurately assess image quality. For example, our selected baseline diffusion-based methods produce noticeable artifacts compared to flow-based methods. However, qualitative evaluations do not always capture these differences effectively.

\begin{figure*}[htbp]
    \centering
    \setlength{\tabcolsep}{2pt} 
    \renewcommand{\arraystretch}{0.8} 

    \begin{minipage}[t]{0.09\textwidth}
        \centering
        \vspace{-3.4cm}
        \includegraphics[width=\linewidth]{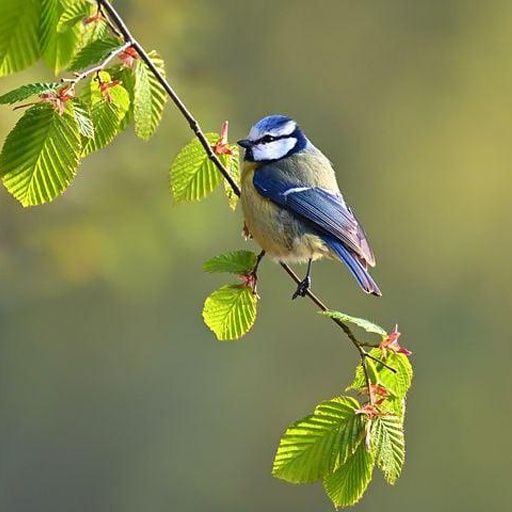}
        \vspace{-0.8cm}
        \begin{center}
        source
    \end{center}
        \vspace{0.3cm} 
        \scalebox{0.9}{\downarrowwithtext[OliveGreen, ultra thick]{\textit{Editing} \\ \textit{Turn}}}
        
    \end{minipage}%
    \hspace{2em}
    \begin{minipage}[t]{0.84\textwidth}
        \begin{tabular}{p{0.3cm}cccccccc}
            &Ours & RF-Inv. & StableFlow & FlowEdit & RF-Solver & FireFlow &  MasaCtrl & PnPInv. \\
            \scalebox{0.9}{\raisebox{0cm}{\rotatebox{90}{bird $\rightarrow$\textcolor{BrickRed}{\textit{butterfly}}}}}&
            \includegraphics[width=0.11\textwidth]{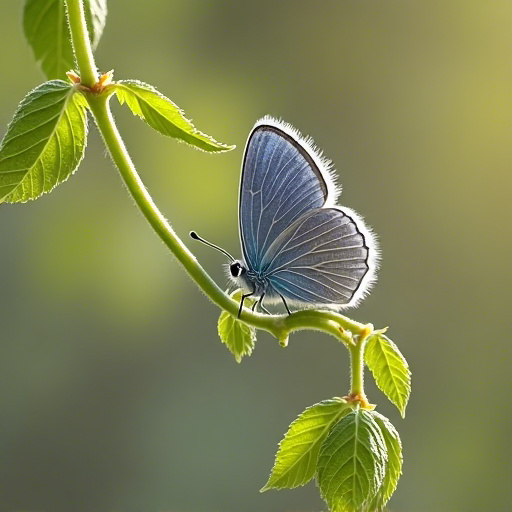} &
            \includegraphics[width=0.11\textwidth]{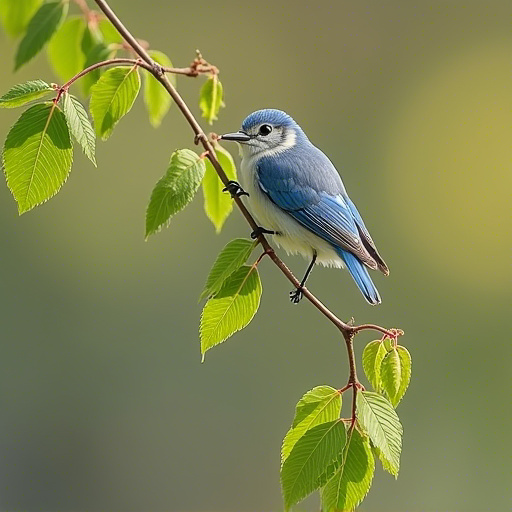} &
            \includegraphics[width=0.11\textwidth]{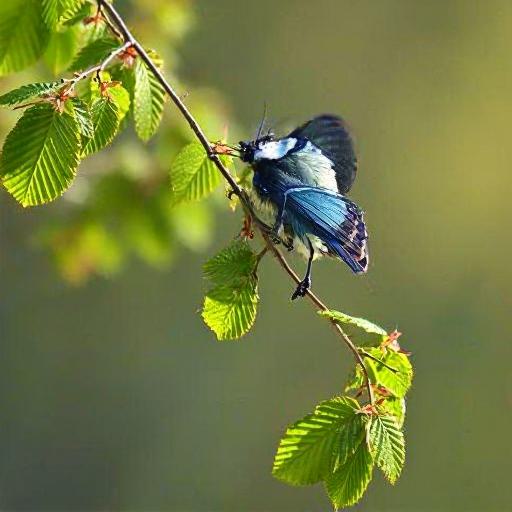} &
            \includegraphics[width=0.11\textwidth]{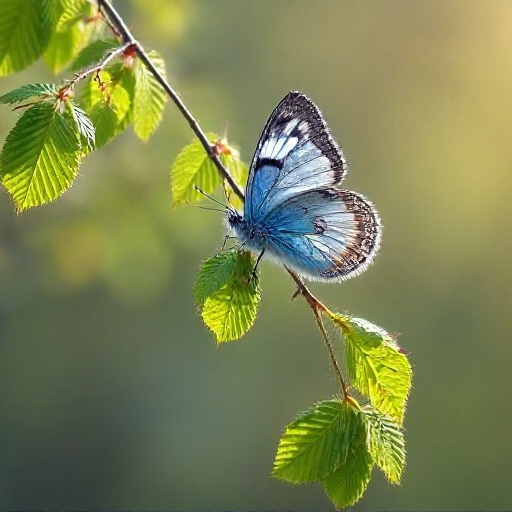} &
            \includegraphics[width=0.11\textwidth]{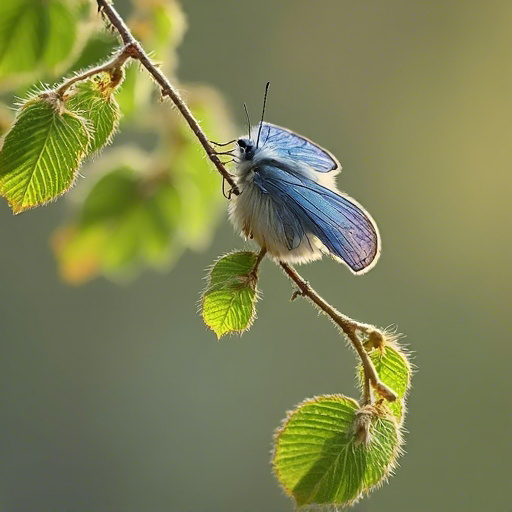} &
            \includegraphics[width=0.11\textwidth]{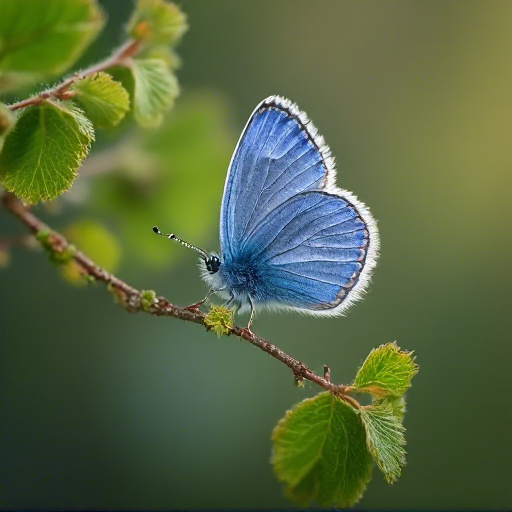} &
            \includegraphics[width=0.11\textwidth]{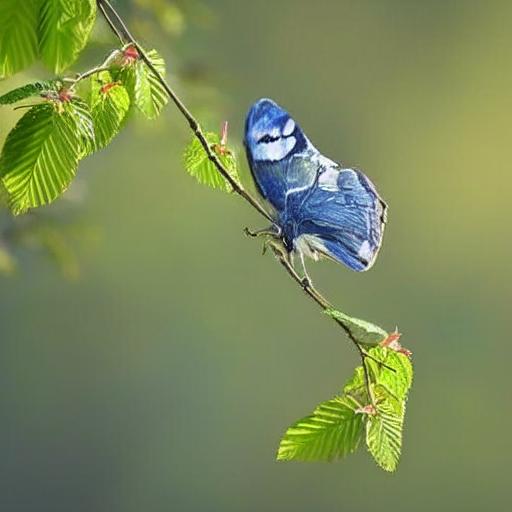} &
            \includegraphics[width=0.11\textwidth]{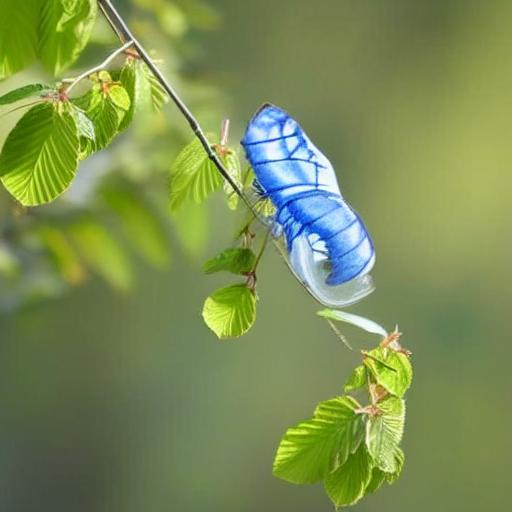} \\

            \scalebox{0.9}{\raisebox{0.0cm}{\rotatebox{90}{\textcolor{BrickRed}{\textit{pink flower}}}}}&
            \includegraphics[width=0.11\textwidth]{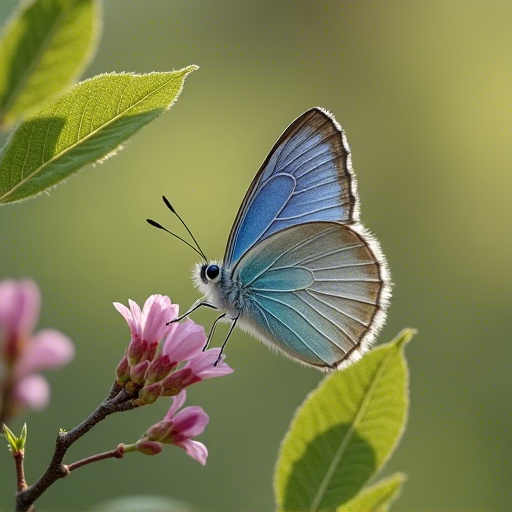} &
            \includegraphics[width=0.11\textwidth]{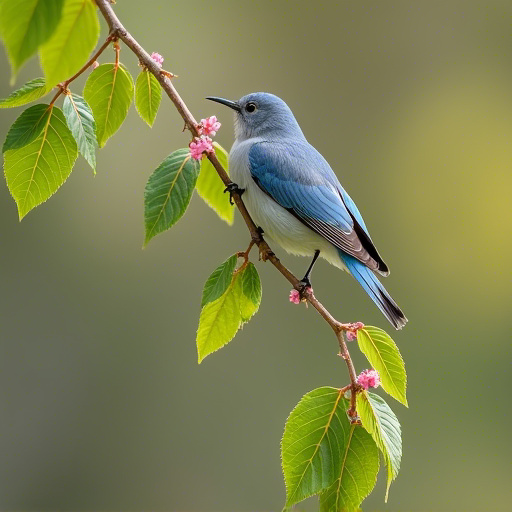} &
            \includegraphics[width=0.11\textwidth]{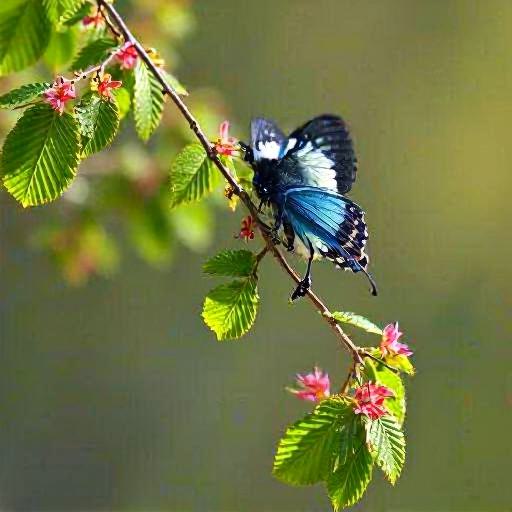} &
            \includegraphics[width=0.11\textwidth]{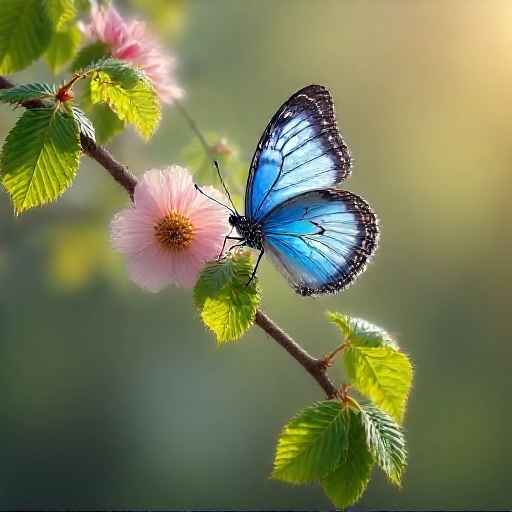} &
            \includegraphics[width=0.11\textwidth]{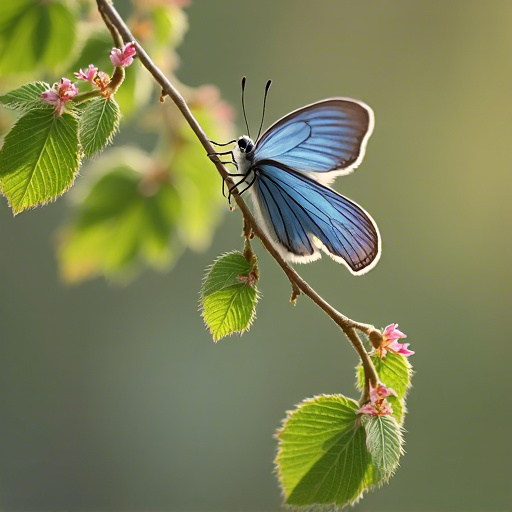} &
             \includegraphics[width=0.11\textwidth]{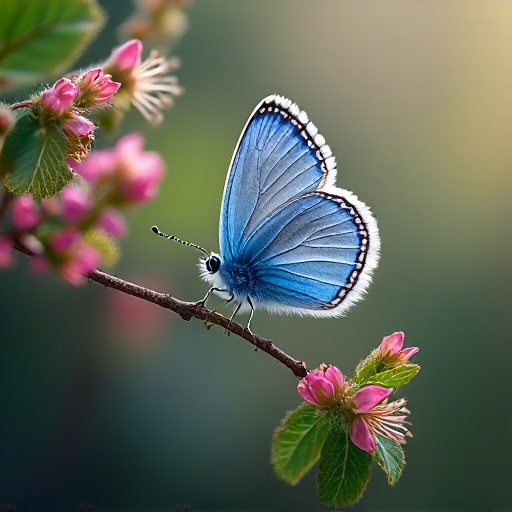} &
            \includegraphics[width=0.11\textwidth]{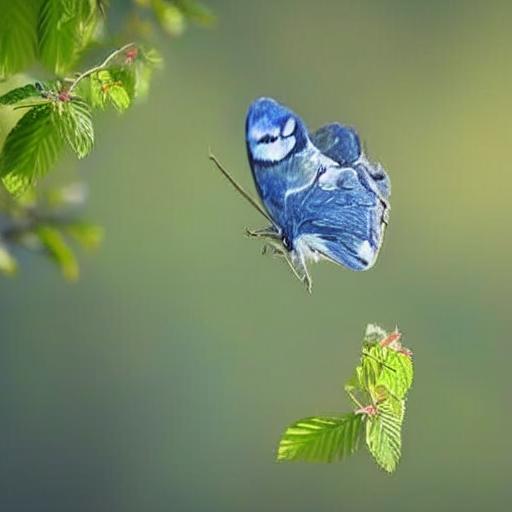} &
            \includegraphics[width=0.11\textwidth]{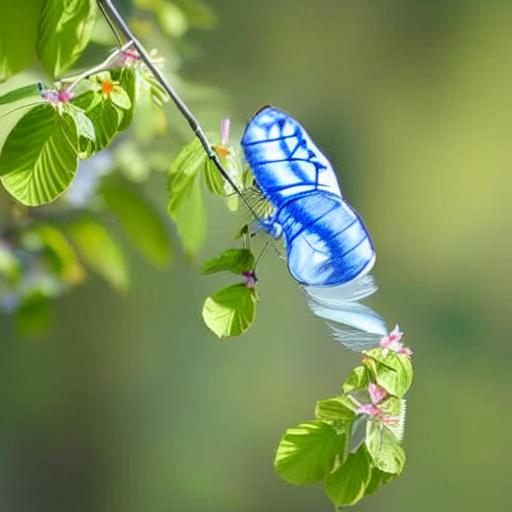} \\
            
            \scalebox{0.9}{\raisebox{+0.3cm}{\rotatebox{90}{+ \textcolor{BrickRed}{\textit{ladybug}}}}}&
            \includegraphics[width=0.11\textwidth]{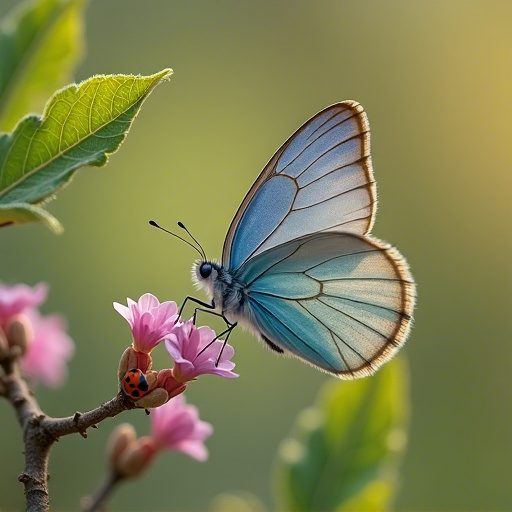} &
            \includegraphics[width=0.11\textwidth]{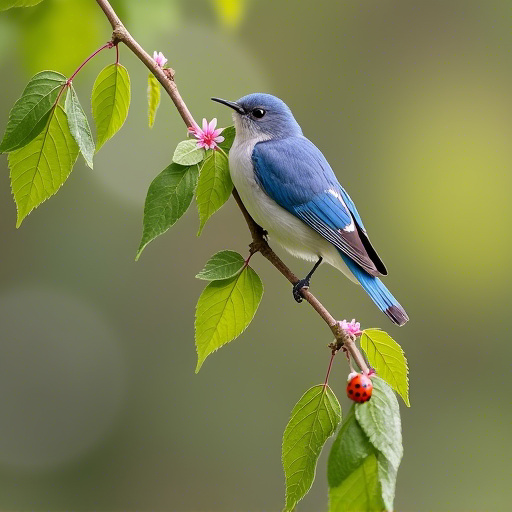} &
            \includegraphics[width=0.11\textwidth]{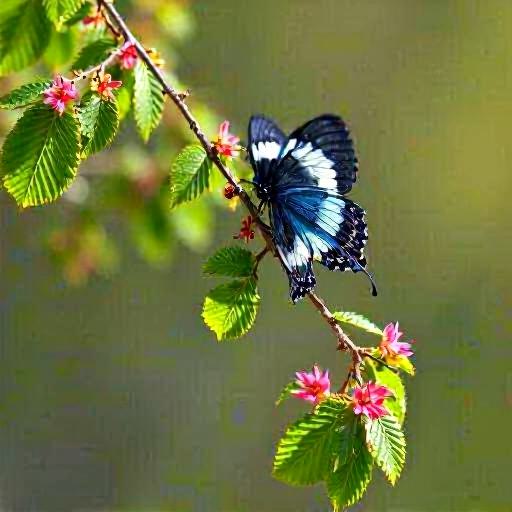} &
            \includegraphics[width=0.11\textwidth]{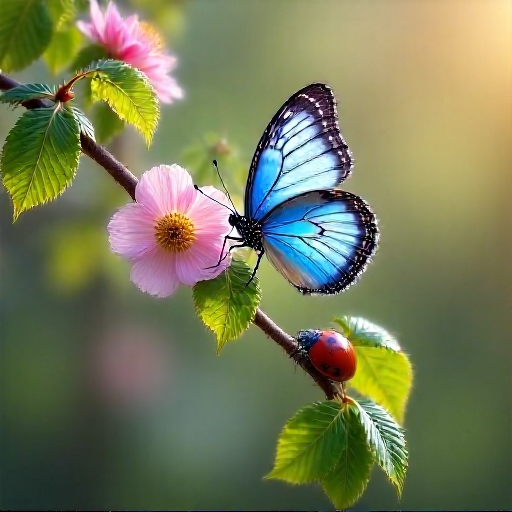} &
            \includegraphics[width=0.11\textwidth]{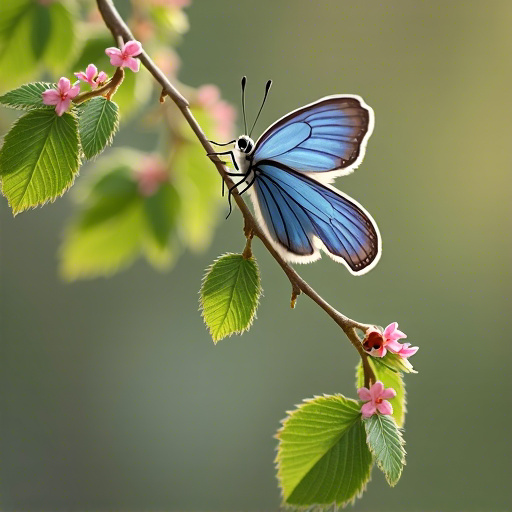} &
            \includegraphics[width=0.11\textwidth]{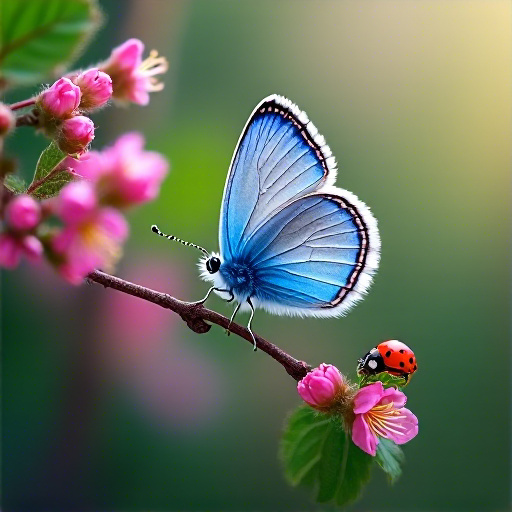} &
            \includegraphics[width=0.11\textwidth]{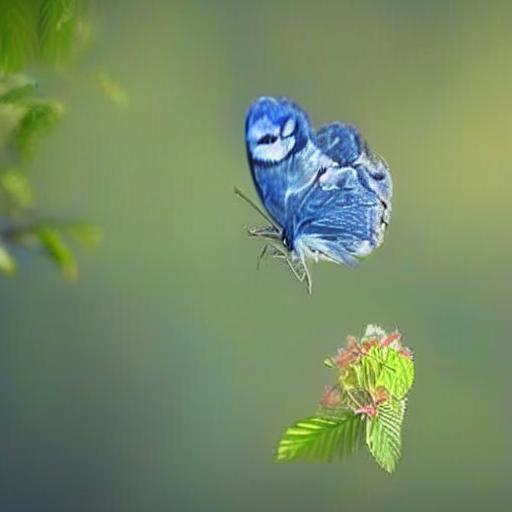} &
            \includegraphics[width=0.11\textwidth]{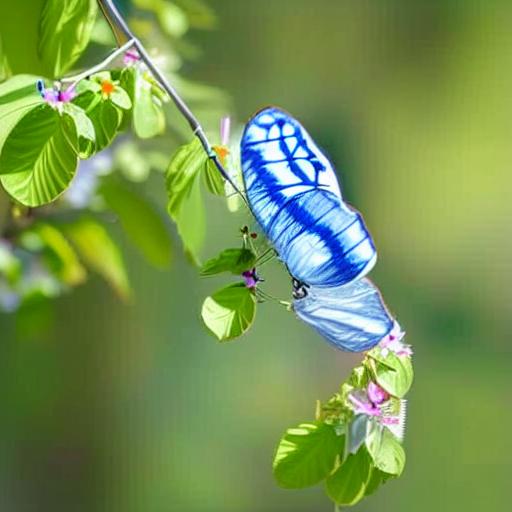} \\
            
            \scalebox{0.9}{\raisebox{0.6cm}{\rotatebox{90}{\textcolor{BrickRed}{\textit{Rec.}}}}}&
           \includegraphics[width=0.11\textwidth]{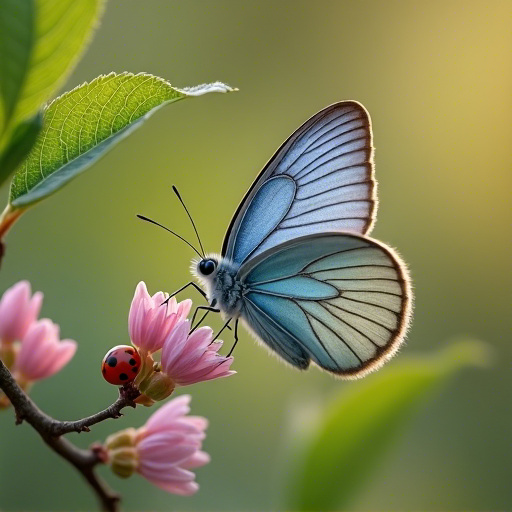} &
            \includegraphics[width=0.11\textwidth]{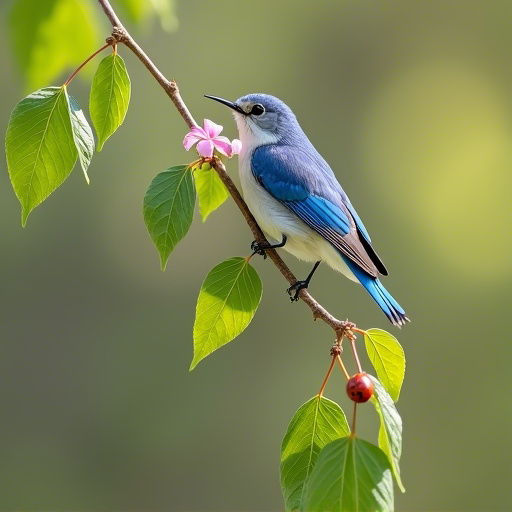} &
            \includegraphics[width=0.11\textwidth]{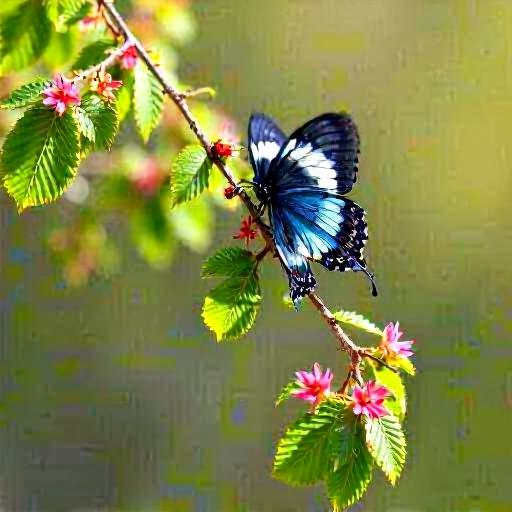} &
            \includegraphics[width=0.11\textwidth]{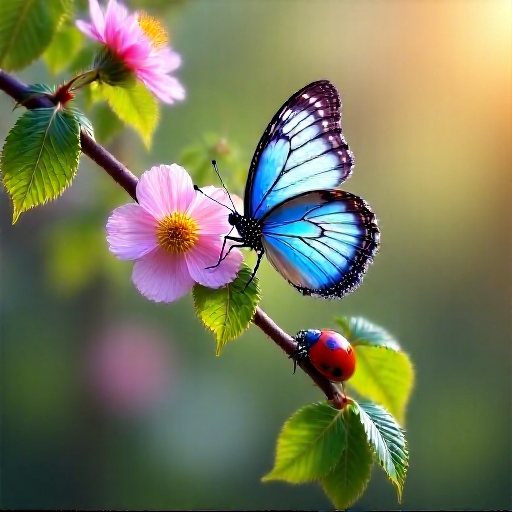} &
            \includegraphics[width=0.11\textwidth]{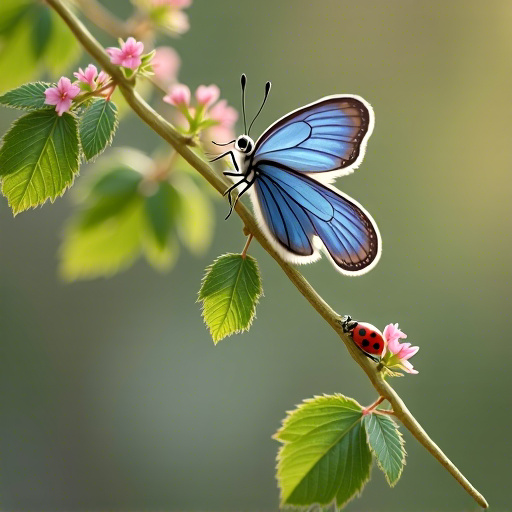} &
            \includegraphics[width=0.11\textwidth]{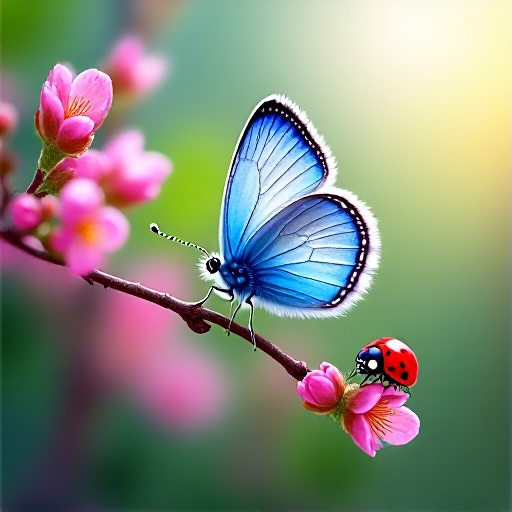} &
            \includegraphics[width=0.11\textwidth]{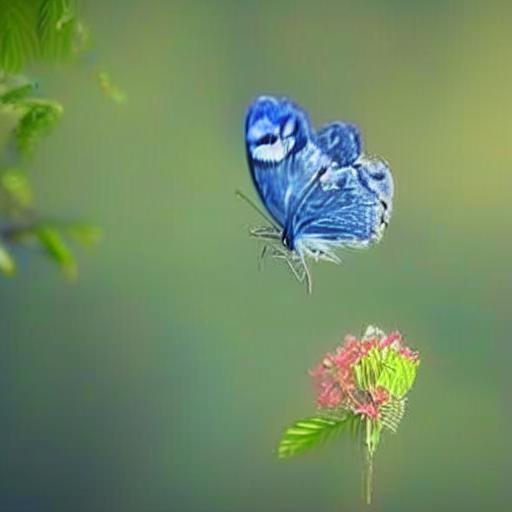} &
            \includegraphics[width=0.11\textwidth]{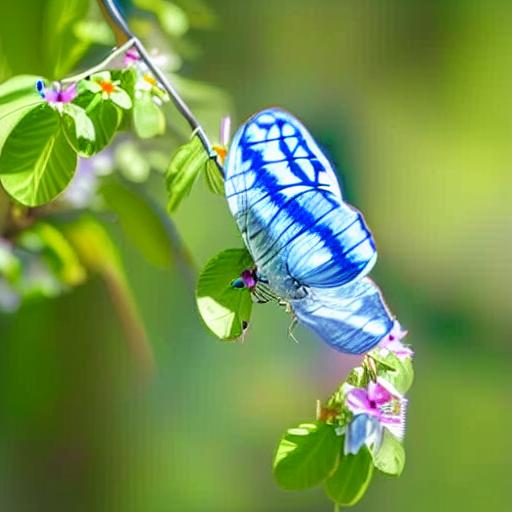}
        \end{tabular}
    \end{minipage}
    \captionsetup{skip=0pt} 
    \caption{Our method consistently follows the color tone of the original image while achieving the desired editing. The second prompt is ``sitting on a pink flower'', while the third prompt is ``with a red ladybug''.}
    \label{fig:butterfly}

\end{figure*}

\begin{figure*}[htbp]
    \centering
    \setlength{\tabcolsep}{2pt} 
    \renewcommand{\arraystretch}{0.8} 

    \begin{minipage}[t]{0.09\textwidth}
        \centering
        \vspace{-3.4cm}
        \includegraphics[width=\linewidth]{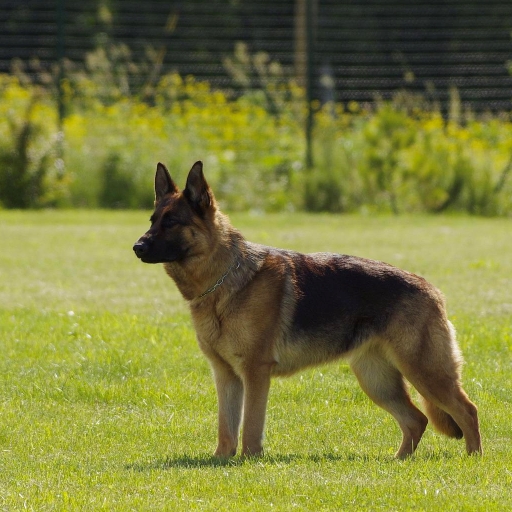}
        \vspace{-0.8cm}
        \begin{center}
        source
    \end{center}
        \vspace{0.3cm} 
        \scalebox{0.9}{\downarrowwithtext[OliveGreen, ultra thick]{\textit{Editing} \\ \textit{Turn}}}
        
    \end{minipage}%
    \hspace{2em}
    \begin{minipage}[t]{0.84\textwidth}
        \begin{tabular}{p{0.3cm}cccccccc}
            &Ours & RF-Inv. & StableFlow & FlowEdit & RF-Solver & FireFlow &  MasaCtrl & PnPInv. \\
            \scalebox{0.9}{\raisebox{0cm}{\rotatebox{90}{\textcolor{BrickRed}{\textit{mouth opened}}}}}&
            \includegraphics[width=0.11\textwidth]{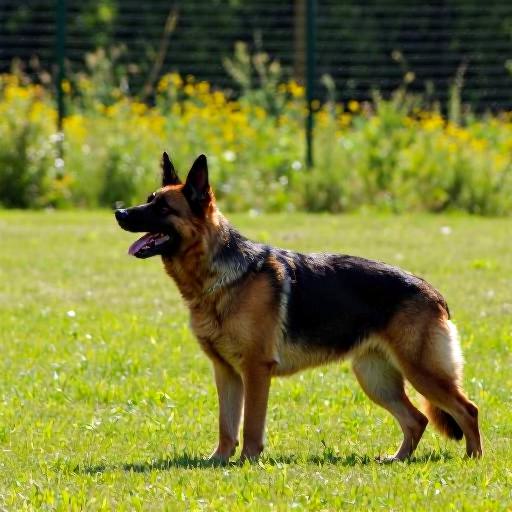} &
            \includegraphics[width=0.11\textwidth]{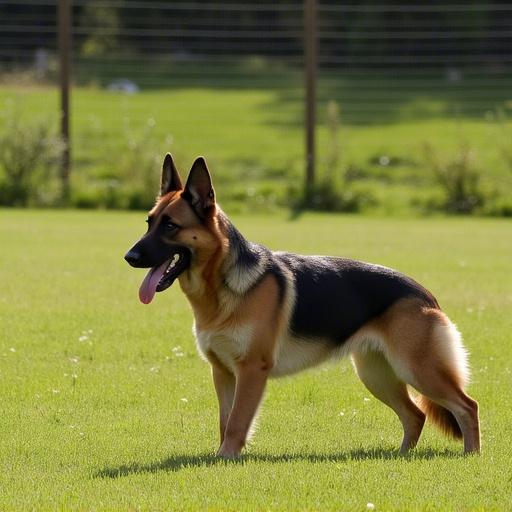} &
            \includegraphics[width=0.11\textwidth]{supplement/imgs/german/german_shepherd/stable_flow_a_German_Shepherd_dog_stands_on_the_grass_with_mouth_opened.jpg} &
            \includegraphics[width=0.11\textwidth]{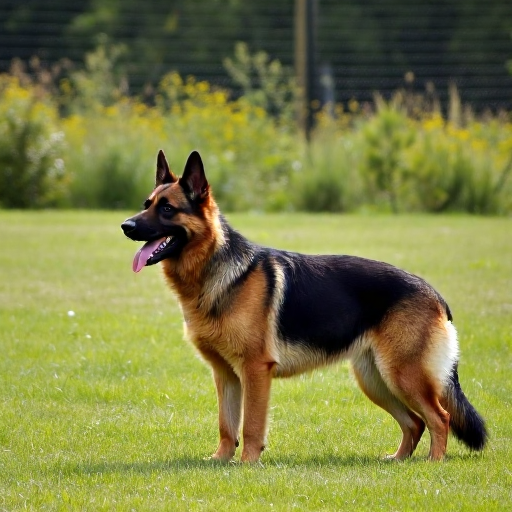} &
            \includegraphics[width=0.11\textwidth]{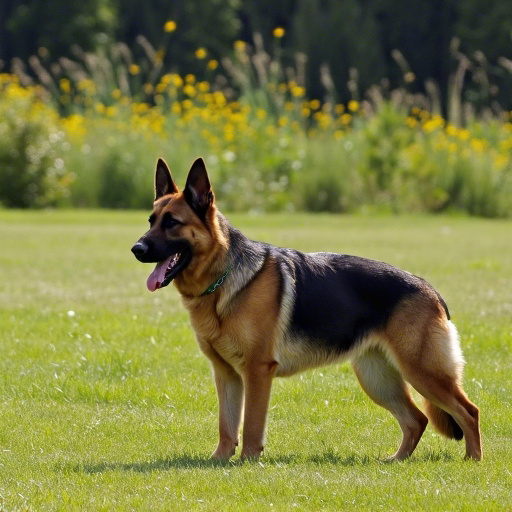} &
            \includegraphics[width=0.11\textwidth]{supplement/imgs/german/german_shepherd/flowedit_a_German_Shepherd_dog_stands_on_the_grass_with_mouth_opened.png} &
            \includegraphics[width=0.11\textwidth]{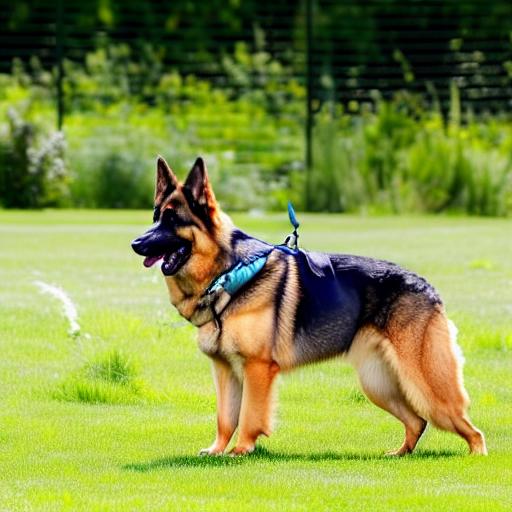} &
            \includegraphics[width=0.11\textwidth]{supplement/imgs/german/german_shepherd/pnp-inv_a_German_Shepherd_dog_stands_on_the_grass_with_mouth_opened.jpg} \\

            \scalebox{0.9}{\raisebox{0.7cm}{\rotatebox{90}{\textcolor{BrickRed}{\textit{Rec.}}}}}&
            \includegraphics[width=0.11\textwidth]{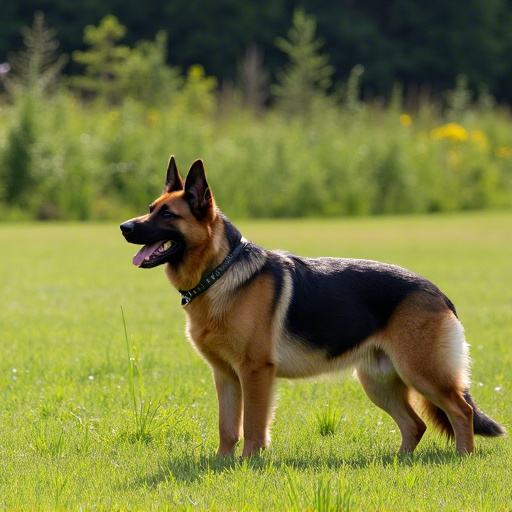} &
            \includegraphics[width=0.11\textwidth]{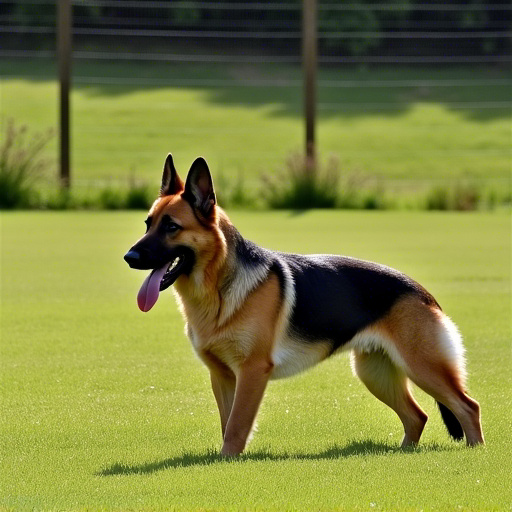} &
            \includegraphics[width=0.11\textwidth]{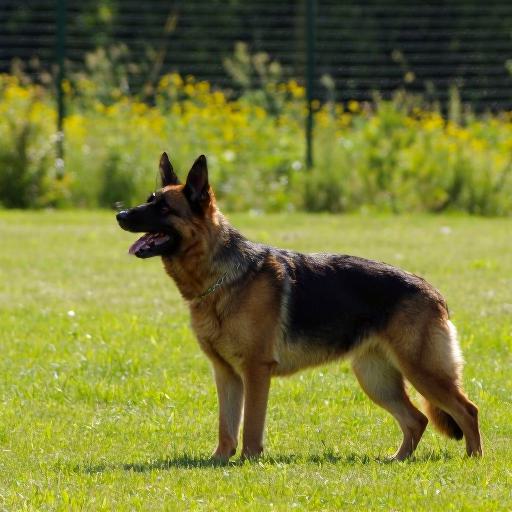} &
            \includegraphics[width=0.11\textwidth]{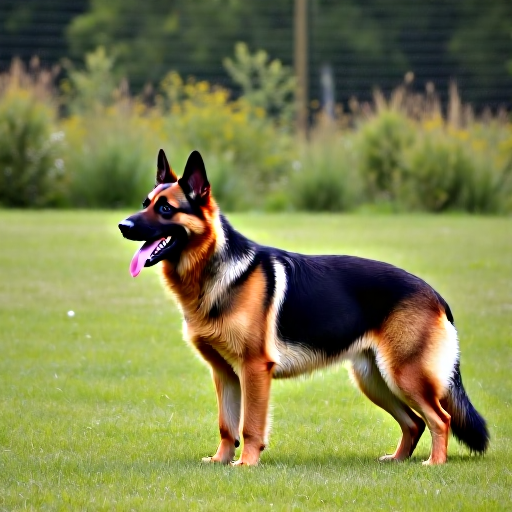} &
            \includegraphics[width=0.11\textwidth]{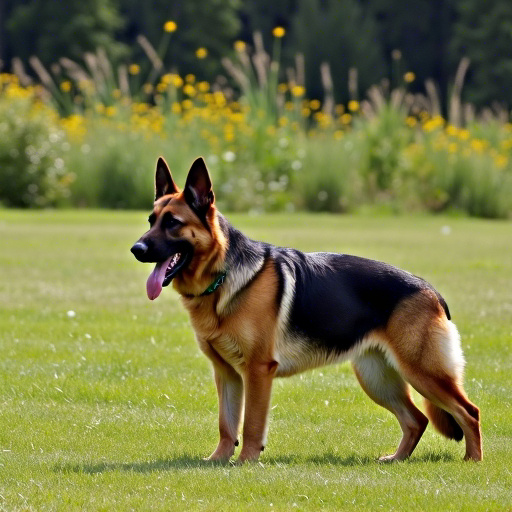} &
            \includegraphics[width=0.11\textwidth]{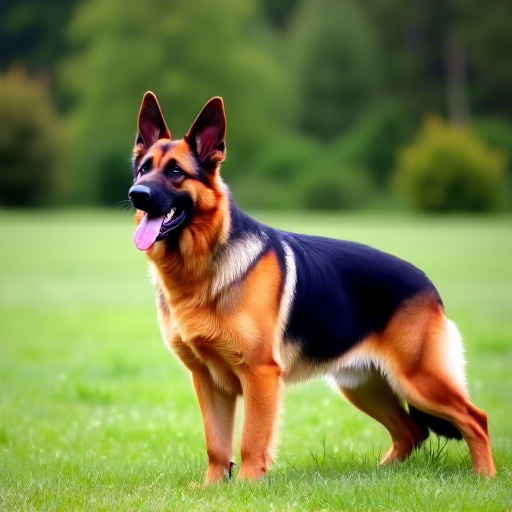} &
            \includegraphics[width=0.11\textwidth]{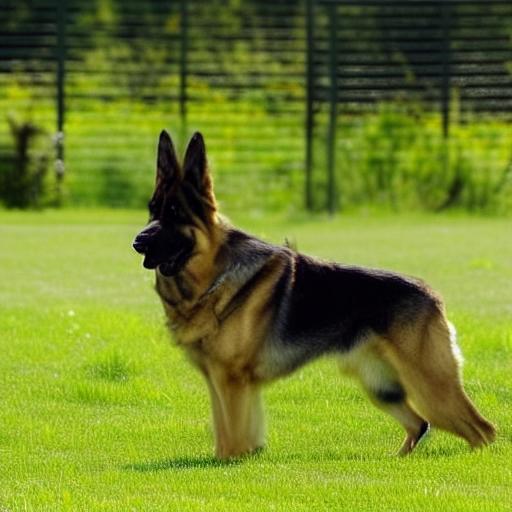} &
            \includegraphics[width=0.11\textwidth]{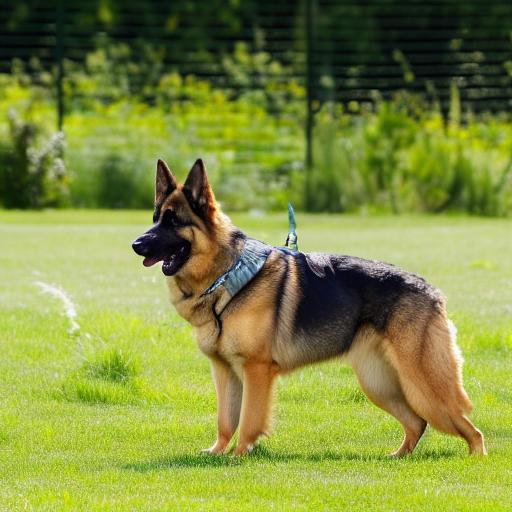} \\
            
            \scalebox{0.9}{\raisebox{+0.3cm}{\rotatebox{90}{+ \textcolor{BrickRed}{\textit{red ball}}}}}&
            \includegraphics[width=0.11\textwidth]{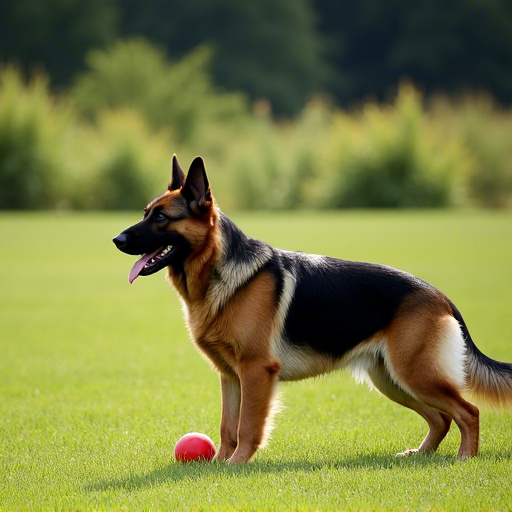} &
            \includegraphics[width=0.11\textwidth]{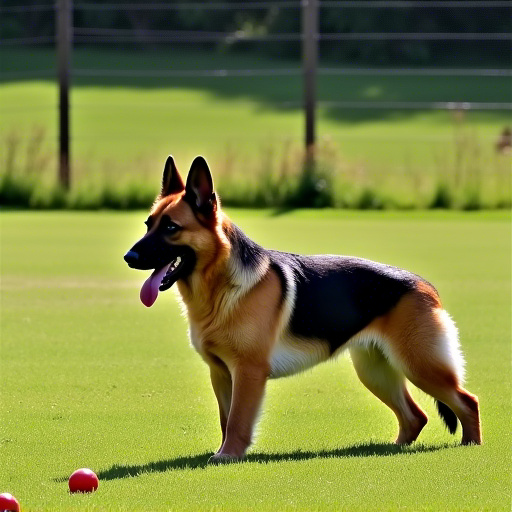} &
            \includegraphics[width=0.11\textwidth]{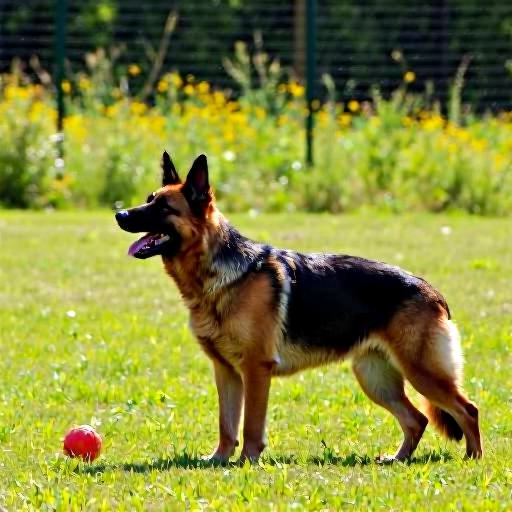} &
            \includegraphics[width=0.11\textwidth]{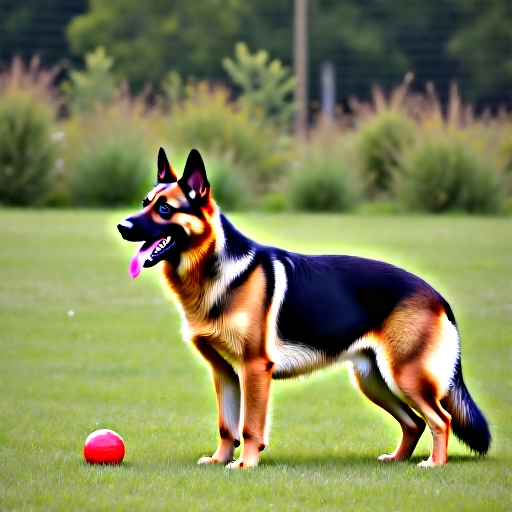} &
            \includegraphics[width=0.11\textwidth]{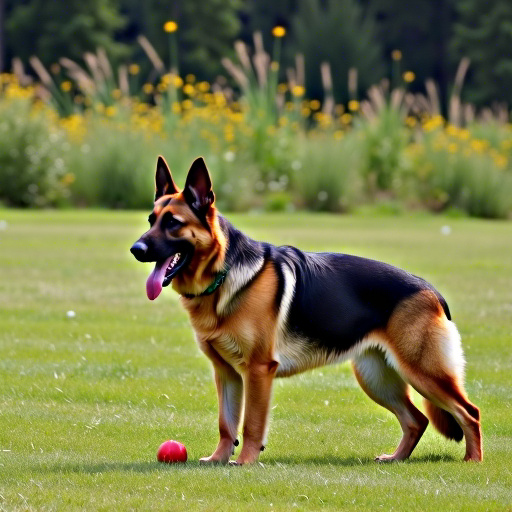} &
            \includegraphics[width=0.11\textwidth]{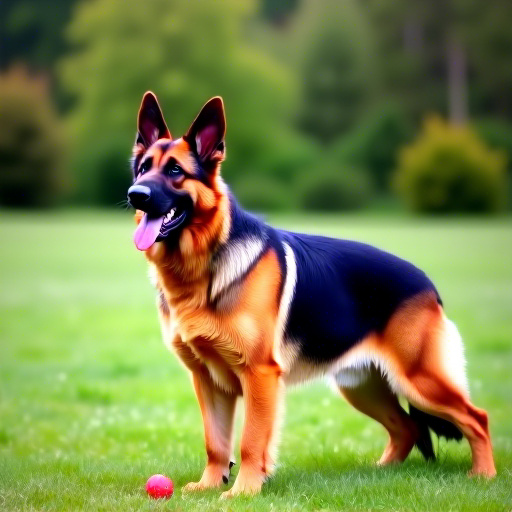} &
            \includegraphics[width=0.11\textwidth]{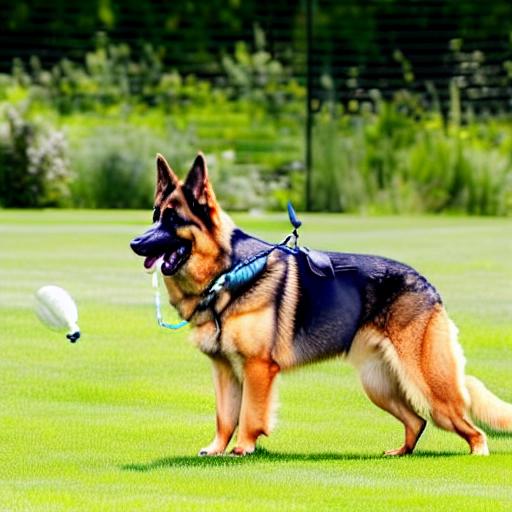} &
            \includegraphics[width=0.11\textwidth]{supplement/imgs/german/german_shepherd/pnp-inv_a_German_Shepherd_dog_stands_on_the_grass_with_mouth_opened_and_a_red_ball_next_to_its_paw.jpg} \\
            
            \scalebox{0.9}{\raisebox{0.1cm}{\rotatebox{90}{+\textcolor{BrickRed}{\textit{blue collar}}}}}&
           \includegraphics[width=0.11\textwidth]{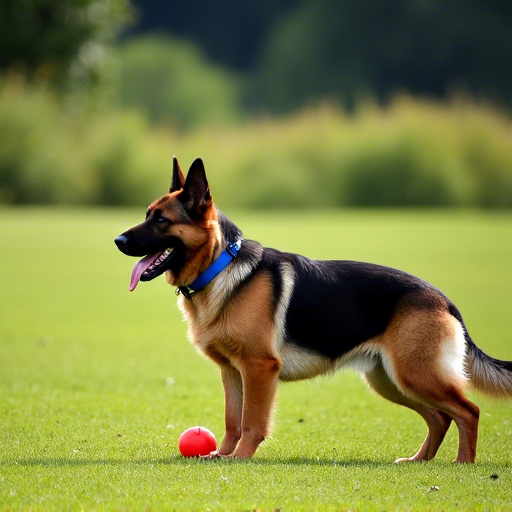} &
            \includegraphics[width=0.11\textwidth]{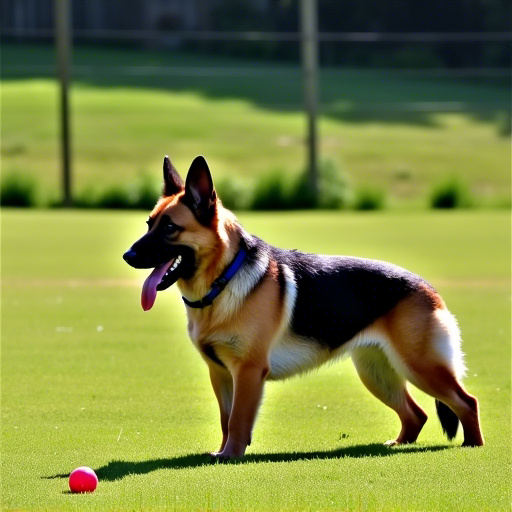} &
            \includegraphics[width=0.11\textwidth]{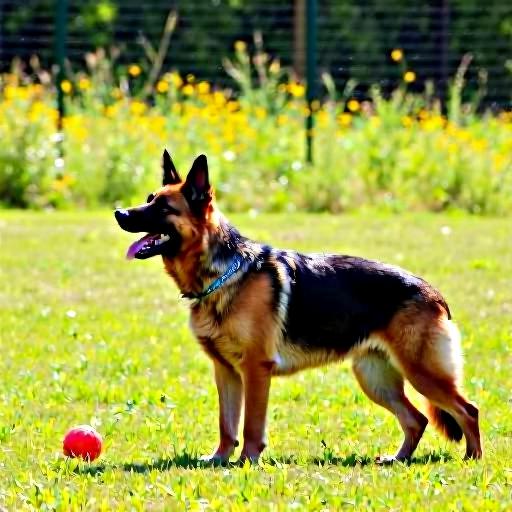} &
            \includegraphics[width=0.11\textwidth]{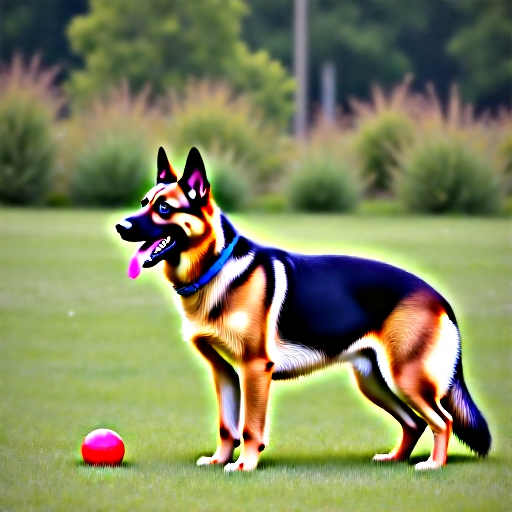} &
            \includegraphics[width=0.11\textwidth]{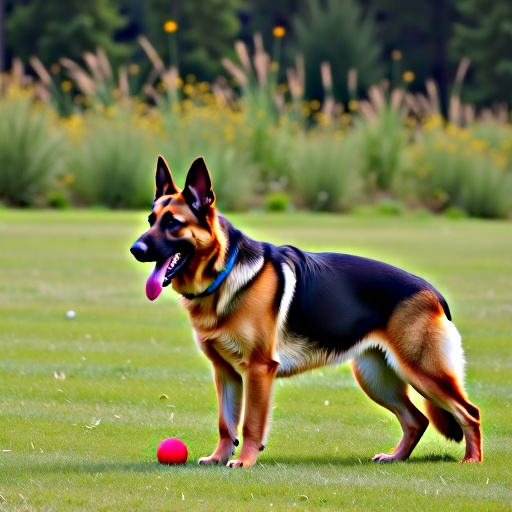} &
            \includegraphics[width=0.11\textwidth]{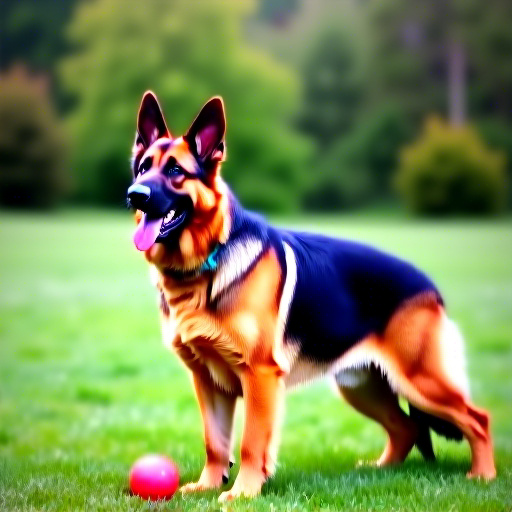} &
            \includegraphics[width=0.11\textwidth]{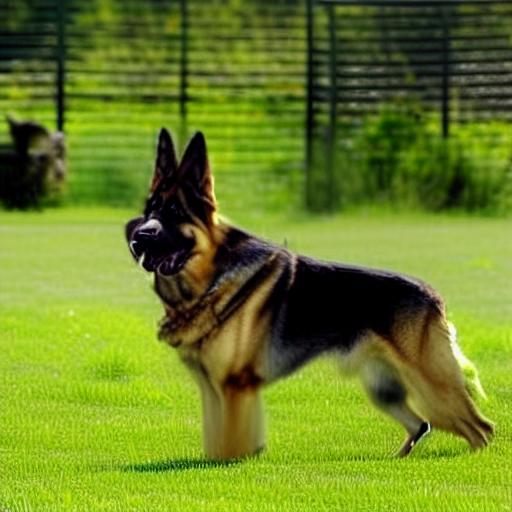} &
            \includegraphics[width=0.11\textwidth]{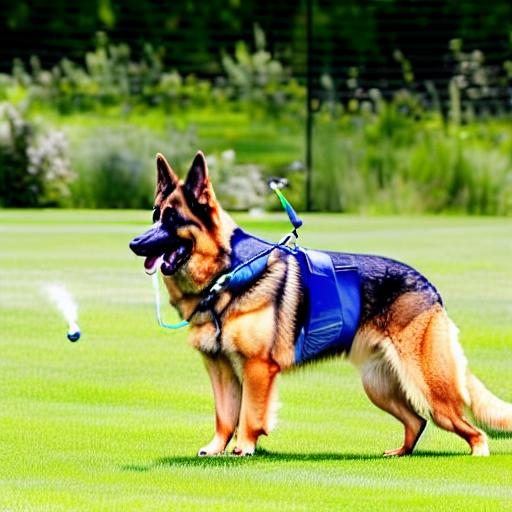}
        \end{tabular}
    \end{minipage}
    \captionsetup{skip=0pt} 
    \caption{Quantitative Results on Natural Animals. Our method successfully performs edits without introducing artifacts.}
    \label{fig:german}

\end{figure*}

\begin{figure*}[htbp]
    \centering
    \setlength{\tabcolsep}{2pt} 
    \renewcommand{\arraystretch}{0.8} 

    \begin{minipage}[t]{0.09\textwidth}
        \centering
        \vspace{-3.4cm}
        \includegraphics[width=\linewidth]{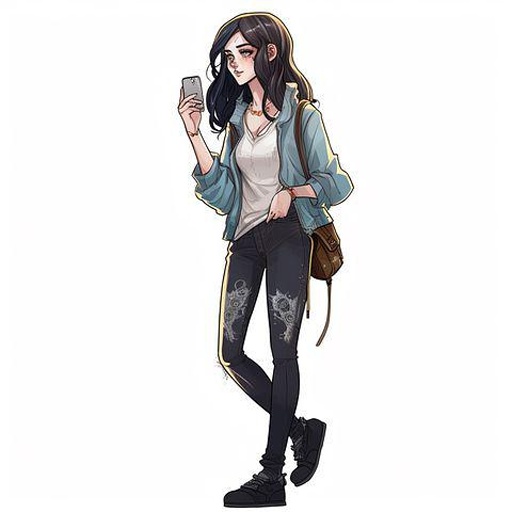}
        \vspace{-0.8cm}
        \begin{center}
        source
    \end{center}
        \vspace{0.3cm} 
        \scalebox{0.9}{\downarrowwithtext[OliveGreen, ultra thick]{\textit{Editing} \\ \textit{Turn}}}
        
    \end{minipage}%
    \hspace{2em}
    \begin{minipage}[t]{0.84\textwidth}
        \begin{tabular}{p{0.3cm}cccccccc}
            &Ours & RF-Inv. & StableFlow & FlowEdit & RF-Solver & FireFlow &  MasaCtrl & PnPInv. \\
            \scalebox{0.9}{\raisebox{0cm}{\rotatebox{90}{phone $\rightarrow$\textcolor{BrickRed}{\textit{coffee}}}}}&
            \includegraphics[width=0.11\textwidth]{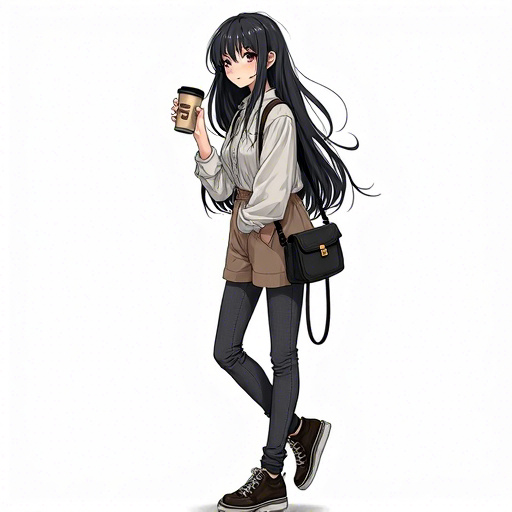} &
            \includegraphics[width=0.11\textwidth]{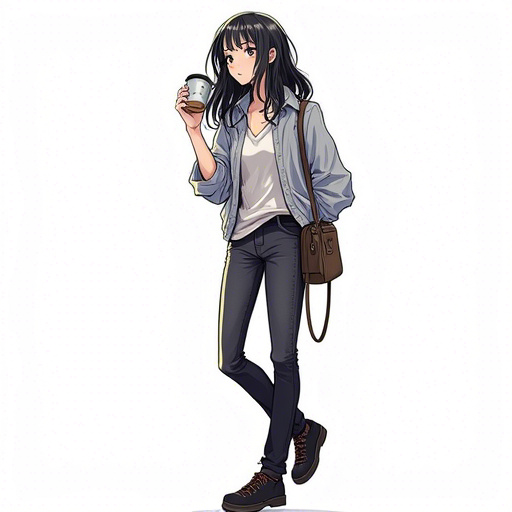} &
            \includegraphics[width=0.11\textwidth]{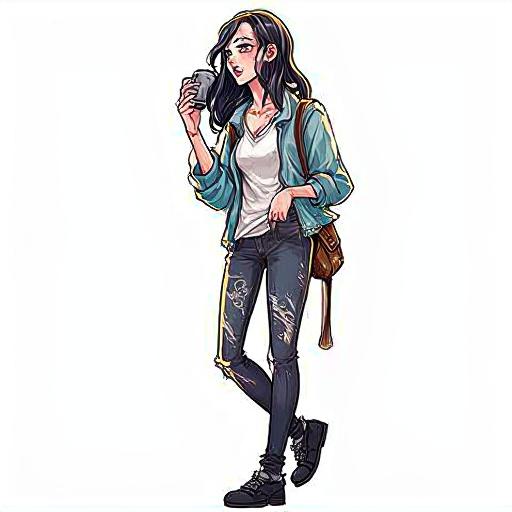} &
            \includegraphics[width=0.11\textwidth]{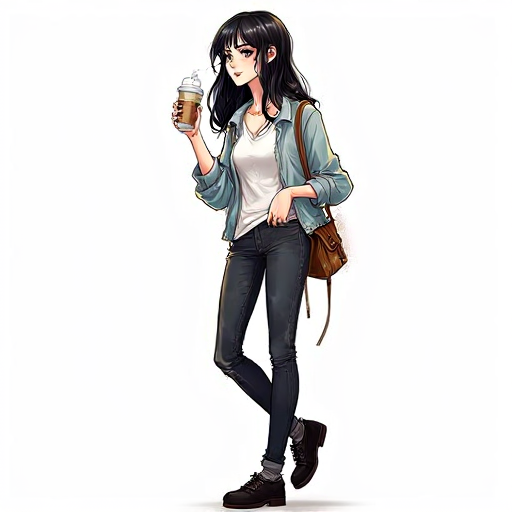} &
            \includegraphics[width=0.11\textwidth]{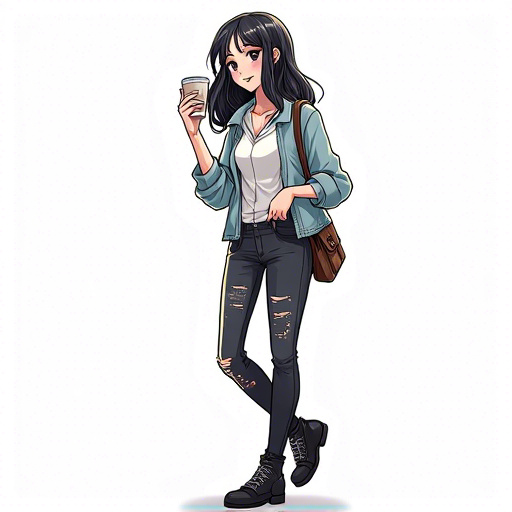} &
            \includegraphics[width=0.11\textwidth]{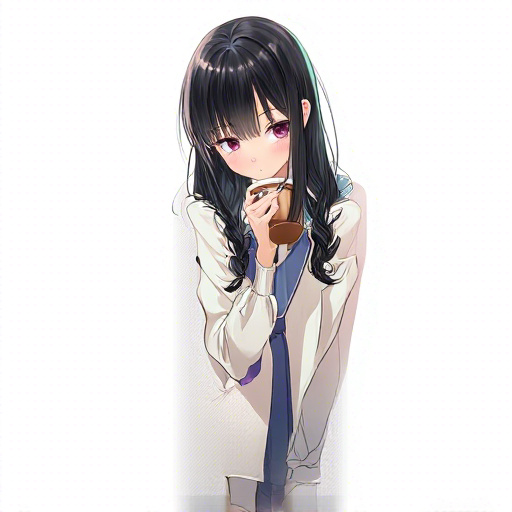} &
            \includegraphics[width=0.11\textwidth]{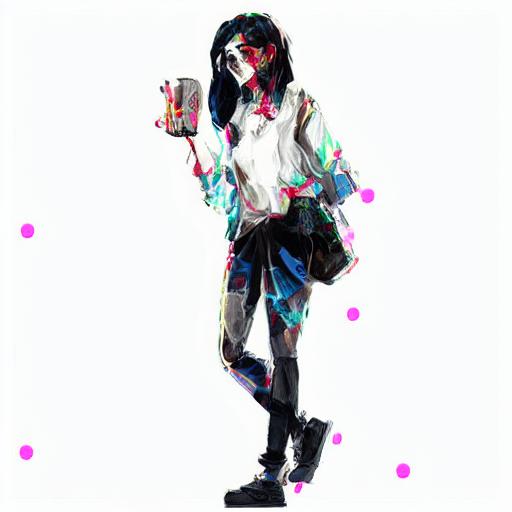} &
            \includegraphics[width=0.11\textwidth]{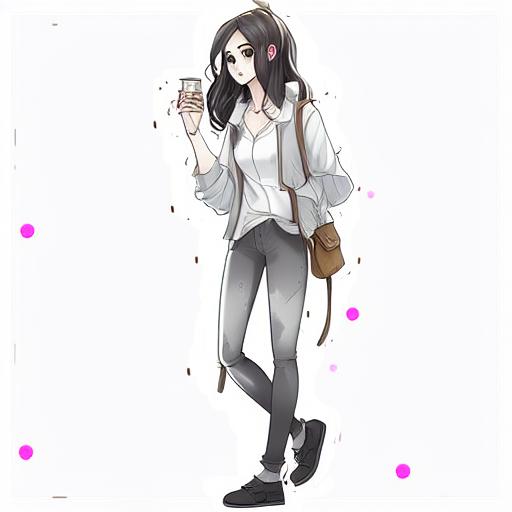} \\

            \scalebox{0.9}{\raisebox{0.2cm}{\rotatebox{90}{+ \textcolor{BrickRed}{\textit{book}}}}}&
            \includegraphics[width=0.11\textwidth]{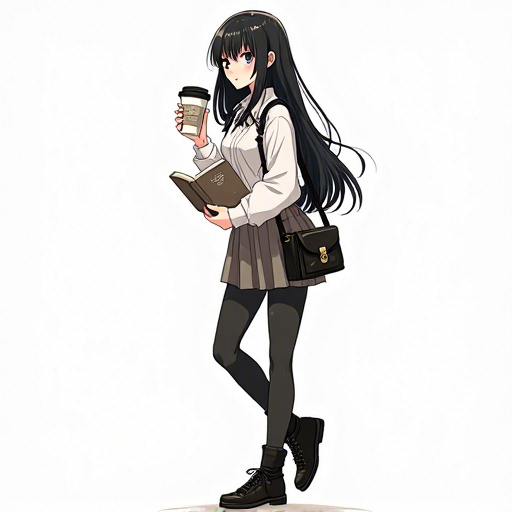} &
            \includegraphics[width=0.11\textwidth]{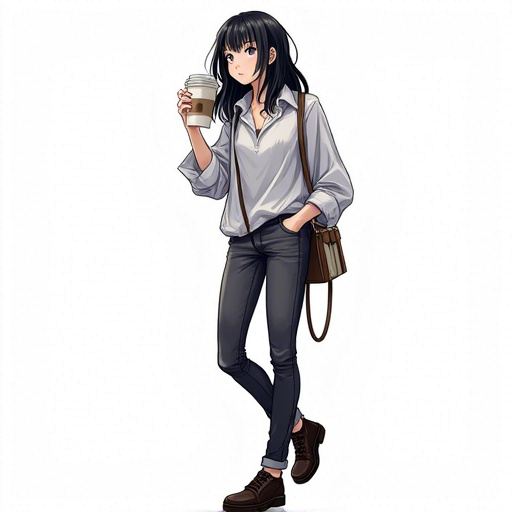} &
            \includegraphics[width=0.11\textwidth]{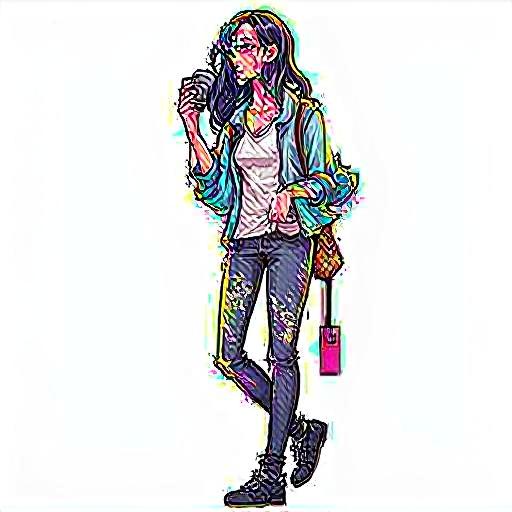} &
            \includegraphics[width=0.11\textwidth]{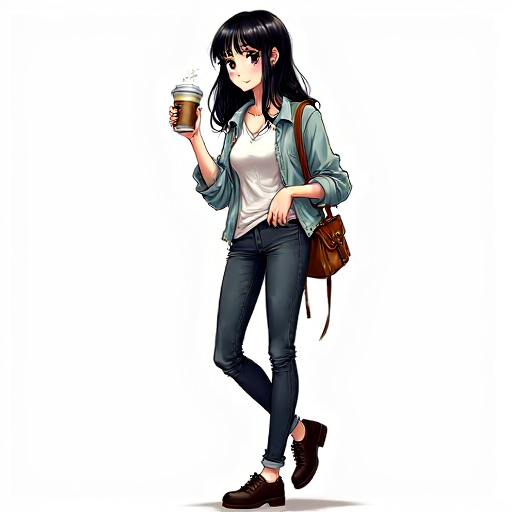} &
            \includegraphics[width=0.11\textwidth]{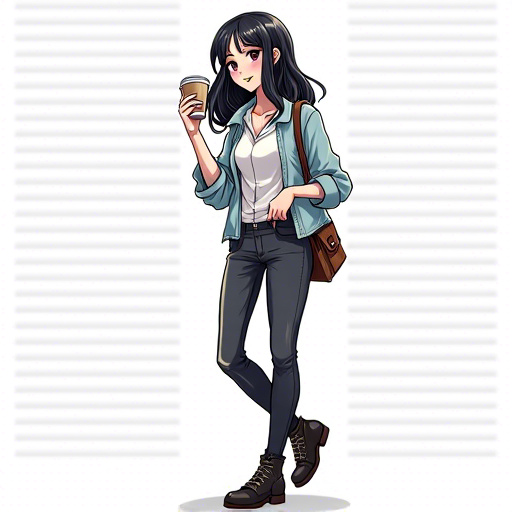} &
            \includegraphics[width=0.11\textwidth]{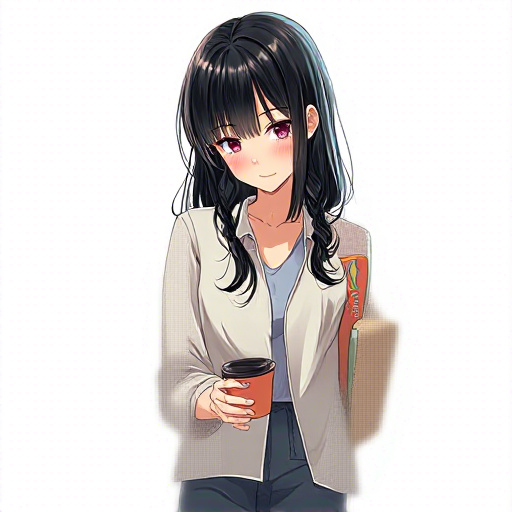}&
            \includegraphics[width=0.11\textwidth]{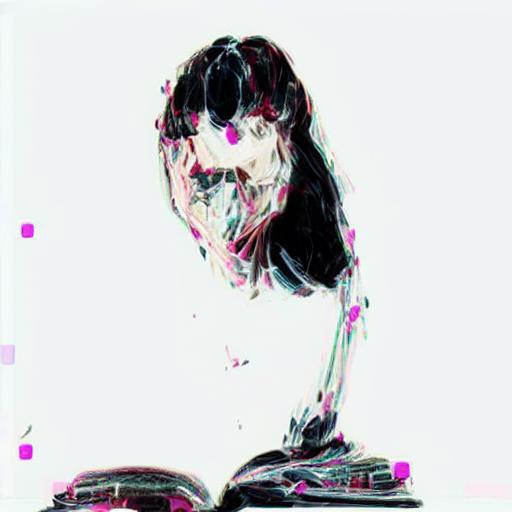} &
            \includegraphics[width=0.11\textwidth]{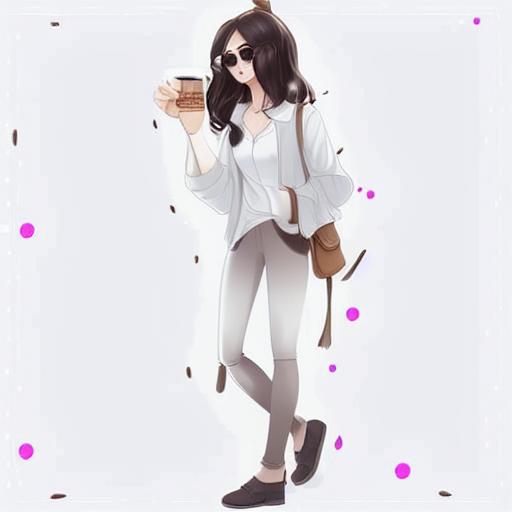}  \\
            
            \scalebox{0.9}{\raisebox{0cm}{\rotatebox{90}{ \textcolor{BrickRed}{\textit{[red]}} skirt}}}&
            \includegraphics[width=0.11\textwidth]{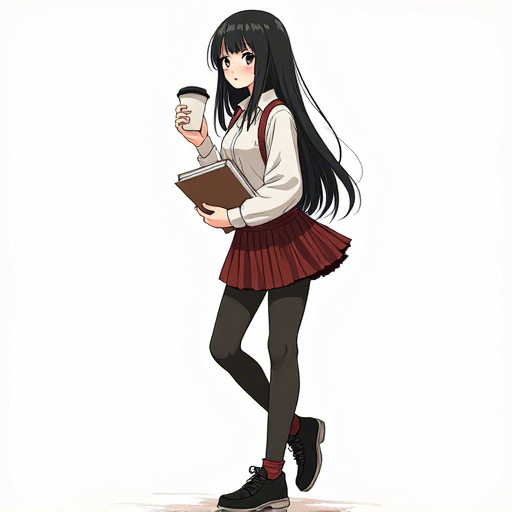} &
            \includegraphics[width=0.11\textwidth]{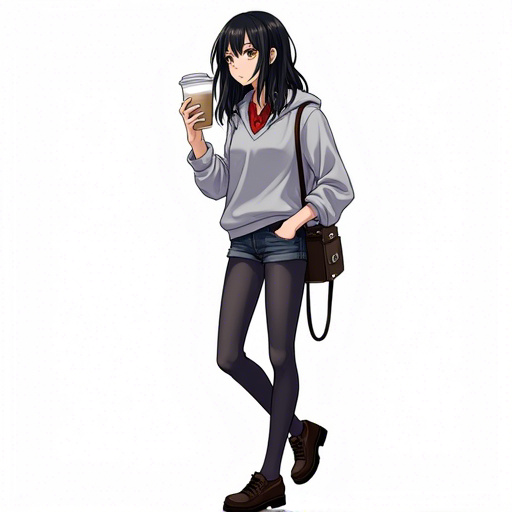} &
            \includegraphics[width=0.11\textwidth]{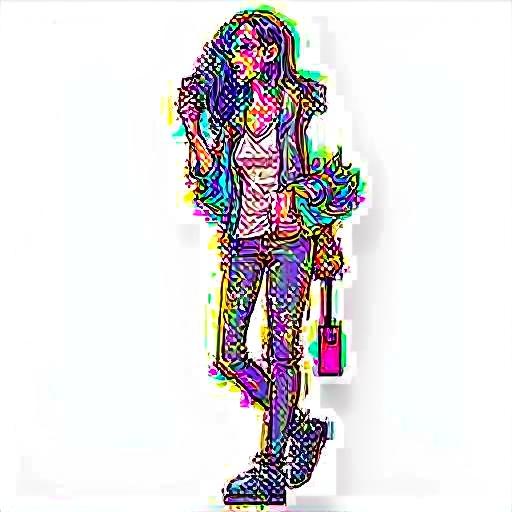} &
            \includegraphics[width=0.11\textwidth]{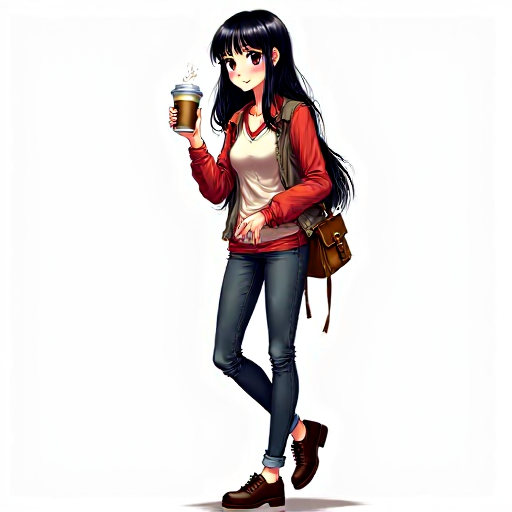} &
            \includegraphics[width=0.11\textwidth]{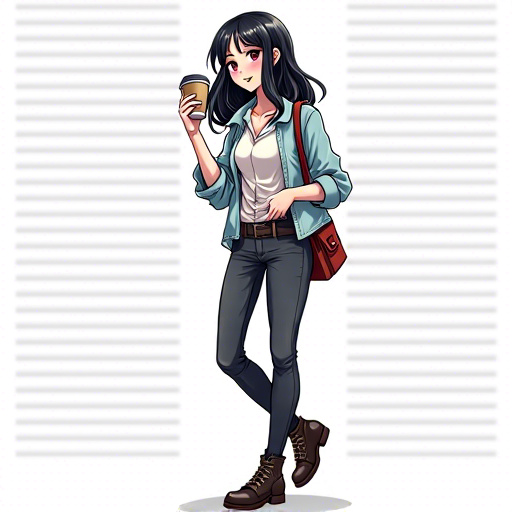} &
            \includegraphics[width=0.11\textwidth]{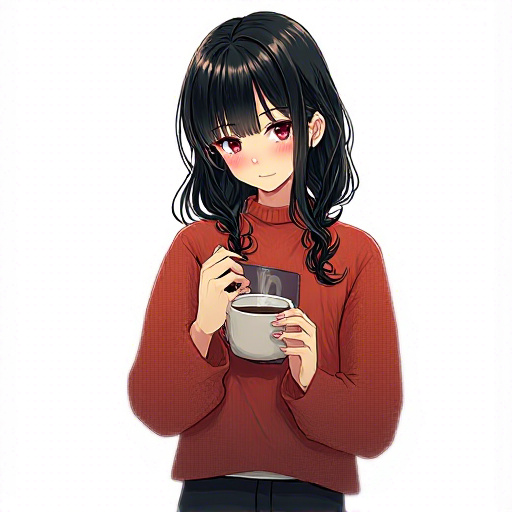} &
            \includegraphics[width=0.11\textwidth]{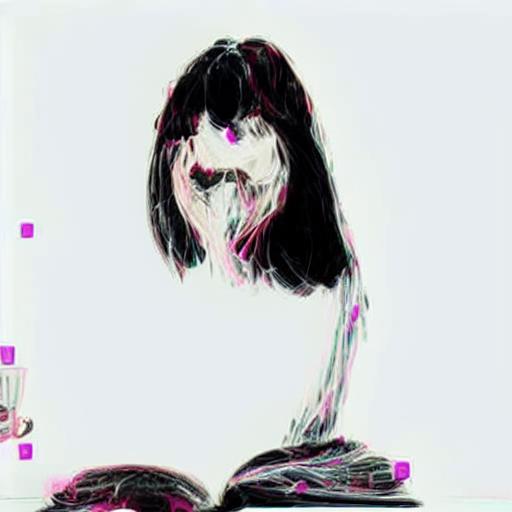} &
            \includegraphics[width=0.11\textwidth]{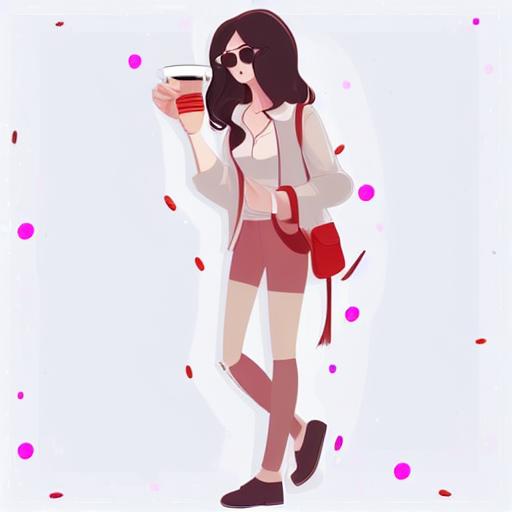} \\
            
            \scalebox{0.9}{\raisebox{0.3cm}{\rotatebox{90}{\textcolor{BrickRed}{\textit{library}}}}}&
           \includegraphics[width=0.11\textwidth]{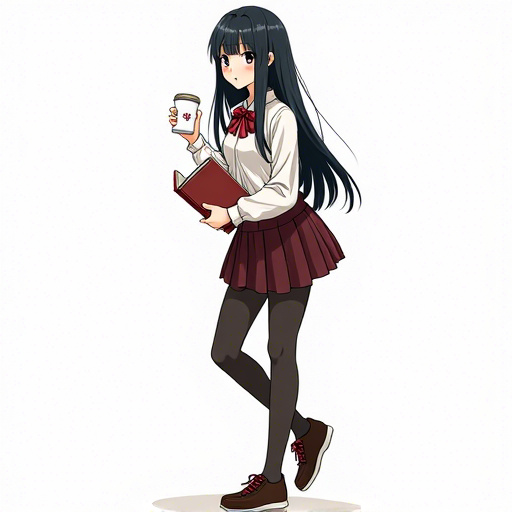} &
            \includegraphics[width=0.11\textwidth]{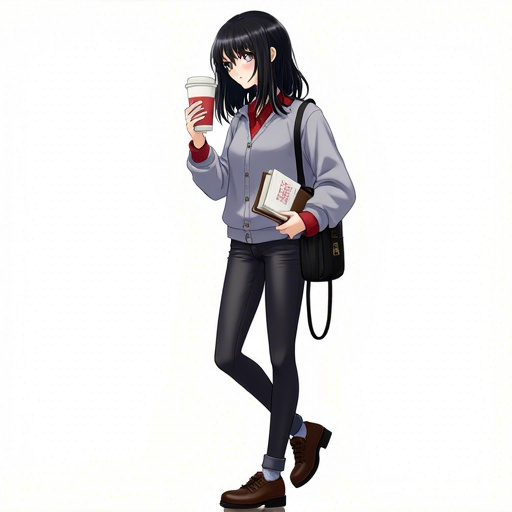} &
            \includegraphics[width=0.11\textwidth]{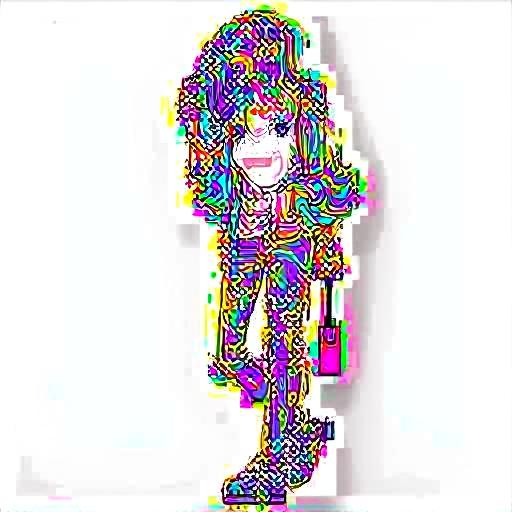} &
            \includegraphics[width=0.11\textwidth]{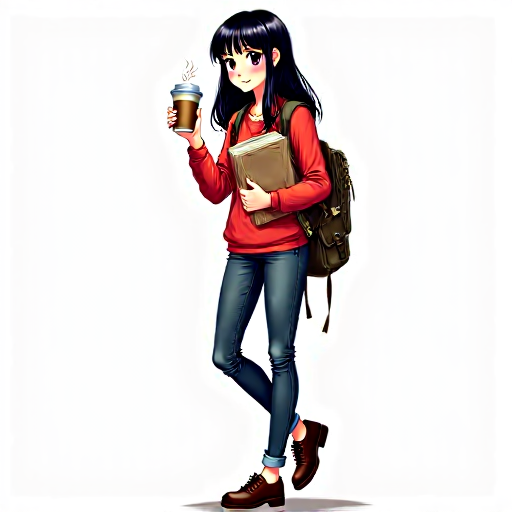} &
            \includegraphics[width=0.11\textwidth]{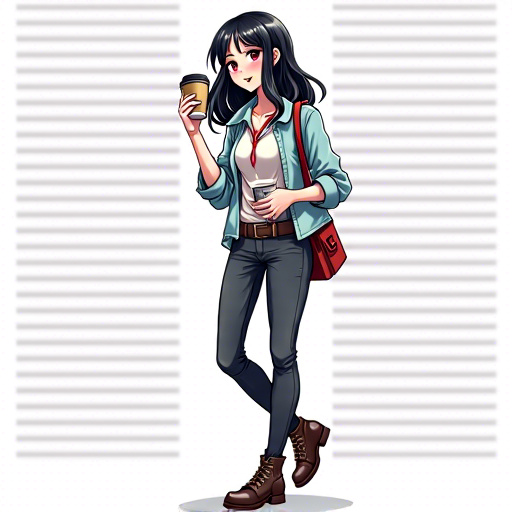} &
            \includegraphics[width=0.11\textwidth]{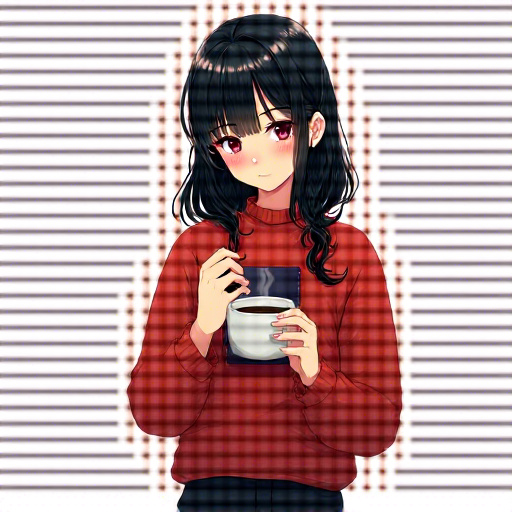} &
            \includegraphics[width=0.11\textwidth]{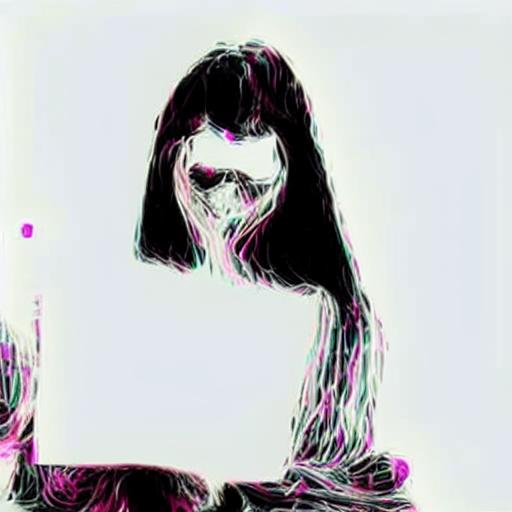} &
            \includegraphics[width=0.11\textwidth]{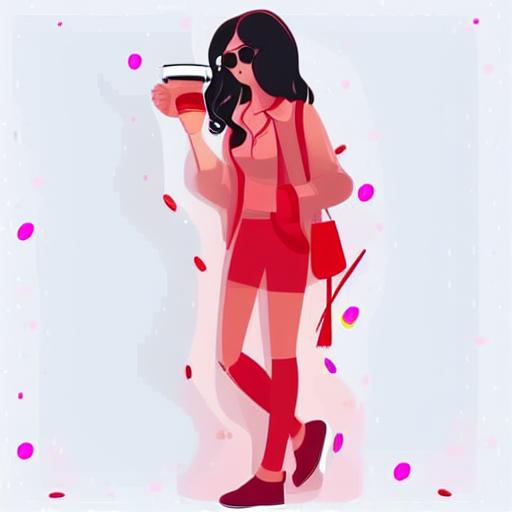}
        \end{tabular}
    \end{minipage}

    \captionsetup{skip=0pt} 
    \caption{Quantitative results on artificial images show that our method successfully preserves the background while performing the editing.}
    \label{fig:girl}

\end{figure*}

\begin{figure*}[htbp]
    \centering
    \setlength{\tabcolsep}{2pt} 
    \renewcommand{\arraystretch}{0.8} 

    \begin{minipage}[t]{0.09\textwidth}
        \centering
        \vspace{-3.4cm}
        \includegraphics[width=\linewidth]{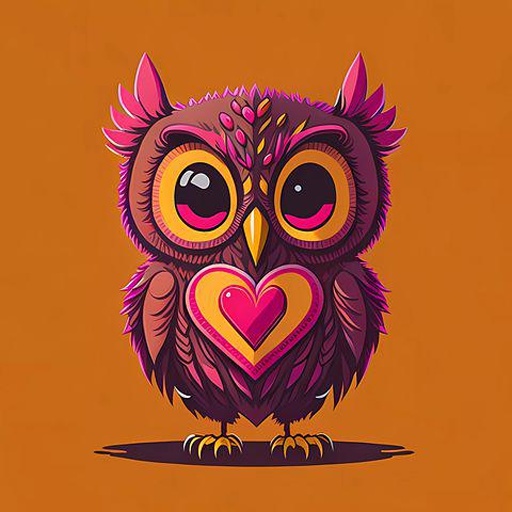}
        \vspace{-0.8cm}
        \begin{center}
        source
    \end{center}
        \vspace{0.3cm} 
        \scalebox{0.9}{\downarrowwithtext[OliveGreen, ultra thick]{\textit{Editing} \\ \textit{Turn}}}
        
    \end{minipage}%
    \hspace{2em}
    \begin{minipage}[t]{0.84\textwidth}
        \begin{tabular}{p{0.4cm}cccccccc}
            &Ours & RF-Inv. & StableFlow & FlowEdit & RF-Solver & FireFlow &  MasaCtrl & PnPInv. \\
            \scalebox{0.9}{\raisebox{0.4cm}{\rotatebox{90}{heart $\rightarrow$\textcolor{BrickRed}{\textit{circle}}}}}&
            \includegraphics[width=0.11\textwidth]{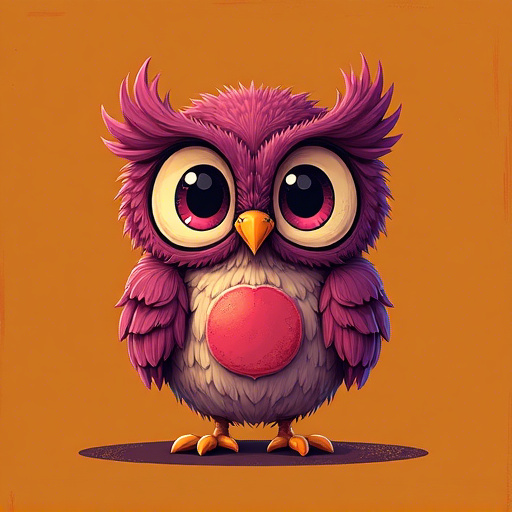} &
            \includegraphics[width=0.11\textwidth]{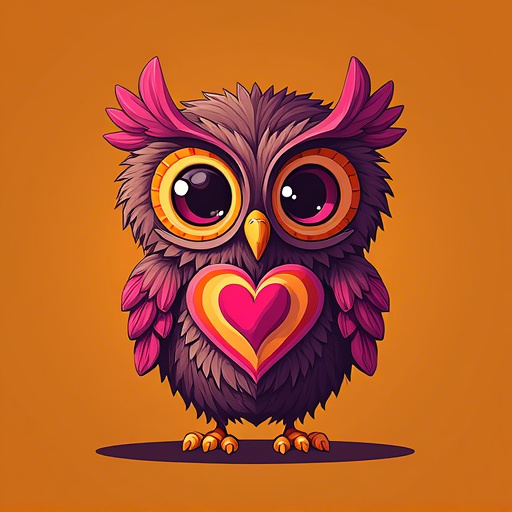} &
            \includegraphics[width=0.11\textwidth]{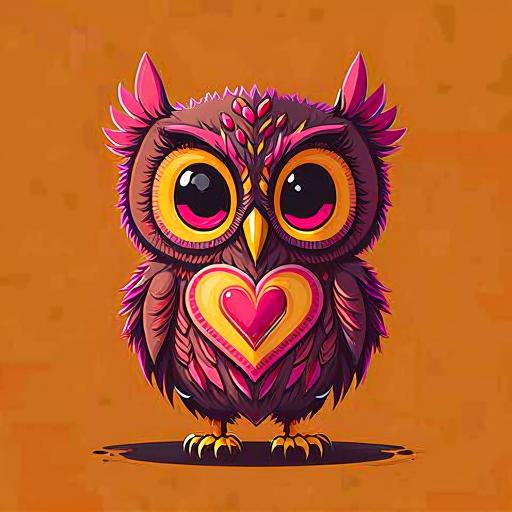} &
            \includegraphics[width=0.11\textwidth]{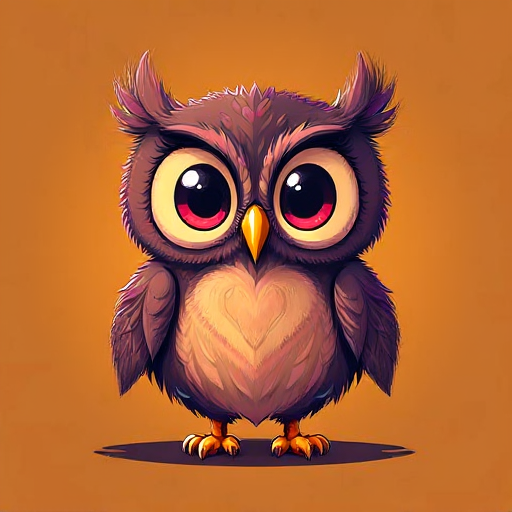} &
            \includegraphics[width=0.11\textwidth]{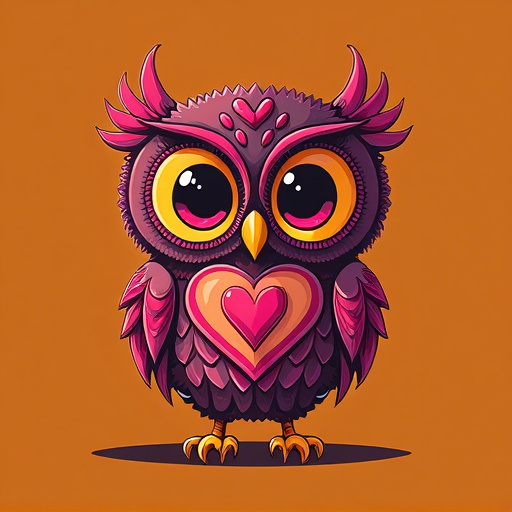} &
            \includegraphics[width=0.11\textwidth]{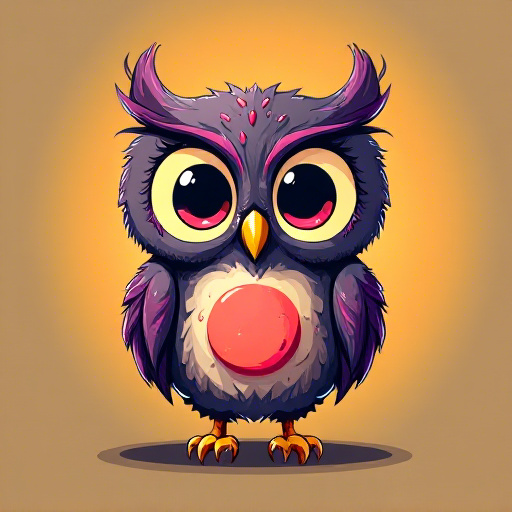} &
            \includegraphics[width=0.11\textwidth]{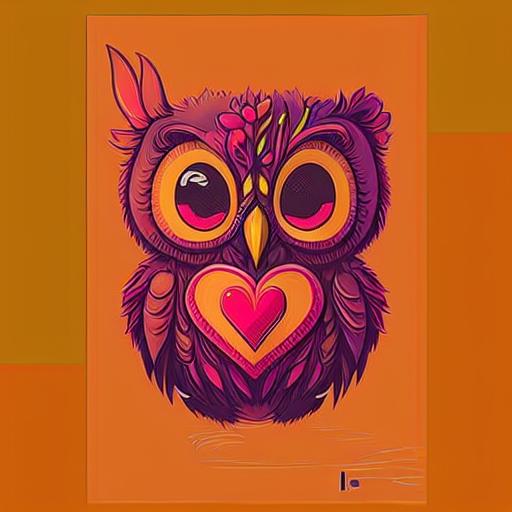} &
            \includegraphics[width=0.11\textwidth]{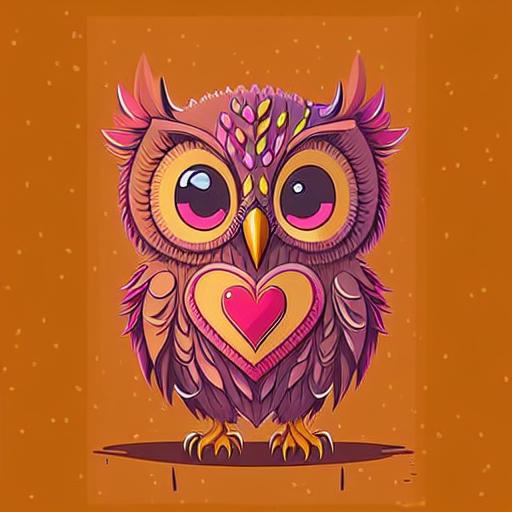} \\

            \scalebox{0.9}{\raisebox{0.cm}{\rotatebox{90}{\textcolor{BrickRed}{\textit{holding flower}}}}}&
            \includegraphics[width=0.11\textwidth]{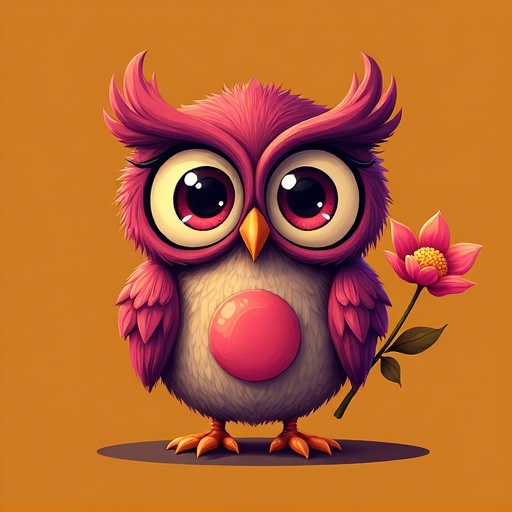} &
            \includegraphics[width=0.11\textwidth]{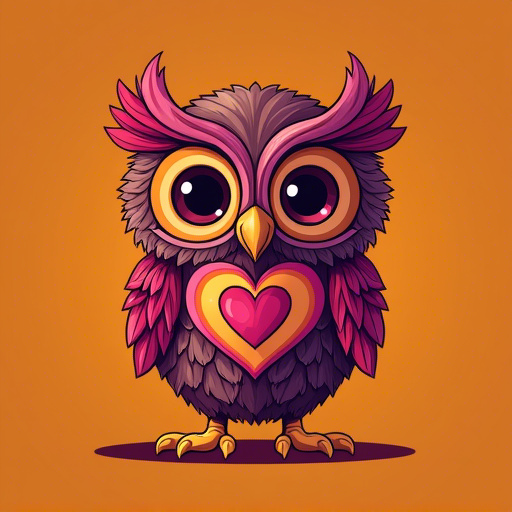} &
            \includegraphics[width=0.11\textwidth]{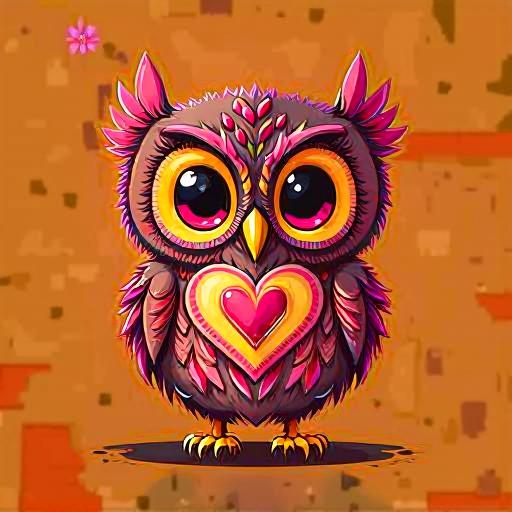} &
            \includegraphics[width=0.11\textwidth]{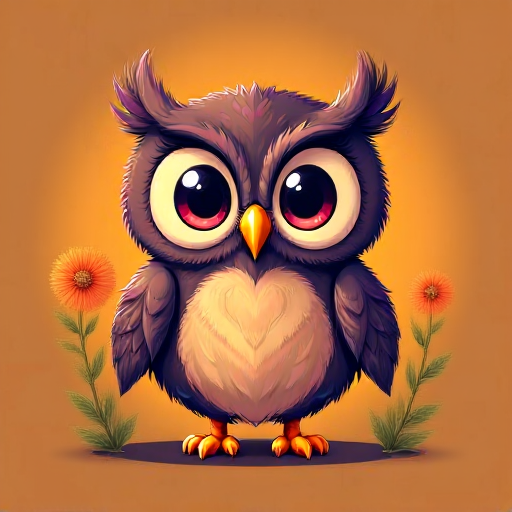} &
            \includegraphics[width=0.11\textwidth]{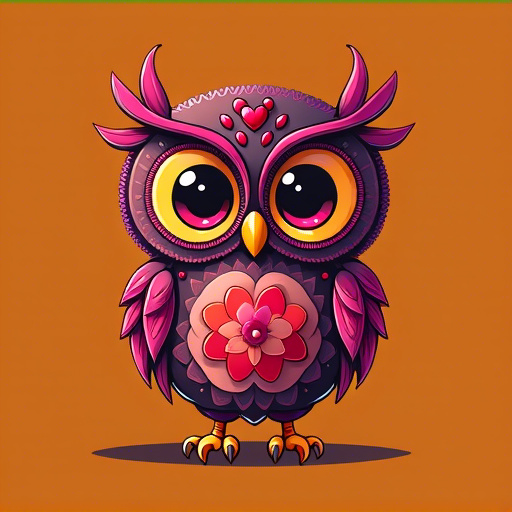} &
            \includegraphics[width=0.11\textwidth]{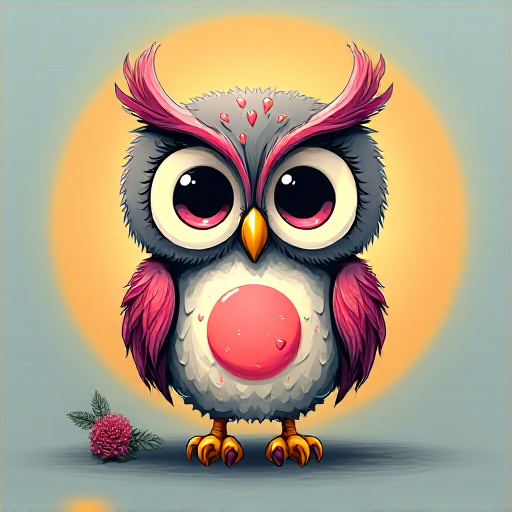}&
            \includegraphics[width=0.11\textwidth]{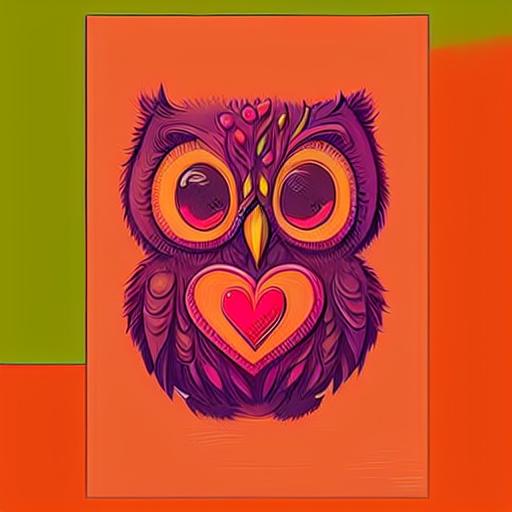} &
            \includegraphics[width=0.11\textwidth]{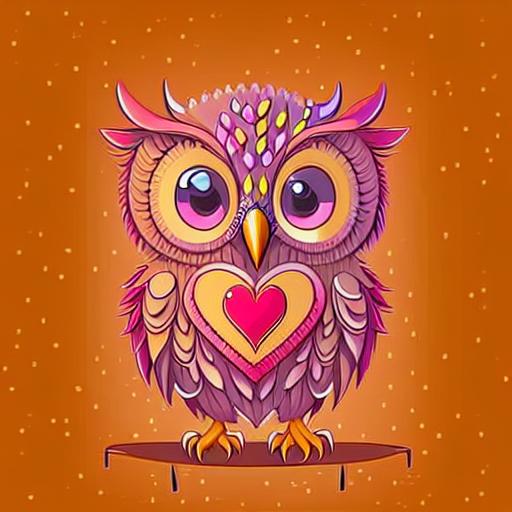}  \\
            
            \scalebox{0.9}{\raisebox{0.2cm}{\rotatebox{90}{+ \textcolor{BrickRed}{\textit{glasses}}}}}&
            \includegraphics[width=0.11\textwidth]{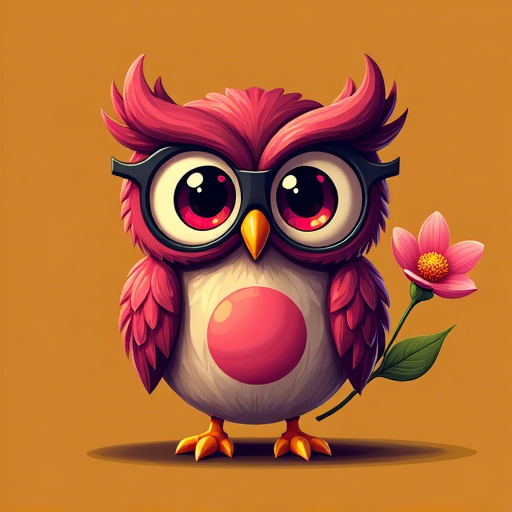} &
            \includegraphics[width=0.11\textwidth]{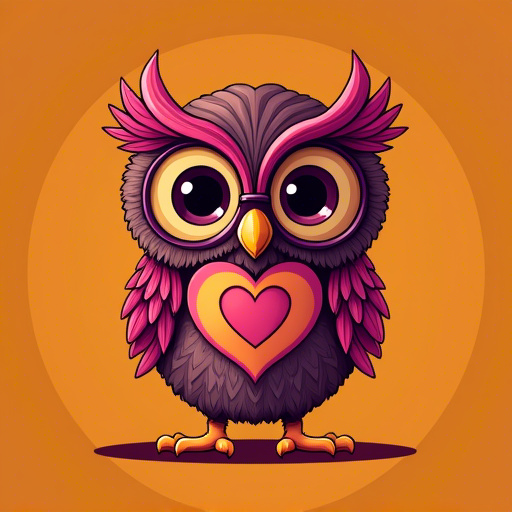} &
            \includegraphics[width=0.11\textwidth]{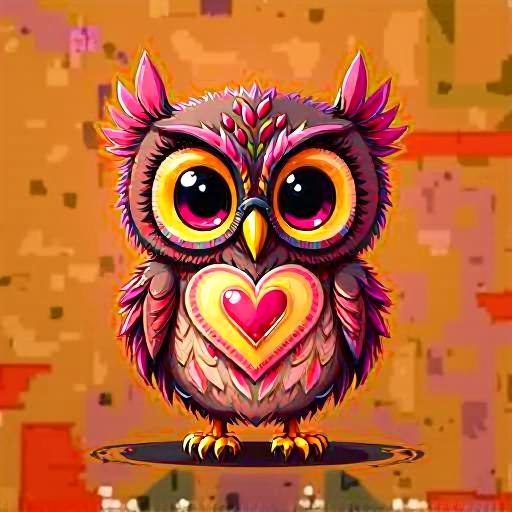} &
            \includegraphics[width=0.11\textwidth]{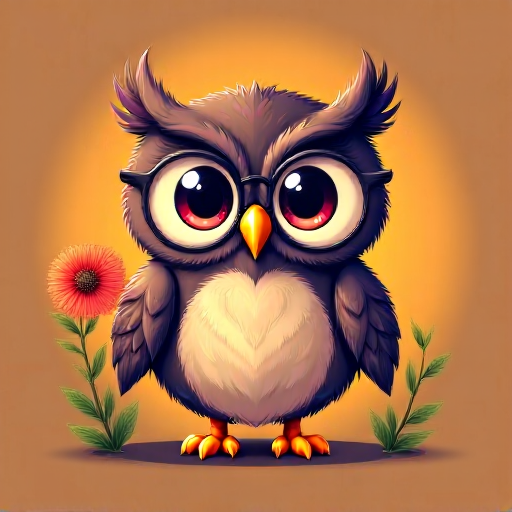} &
            \includegraphics[width=0.11\textwidth]{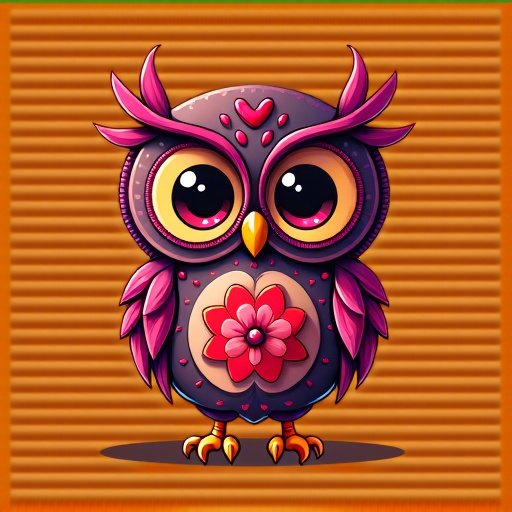} &
            \includegraphics[width=0.11\textwidth]{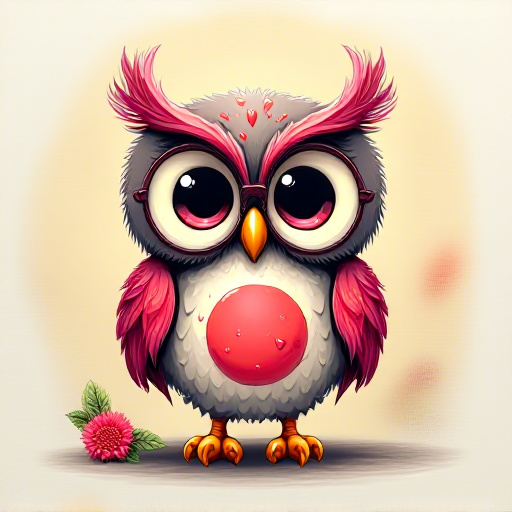} &
            \includegraphics[width=0.11\textwidth]{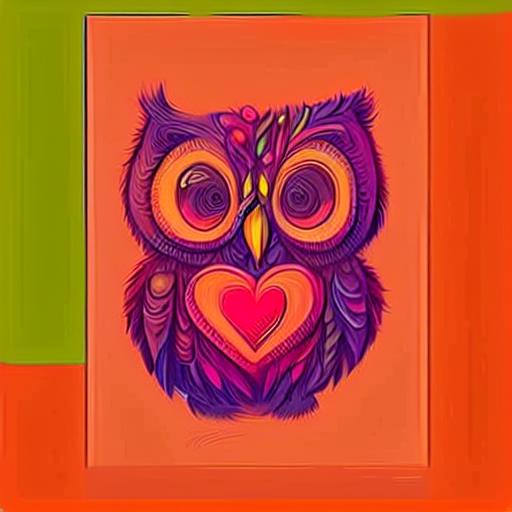} &
            \includegraphics[width=0.11\textwidth]{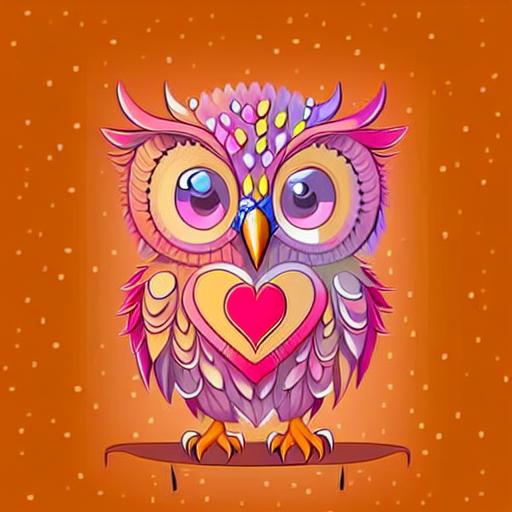} \\
            
            \scalebox{0.9}{\raisebox{-0.3cm}{\rotatebox{90}{standing on \textcolor{BrickRed}{\textit{branch}}}}}&
           \includegraphics[width=0.11\textwidth]{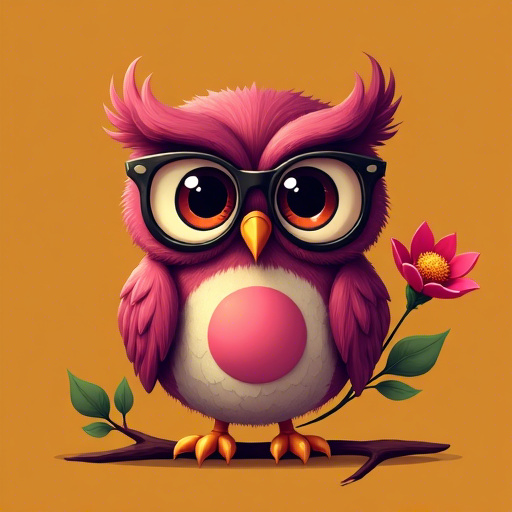} &
            \includegraphics[width=0.11\textwidth]{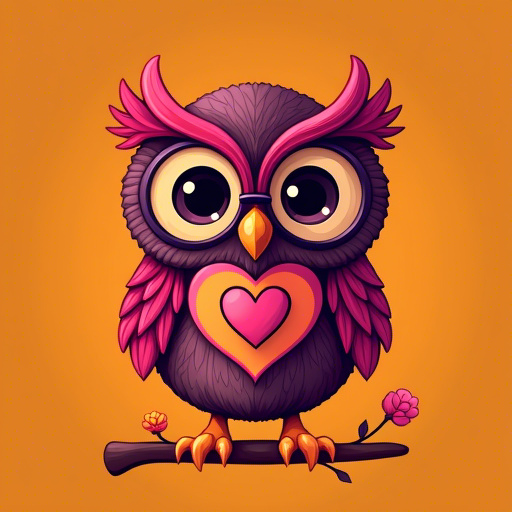} &
            \includegraphics[width=0.11\textwidth]{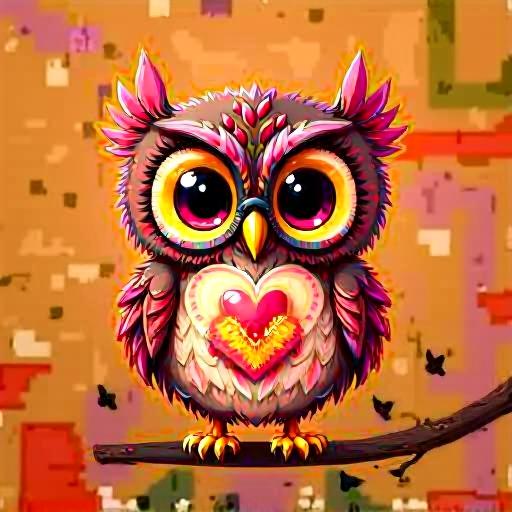} &
            \includegraphics[width=0.11\textwidth]{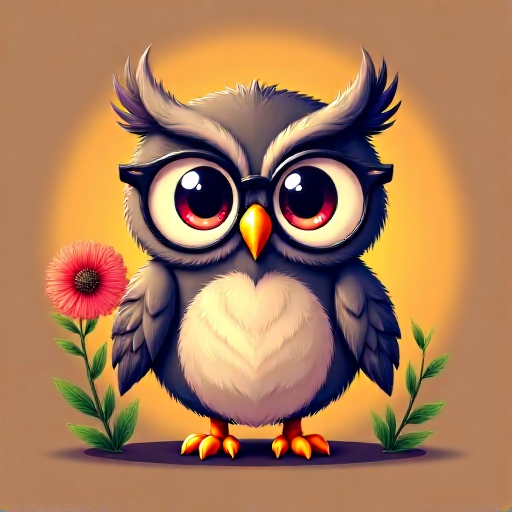} &
            \includegraphics[width=0.11\textwidth]{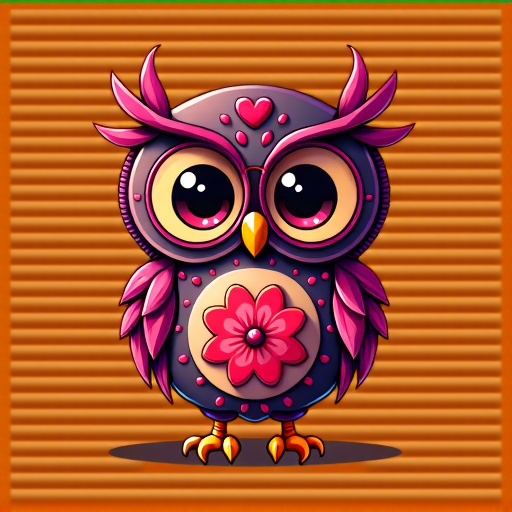} &
            \includegraphics[width=0.11\textwidth]{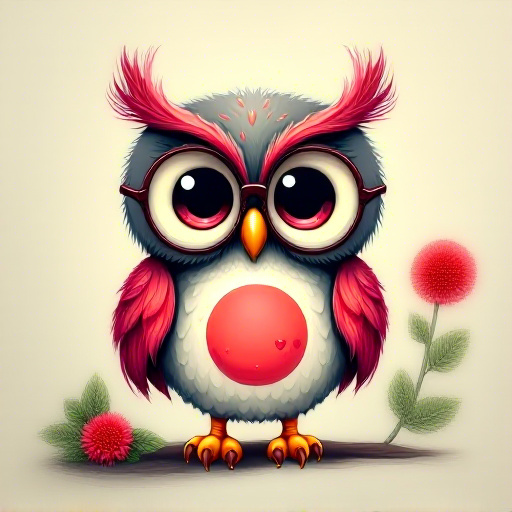} &
            \includegraphics[width=0.11\textwidth]{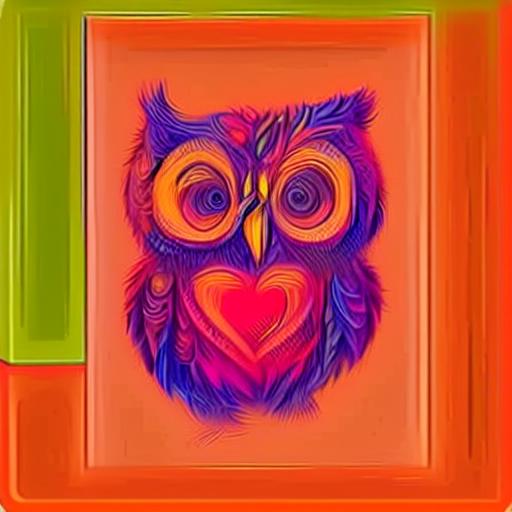} &
            \includegraphics[width=0.11\textwidth]{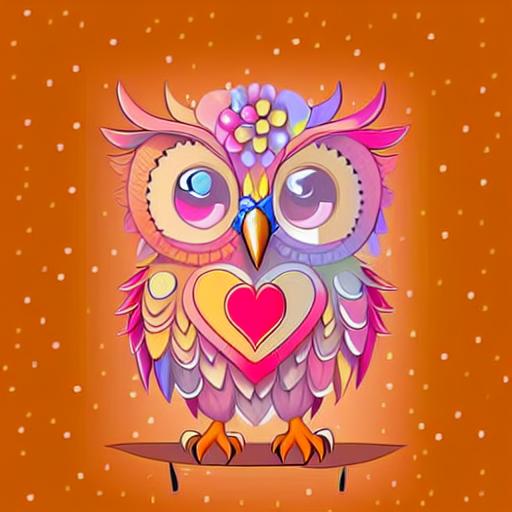}
        \end{tabular}
    \end{minipage}

    \captionsetup{skip=0pt} 
    \caption{Quantitative results on artificial images show that our method successfully preserves the background while performing the editing.}
    \label{fig:owl}

\end{figure*}

To address this, we conduct additional qualitative experiments on both natural and artificial images. The results for natural image editing are shown in ~\cref{fig:german} and ~\cref{fig:butterfly}, while artificial image results are presented in ~\cref{fig:girl} and ~\cref{fig:owl}.

In both categories, our method achieves a high success rate in image editing. Equally important, our edited images consistently preserve key features of the original image across multiple editing steps, including color, lighting, background, and pose. This balance between content preservation and editing effectiveness aligns with the quantitative results in ~\cref{sup: quantative}.
Artistc paintings are almost the most dofficult category if image to editing and reconstruction.
We present our full experimtnal rresults multi-turn reconstruciton on artisic painting. Our methods is almost can do all the way done.

\end{document}